\documentclass[10pt,twocolumn,letterpaper]{article}

\usepackage[pagenumbers]{cvpr}      %

\usepackage{microtype}

\usepackage{amsmath}
\usepackage{amssymb}
\usepackage{booktabs}
\usepackage{tabularx}
\usepackage{array}
\usepackage{xcolor}
\usepackage{multirow}
\usepackage{multicol}
\usepackage{colortbl}
\usepackage{graphicx}
\usepackage{tikz}
\usetikzlibrary{shapes.misc}
\usepackage{cuted}

\definecolor{FlowerSanityTint}{gray}{0.94}
\definecolor{FlowerSanityInk}{gray}{0.30}
\definecolor{FlowerOursTint}{RGB}{235,245,243}
\definecolor{FlowerOursInk}{RGB}{24,92,82}
\definecolor{FlowerSPITint}{RGB}{244,240,250}
\definecolor{FlowerSPIInk}{RGB}{92,62,138}
\definecolor{FlowerDivider}{gray}{0.80}
\newcommand{\FlowerScore}[1]{$#1$}
\newcommand{\FlowerBest}[1]{\FlowerScore{\mathbf{#1}}}
\newcommand{\FlowerSecond}[1]{\FlowerScore{\underline{#1}}}
\newcommand{\FlowerWarning}[1]{\begingroup\setlength{\fboxsep}{1pt}\setlength{\fboxrule}{0.4pt}\fbox{\FlowerScore{#1}}\endgroup}
\newcommand{\FlowerUnavailable}{\textnormal{n/a}}
\newcommand{\FlowerSanityLabel}{\multirow[c]{-2}{\linewidth}{\centering\resizebox{\linewidth}{!}{\rotatebox[origin=c]{90}{\shortstack{\textsf{Sanity}\\\textsf{checks}}}}}}
\newcommand{\FlowerIGLabel}{\multirow[c]{-10}{\linewidth}[\dimexpr\aboverulesep+\belowrulesep+\cmidrulewidth\relax]{\centering\rotatebox[origin=c]{90}{\textsf{IG Family}}}}
\newcommand{\FlowerSPILabel}{\multirow[c]{-2}{\linewidth}{\centering\rotatebox[origin=c]{90}{\textcolor{FlowerSPIInk}{\textsf{\textbf{Abl.}}}}}}
\newcommand{\FlowerOursLabel}{\multirow[c]{-2}{\linewidth}{\centering\rotatebox[origin=c]{90}{\textcolor{FlowerOursInk}{\textsf{\textbf{Ours}}}}}}
\newcommand{\FlowerTableLayout}{%
  \setlength{\tabcolsep}{2.5pt}%
  \renewcommand{\arraystretch}{1.15}%
  \setlength{\aboverulesep}{1.5pt}%
  \setlength{\belowrulesep}{1.5pt}%
}
\newcommand{\FlowerGroupHeaders}{%
  \multicolumn{3}{l}{}
  & \multicolumn{5}{c}{\scriptsize\textsf{Faithfulness}}
  & \multicolumn{3}{c}{\scriptsize\textsf{Localization}}
  & \multicolumn{1}{c}{\scriptsize\textsf{Class sensitivity}} \\
  \cmidrule(lr){4-8}\cmidrule(lr){9-11}\cmidrule(lr){12-12}
}
\newcommand{\FlowerHeaders}{%
  & & \textbf{Explainer}
  & Del.\ AUC\,$\downarrow$
  & Ins.\ AUC\,$\uparrow$
  & FC\,$\uparrow$
  & MC\,$\uparrow$
  & FE\,$\uparrow$
  & PG\,$\uparrow$
  & TopK\,$\uparrow$
  & RRA\,$\uparrow$
  & RandomLogit\,$\downarrow$ \\
}

\newcommand{\FlowerAblationIGLabel}{\multirow[c]{-4}{\linewidth}[\dimexpr(\aboverulesep+\belowrulesep+\cmidrulewidth)/2\relax]{\centering\rotatebox[origin=c]{90}{\textsf{IG Family}}}}
\newcommand{\FlowerAblationOursLabel}{\centering\rotatebox[origin=c]{90}{\textcolor{FlowerOursInk}{\tiny\textsf{\textbf{Ours}}}}}

\definecolor{accentblue}{RGB}{25, 85, 150}
\definecolor{slategray}{RGB}{70, 85, 100}

\newtheorem{proposition}{Proposition}

\definecolor{InterpLinearColor}{HTML}{007F7B} %
\definecolor{ExcursionColor}{HTML}{7856A6}    %
\definecolor{CouplingColor}{HTML}{AC6B16}     %

\newcommand{\linpath}[1]{\textcolor{InterpLinearColor}{#1}}
\newcommand{\excurs}[1]{\textcolor{ExcursionColor}{#1}}
\newcommand{\coupl}[1]{\textcolor{CouplingColor}{#1}}

\definecolor{GroupA}{HTML}{482878}
\definecolor{GroupB}{HTML}{E30280}
\definecolor{GroupC}{HTML}{D91621}
\definecolor{GroupD}{HTML}{E46422}

\DeclareRobustCommand{\circnumPill}[2]{\texorpdfstring{%
\tikz[baseline=(char.base)]{
    \node[
        rounded rectangle,
        fill=#1,
        text=white,
        inner xsep=2.5pt,
        inner ysep=2pt,
        minimum height=9pt,
        font=\bfseries\scriptsize
    ] (char) {#2};
}}{(#2)}}

\DeclareRobustCommand{\circnumA}[1]{\circnumPill{GroupA}{#1}}
\DeclareRobustCommand{\circnumB}[1]{\circnumPill{GroupB}{#1}}
\DeclareRobustCommand{\circnumC}[1]{\circnumPill{GroupC}{#1}}
\DeclareRobustCommand{\circnumD}[1]{\circnumPill{GroupD}{#1}}

\definecolor{cvprblue}{rgb}{0.21,0.49,0.74}
\usepackage[pagebackref,breaklinks,colorlinks,allcolors=cvprblue]{hyperref}

\def\confName{CVPR}
\def\confYear{2027}

\title{Do Feature Attributions Need Generative Flows? \\ A Theoretical and Empirical Reality Check}

\author{
  Emre Kavak\textsuperscript{1,2,3} \qquad
  Christian Wachinger\textsuperscript{1,2,3} \\
  \textsuperscript{1}Technical University of Munich, Germany \\
  \textsuperscript{2}Konrad Zuse School of Excellence in Reliable AI (relAI), Germany \\
  \textsuperscript{3}Munich Center for Machine Learning (MCML), Germany \\
  {\tt\small \{emre.kavak, christian.wachinger\}@tum.de}
}

\begin{document}
\maketitle
\begin{strip}
\centering
\includegraphics[width=\linewidth]{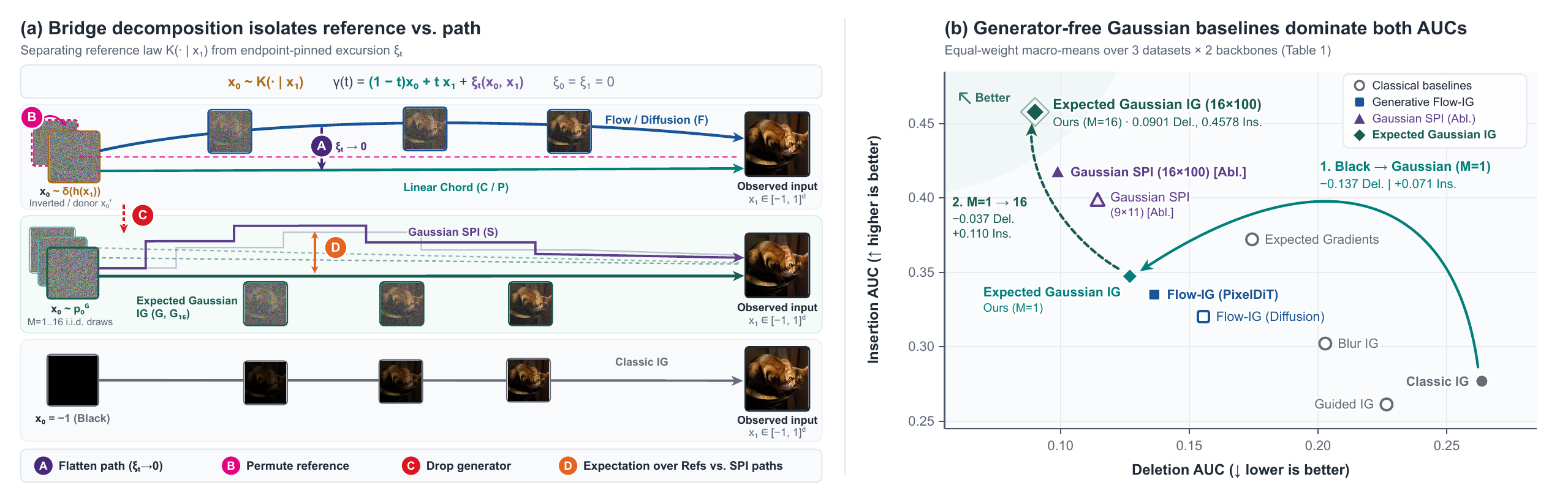}\hfill
\captionsetup{hypcap=false}
\captionof{figure}{\textbf{Baseline--path (bridge) decomposition and empirical reality check.} (a)~Any path-based explainer separates into two choices: \coupl{baseline selection $K(\cdot \mid x_1)$} (fixing both the joint pairing $\coupl{\pi}$ and baseline marginal $p_0$) and a \excurs{path deviation $\xi_t$} bending away from the \linpath{straight chord $\gamma_{\text{lin}}(t)$}. (b)~Across three datasets and two classifier backbones, replacing the black baseline with independent Gaussian draw(s) clamped to $[-1, 1]^d$ along a straight line (\textbf{Expected Gaussian IG}, M=1 \& M=16) outperforms both generative Flow-IG backends on Deletion and Insertion AUC without any generator (ablation labels $F, C, P, G, S$ are defined in Section~\ref{sec:experiments}).}
\captionsetup{hypcap=true}
\label{fig:main_fig}%
\end{strip}

\begin{abstract}
Integrated Gradients (IG) rests on a principled axiomatic foundation, yet its standard black-baseline straight-line formulation suffers from baseline blindness and out-of-distribution artifacts. Recent work replaces simple IG with curved trajectories from diffusion and flow-matching models, reporting strong gains while altering multiple design factors at once.
Through a baseline--path decomposition, a systematic analysis of recent methods, and controlled experiments, we debunk the claim that generative paths benefit IG. Our framework separates baseline selection from path geometry, showing that exact Euclidean quadratic optimal-transport flows collapse to straight lines and that recent generative explainers confound baseline shifts with curved trajectories. Across vision benchmarks, architectures, and generative backends, neither curved generative paths nor input-specific baseline pairings (couplings) consistently improve attribution quality; instead, a simple generator-free Gaussian baseline matches or outperforms heavy generative flows at zero generator cost, and averaging straight lines over independent Gaussian baselines (Expected Gaussian IG) achieves the strongest performance overall.
Our results demonstrate that generative attribution pipelines must be ablated sequentially, separating baseline distribution, input-specific pairing, and trajectory curvature, against straight-line and random-baseline controls before attributing gains to learned manifolds or flows. Our code is available at \url{https://github.com/yakamoz5/expected-gaussian-ig}.
\end{abstract}
    
\section{Introduction}
\label{sec:intro}

Integrated Gradients (IG)~\cite{sundararajan2017axiomatic} explains an image classifier by scoring each pixel's contribution to the prediction. It does so by drawing a straight line from a reference image $x_0$ (conventionally an all-black image) to the input $x_1$, and integrating the model's gradients along that line. In practice, this often produces noisy, unfaithful explanations: dark pixels receive zero credit simply because they match the black reference, even when they are semantically important (\emph{baseline blindness}), and, some argue, the straight line traverses out-of-distribution interpolations (e.g., unnaturally darkened versions of $x_1$) where classifier gradients are unreliable~\cite{sturmfels2020visualizing,kapishnikov2021guided}. IG nonetheless rests on a uniquely principled foundation: grounded in continuous Aumann--Shapley game theory~\cite{aumann1974values}, its line integral is the unique fixed-path credit-allocation rule satisfying completeness (attributions sum to the prediction gap between $x_1$ and $x_0$), linearity (linear combinations of models yield linearly combined attributions), and dummy sensitivity (features the model never depends on receive zero credit)~\cite{sundararajan2017axiomatic,zhang2026otflowshap}. These guarantees are conditional on the chosen endpoints and trajectory, however: they ensure attributions are consistent \emph{along that line}, but say nothing about whether the line, or the reference it starts from, is a good choice.

To fix this, a wave of recent work replaces straight lines from black images with curved trajectories synthesized by deep generative models: shortest curved paths (Riemannian geodesics) and latent trajectories in Variational Autoencoders (VAEs)~\cite{zaher2024manifold,kim2026manifoldaligned}, classifier- or reward-steered reverse diffusion processes~\cite{lei2024denoising,kim2026diffusion}, probability-flow ODEs~\cite{song2021scorebased}, and optimal transport (OT) flow-based attributions~\cite{zhang2026otflowshap}. While these generative explainers report striking gains over black-baseline IG, they come at a steep price: training or numerically inverting (mapping $x_1$ back to a noise seed $x_0$ via) large generative networks to synthesize intermediate waypoints, and frequently coupling path generation to classifier or reward gradients in ways that break Aumann--Shapley axioms or optimize the test metric itself.

More fundamentally, replacing black-baseline IG with a generative flow or diffusion bridge alters \emph{two confounded factors at once} (Figure~\ref{fig:main_fig}a): (i)~\coupl{\textbf{reference selection}}, replacing the constant black image with a generator-inverted noise seed tailored to $x_1$ (which simultaneously shifts the baseline distribution $p_0$ from a single black point mass to full-support noise spanning the pixel domain, and pairs each input with an image-specific seed), and (ii)~\excurs{\textbf{path shape}}, bending the trajectory away from a \linpath{straight line} along a learned flow or diffusion path. Because these expensive methods alter both factors at once, the reported improvements remain fundamentally ambiguous: \emph{which of these changes actually drives the gain, and, consequently, do we need generative models at all for faithful explanations?}

We answer these questions through a theoretical and empirical reality check built on a simple \emph{baseline--path decomposition} (Figure~\ref{fig:main_fig}a). Writing any trajectory as a straight line between endpoints plus an off-line deviation separates any explainer into two components: (i)~the \coupl{\textbf{baseline distribution}} $\coupl{K(\cdot \mid x_1)}$ (which fixes both the overall baseline pool and how baselines are paired with inputs) and (ii)~the \excurs{\textbf{path deviation}} $\excurs{\xi_t}$ away from the straight line. Under this decomposition, recent generative explainers alter both components simultaneously relative to black-baseline IG, obscuring which change drives their gains.

Concretely, building on the axiomatic transport framework of \citet{zhang2026otflowshap}, we prove that under exact quadratic Optimal Transport in Euclidean space, the optimal-transport path between distributions (McCann's displacement interpolation~\cite{McCann1997ACP,villani2009optimal}) follows straight chords with \emph{zero path deviation}, collapsing transport-geodesic attribution to straight-line IG from the coupled reference (Proposition~\ref{prop:ot_collapse}). Optimal flow-matching IG explainers therefore collapse to \linpath{straight lines}, leaving \coupl{baseline selection} as the only remaining source of improvement over classical IG. Examining concurrent diffusion explainers reveals similar confounds: DDPath~\cite{lei2024denoising} averages raw gradients without weighting by step displacements and loses completeness, while DiffIG~\cite{kim2026diffusion} steers paths with a reward trained on its own test metric even though its own ablations show that multi-path schedule averaging drives its gains (and not diffusion guidance).

These insights motivate four falsifiable hypotheses (Section~\ref{sec:method_generative}), tested via controlled ablations across three datasets, two classifier backbones, and nine evaluation metrics (Section~\ref{sec:experiments}). Because both continuous-time flow matching~\cite{liu2022flow} and deterministic DDIM~\cite{song2021denoising} trace probability-flow ODE trajectories between noise and data~\cite{song2021scorebased,albergo2025stochastic}, we evaluate two path-diverse generative backends under a unified \textbf{Flow-IG} framing (following~\citet{zhang2026otflowshap}): \textbf{Flow-IG (PixelDiT)}, using the PixelDiT pixel-space flow-matching model~\cite{yu2026pixeldit} for straighter OT-style paths, and \textbf{Flow-IG (Diffusion)}, using OpenAI's unconditional diffusion model~\cite{dhariwal2021diffusion} trained with DDPM~\cite{ho2020denoising} and inverted via deterministic DDIM~\cite{song2021denoising} as a naturally curved-path alternative.
\begin{itemize}[leftmargin=*, align=left, labelsep=0.5em, itemsep=1.5pt, topsep=2pt]
    \item[\circnumA{A}] \textbf{Generative curved paths are unnecessary:} Flattening learned flow or diffusion trajectories to straight lines between the exact same endpoints preserves or significantly \emph{improves} attribution quality across faithfulness, localization, and class sensitivity.
    \item[\circnumB{B}] \textbf{Generative image-specific baseline pairing is unnecessary:} Randomly permuting inverted noise baselines across evaluation images preserves the baseline distribution while breaking any semantic pairing. Far from hurting, breaking the pairing consistently \emph{improves} perturbation faithfulness (Deletion AUC) across every dataset--backbone--generator setting, refuting the coupling hypothesis of~\citet{zhang2026otflowshap}.
    \item[\circnumC{C}] \textbf{Generator-free Gaussian Linear IG is competitive:} Replacing the entire generative pipeline with a single independent Gaussian noise baseline (clamped to the pixel range) along a straight line (\textbf{Gaussian Linear IG}) matches or outperforms both Flow-IG backends across all six dataset--backbone evaluations at zero generator cost (Figure~\ref{fig:main_fig}b).
    \item[\circnumD{D}] \textbf{Path design is overrated; baseline averaging dominates:} Under an identical compute budget, we compare averaging randomized coordinate schedules from a fixed Gaussian baseline via Stick-Breaking Path Integration~\cite{jeon2023beyond} (\textbf{Gaussian SPI}) against our \textbf{Expected Gaussian IG}, which averages straight lines over independent clamped Gaussian baselines. Averaging baselines decisively outperforms designing curved paths across all metrics, making Expected Gaussian IG our strongest method overall, ranking first among all $15$ evaluated methods on $7$ of $9$ metrics (and second on the remaining $2$).
\end{itemize}

\section{Related Work}
\label{sec:rel_work}

\paragraph{Axiomatic path attribution and heuristic baselines.}
Grounded in Aumann--Shapley game theory~\cite{aumann1974values}, Integrated Gradients (IG)~\cite{sundararajan2017axiomatic} uniquely satisfies completeness, linearity, symmetry, and dummy axioms along the straight line from a reference $x_0$ to the input $x_1$. Because a constant black baseline induces saturation and fails to credit dark pixels~\cite{sturmfels2020visualizing}, classical extensions either randomize the baseline over empirical references: marginalizing missing features over a background dataset in SHAP~\cite{lundberg2017shap} or averaging straight-line integrals over dataset references in Expected Gradients~\cite{erion2021improving}. Blur IG~\cite{xu2020attribution} integrates along a Gaussian scale-space path, Guided IG~\cite{kapishnikov2021guided} greedily advances low-gradient coordinates along a gradient-adaptive path, and Stick-Breaking Path Integration (SPI)~\cite{jeon2023beyond} averages attributions over non-adaptive, randomized Dirichlet coordinate schedules from a black baseline.

\paragraph{VAE manifold guided attribution.}
To avoid evaluating gradients on out-of-distribution linear interpolants, recent vision explainers route paths through deep generative models. Within Variational Autoencoders (VAEs), \textbf{Manifold Integrated Gradients (MIG)}~\cite{zaher2024manifold} integrates along shortest curved paths (geodesics) in the VAE's latent space between encoded endpoints, while \textbf{Manifold-Aligned Guided IG (MA-GIG)}~\cite{kim2026manifoldaligned} performs Guided IG~\cite{kapishnikov2021guided} in VAE latent space and decodes waypoints back to image space.

\paragraph{Optimal transport flows and diffusion flows.}
Using diffusion models, \textbf{Denoising Diffusion Path (DDPath)}~\cite{lei2024denoising} generates trajectories via classifier-guided reverse diffusion~\cite{dhariwal2021diffusion} from Gaussian noise without guaranteeing endpoint matching at $x_1$. \textbf{Diffusion Integrated Gradients (DiffIG)}~\cite{kim2026diffusion} trains a latent diffusion model over Stick-Breaking paths in a $\beta$-VAE and uses reward-gradient guidance during stochastic sampling to maximize a learned perturbation-faithfulness reward, introducing auxiliary training, hyperparameters, and axiom violations.
Parallel to this line of work, \textbf{OTFlowSHAP}~\cite{zhang2026otflowshap} separates fixed-path credit allocation from path selection and proposes integrating along Benamou--Brenier optimal transport (OT) geodesics~\cite{benamou2000computational}, approximated via Rectified Flow and Reflow~\cite{liu2022flow}, arguing that transport-consistent trajectories and coupled references yield more stable, spatially structured explanations with competitive faithfulness. Crucially, all of these generative methods require training generative models and evaluating large neural networks during inference to produce waypoints.

Building directly on \citet{zhang2026otflowshap}, we unify classical, schedule-based, and generative explainers via our \textit{baseline--path decomposition}. In Section~\ref{sec:method}, we use this formalism to isolate what generative pipelines actually change, examine the theoretical and estimator properties of recent flow and diffusion methods, and show that a simple generator-free Gaussian baseline captures their primary gains.

\section{Method}
\label{sec:method}
Given a trained image classifier and an input image $x_1$, our goal is to explain \emph{why} the model predicts a target class $c$ by assigning each input feature (pixel) $i$ an importance score $\Phi_i$ measuring its contribution to the class score $f_c(x_1)$. Because feature importance is only meaningful relative to what is ``missing,'' attribution methods compare $x_1$ against an uninformative reference image $x_0$ (such as an all-black canvas) and seek scores that are both \emph{faithful} (pinpointing features the classifier genuinely relies on rather than visual artifacts) and \emph{complete}, so that individual credits sum cleanly to the total prediction change $f_c(x_1) - f_c(x_0)$. Path-based explainers compute these scores by continuously transforming $x_0$ into $x_1$ along a path $\gamma(t)$ and accumulating the classifier's sensitivity along the way.

Formally, let $x_1 \in [-1, 1]^d \subset \mathbb{R}^d$ denote the observed input image, $x_0 \in \mathbb{R}^d$ a reference baseline, and $f_c : \mathbb{R}^d \to \mathbb{R}$ the classifier's pre-softmax logit for target class $c$. Given a piecewise-$C^1$ path $\gamma : [0, 1] \to \mathbb{R}^d$ connecting $\gamma(0) = x_0$ to $\gamma(1) = x_1$, axiomatic fixed-path attribution uniquely allocates credit via the Aumann--Shapley line integral~\cite{aumann1974values,sundararajan2017axiomatic,zhang2026otflowshap}:
\begin{equation}
    \Phi_i(f_c, \gamma) = \int_0^1 \frac{\partial f_c(\gamma(t))}{\partial x_i} \dot{\gamma}_i(t) \, dt.
    \label{eq:as_integral}
\end{equation}
We assume $f_c$ is continuously differentiable along the paths under consideration; the same calculation holds for piecewise-smooth networks along paths satisfying the almost-everywhere chain rule. By the fundamental theorem of calculus, summing across all $d$ coordinates yields \emph{completeness}:
\begin{equation}
    \sum_{i=1}^d \Phi_i(f_c, \gamma) = f_c(x_1) - f_c(x_0).
    \label{eq:completeness}
\end{equation}
Equation~\eqref{eq:completeness} exposes a fundamental separation: choosing the reference $x_0$ changes the total explained logit gap $f_c(x_1) - f_c(x_0)$~\cite{sturmfels2020visualizing}, whereas choosing the interior trajectory $\gamma(t)$ at fixed endpoints leaves the total explained score invariant~\cite{friedman2004paths,sundararajan2017axiomatic} and only reallocates credit across features.

\subsection{Baseline--Path Decomposition}
\label{sec:method_bridges}

Once fixed-path credit allocation is locked to Eq.~\eqref{eq:as_integral}, any path-based explainer is fully specified by how it selects baselines and connects them to $x_1$. We represent an explainer by a conditional baseline distribution $\coupl{K(\cdot \mid x_1)}$ and a path deviation $\excurs{\xi_t}$ around the straight line:
\begin{equation}
\begin{aligned}
    x_0 &\sim \coupl{K(\cdot \mid x_1)}, \qquad \excurs{\xi_0} = \excurs{\xi_1} = \mathbf{0}, \\
    \gamma(t) &= \linpath{\underbrace{(1 - t)x_0 + t x_1}_{\text{Linear chord } \gamma_{\text{lin}}(t)}} + \excurs{\underbrace{\xi_t(x_0, x_1)}_{\text{Path deviation}}}.
\end{aligned}
\label{eq:interpolant}
\end{equation}
For every continuous path with fixed endpoints, $\excurs{\xi_t} = \gamma(t) - \linpath{\gamma_{\text{lin}}(t)}$ is uniquely defined; its conditional law given $(x_0, x_1)$ captures any interior path guidance (e.g., classifier gradients $\nabla f_c$ or a generator's learned velocity or denoising field) and path randomness.

For inputs $x_1 \sim p_1$, the data distribution $p_1$ is fixed by the task, so choosing the conditional law $\coupl{K(\cdot \mid x_1)}$ alone determines both how baselines are paired with inputs (the joint coupling $\coupl{\pi}$) and what the overall baseline pool looks like (the reference marginal $p_0$, evaluated on any measurable set $A \subset \mathbb{R}^d$):
\begin{equation}
\begin{aligned}
    \coupl{\pi(dx_0, dx_1)} &= \coupl{K(dx_0 \mid x_1)}\, p_1(dx_1), \\
    p_0(A) &= \int \coupl{K(A \mid x_1)}\, p_1(dx_1).
\end{aligned}
\label{eq:coupling}
\end{equation}
Although $p_0$ is determined by $p_1$ and $\coupl{K}$ (or $\coupl{\pi}$), writing $p_0$ explicitly matches classical input-independent explainers ($\coupl{K(\cdot \mid x_1)} = p_0$, e.g., Vanilla IG~\cite{sundararajan2017axiomatic} and Expected Gradients~\cite{erion2021improving}) and lets our ablations separate shifts in the overall baseline pool ($p_0$) from per-image pairing ($\coupl{\pi}$) in inversion-based methods ($\coupl{K(\cdot \mid x_1)} = \delta_{h_\theta(x_1)}$, a point mass at the baseline $h_\theta(x_1)$ recovered by numerically inverting the generator on $x_1$ and mapping it back to classifier pixel space).

\paragraph{What can the path deviation change?}
With endpoints $(x_0, x_1)$ fixed, curving the path away from $\linpath{\gamma_{\text{lin}}}$ cannot change the total explained score; it can only redistribute credit across features through cross-feature interactions. Formally, let $\Delta_i(\excurs{\xi}) := \Phi_i(f_c, \linpath{\gamma_{\text{lin}}} + \excurs{\xi}) - \mathrm{IG}_i(f_c; x_0, x_1)$ denote the shift relative to straight-line $\mathrm{IG}_i$. By completeness~\eqref{eq:completeness}, $\sum_{i=1}^d \Delta_i(\excurs{\xi}) = 0$, so any path deviation is strictly zero-sum. Moreover, for an additive model $f_c(x) = b + \sum_i g_i(x_i)$, each partial derivative $\partial_i f_c(x) = g_i'(x_i)$ depends only on $x_i$, so every path yields the same attribution $g_i(x_{1,i}) - g_i(x_{0,i})$ and $\Delta_i(\excurs{\xi}) = 0$ for all $i$ and all $\excurs{\xi}$. Path dependence consequently arises solely from feature interactions ($\partial_{ij}^2 f_c \neq 0$)~\cite{ward2020exploration,janizek2021explaining}. When such interactions are present, path geometry governs which pixels can receive credit: coordinate-monotone paths (such as Guided IG~\cite{kapishnikov2021guided} and SPI~\cite{jeon2023beyond}) stay inside the coordinate box between $x_0$ and $x_1$ and therefore assign zero credit whenever $x_{1,i} = x_{0,i}$, whereas unconstrained curved paths (such as Blur IG~\cite{xu2020attribution} and generative flows~\cite{zhang2026otflowshap,kim2026diffusion}) can wander away from $x_{0,i}$ at intermediate $t$ and assign non-zero credit even to unchanged pixels. Appendix~\ref{app:path_dependence} and Table~\ref{tab:path_attribution_taxonomy} (Appendix~\ref{app:full_tables}) formalize this three-way geometric classification (straight/reparameterized chords, coordinate-monotone paths, and unconstrained curves).

\subsection{The Role of Generative Trajectories}
\label{sec:method_generative}

\paragraph{Optimal transport selects endpoint pairs}
Transport-geodesic attribution~\cite{zhang2026otflowshap} selects trajectories by minimizing the Benamou--Brenier kinetic action~\cite{benamou2000computational} $\int_0^1 \int_{\mathbb{R}^d} \|v_t(x)\|_2^2 \, d\rho_t(x)\, dt$ over time-dependent velocity fields $v_t(x)$ and interpolating densities $\rho_t$ satisfying the continuity equation $\partial_t \rho_t + \nabla \cdot (\rho_t v_t) = 0$ between $p_0$ and $p_1$, approximating this optimum via Rectified Flow and Reflow~\cite{liu2022flow}. In Euclidean space with quadratic cost, optimal displacement interpolation connects each optimally coupled pair by a straight segment~\cite{McCann1997ACP,villani2009optimal}. Applying Eq.~\eqref{eq:as_integral} yields an immediate collapse:

\begin{proposition}[Attribution along quadratic OT]
\label{prop:ot_collapse}
Let $p_0, p_1 \in \mathcal{P}_2(\mathbb{R}^d)$ (the space of probability measures on $\mathbb{R}^d$ with finite second moment) be absolutely continuous with respect to Lebesgue measure, and let $T^* : \mathbb{R}^d \to \mathbb{R}^d$ be their quadratic optimal transport map pushing $p_0$ to $p_1$ ($T^*_\# p_0 = p_1$). For $p_1$-almost every $x_1$, set $x_0^* = (T^*)^{-1}(x_1)$. Then the optimal displacement-interpolation trajectory has zero path deviation ($\excurs{\xi_t^*} \equiv \mathbf{0}$) and satisfies
\begin{equation}
\begin{aligned}
    \gamma^*(t) &= (1 - t)x_0^* + t x_1, \\
    \Phi_i(f_c, \gamma^*) &= \mathrm{IG}_i(f_c; x_0^*, x_1).
\end{aligned}
\label{eq:ot_collapse}
\end{equation}
\end{proposition}

The first identity is McCann's displacement interpolation~\cite{McCann1997ACP,villani2009optimal}, and substituting its constant velocity $\linpath{\dot{\gamma}^*(t)} = x_1 - x_0^*$ into Eq.~\eqref{eq:as_integral} gives the second (also noted in~\cite[App.~C.3]{zhang2026otflowshap}). Thus, exact quadratic OT contributes strictly through the \coupl{coupled reference} $\coupl{x_0^*}$, not through curved trajectories ($\excurs{\xi_t^*} \equiv \mathbf{0}$). Moreover, as proved in Appendix~\ref{app:energy_timing}, expected kinetic energy conflates endpoint pairing, off-chord deviation, and attribution-invariant reparameterization speed along the straight line. Minimizing kinetic action~\cite{zhang2026otflowshap} therefore has no formal link to classifier faithfulness, while zero path deviation is already achieved trivially by any \linpath{straight chord} $\linpath{\gamma_{\mathrm{lin}}}$.

\paragraph{Generative complexity lacks a clear justification.}
Just as quadratic OT collapses curved flows to straight chords, closer inspection of concurrent diffusion and flow explainers shows that their heavy generative machinery is neither theoretically sound nor empirically isolated as the driver of gains (detailed in Appendix~\ref{app:diffusion_estimators}). Rather than tracing a well-defined path integral from $x_0$ to $x_1$, DDPath~\cite{lei2024denoising}, for which no official code is released, combines internal formula mismatches with a non-standard, hard classifier-guided diffusion loop whose sampling mean is anchored directly to the target image $x_1$ rather than the noisy state; this forfeits any formal guarantee of ending at $x_1$, and its displayed algorithm drops step displacements $\Delta x_t$ altogether to return the unweighted gradient average $T^{-1}\sum_{t=0}^{T-1}\nabla f_c(x_t)$, losing completeness and behaving like noise-averaged saliency~\cite{smilkov2017smoothgrad}. DiffIG~\cite{kim2026diffusion} goes further by training a latent diffusion model alongside guidance predictors regressed on its own Insertion--Deletion test metric (\texttt{DiffID}), yet its own ablations retain $96\text{--}99\%$ of peak \texttt{DiffID} when guidance weights are zeroed and show guided \texttt{Best-of-}$N$ path selection trailing far behind multi-path averaging, indicating that simple stochastic schedule averaging as done by SPI \cite{jeon2023beyond}, not diffusion guidance, does the work. Meanwhile, diagnostic metrics such as Zhang et al.'s Flow Consistency Error~\cite{zhang2026otflowshap} only check how faithfully discrete steps follow the generator's velocity field, saying nothing about whether the resulting scores reflect the classifier. Across all three lines of work, expensive generative training and trajectory steering lack justification over much simpler controls, motivating the four falsifiable hypotheses tested next.

\paragraph{Falsifiable hypotheses.}
Proposition~\ref{prop:ot_collapse} and the diagnostics above isolate four hypotheses tested in Section~\ref{sec:experiments}:
\circnumA{A}~whether curved path deviations ($\excurs{\xi_t} \neq \mathbf{0}$) add value beyond \linpath{straight chords} ($\excurs{\xi_t^*} \equiv \mathbf{0}$);
\circnumB{B}~whether the empirical own-image inversion \coupl{coupling} $\coupl{\hat{\pi}_{h_\theta}}$ (induced by finite-step numerical inversion $h_\theta$) drives OT performance once $\excurs{\xi_t} \equiv \mathbf{0}$;
\circnumC{C}~whether sampling a single baseline directly from a clamped Gaussian prior along a \linpath{straight chord} suffices without any generator or \coupl{coupling}; and
\circnumD{D}~whether averaging straight chords over multiple independent Gaussian baselines (\textbf{Expected Gaussian IG}) outweighs averaging equally many Stick-Breaking path deviations~\cite{jeon2023beyond} from one Gaussian baseline (\textbf{Gaussian SPI}) at equal compute.

\subsection{Expected Gaussian Integrated Gradients}
\label{sec:method_direct}

\paragraph{Clamped Gaussian reference prior.}
Standard black baselines ($x_{\mathrm{blk}} = -\mathbf{1}$) suffer from \emph{baseline blindness} in dark image regions ($x_{1,i} - x_{\mathrm{blk},i} \approx 0$) and shift global image luminance strongly off-manifold. Motivated by Hypothesis~\circnumC{C}, that generative flows primarily benefit from replacing $-\mathbf{1}$ with a zero-mean, full-support noise prior, we eliminate the generator entirely and draw $Z \sim \mathcal{N}(0, I_d)$ independently of $x_1$, clamping to the pixel domain $[-1, 1]^d$:
\begin{equation}
    x_0 = \operatorname{clip}(Z, -1, 1), \qquad
    p_0^{\mathrm{G}} = \operatorname{Law}(x_0).
    \label{eq:gaussian_ref}
\end{equation}
Within our baseline--path decomposition (Eqs.~\eqref{eq:interpolant}--\eqref{eq:coupling}), this specifies input-independent baseline selection $\coupl{K(\cdot \mid x_1)} = p_0^{\mathrm{G}}$ and zero path deviation ($\excurs{\xi_t} \equiv \mathbf{0}$).

\paragraph{Expected Gaussian IG.}
Using this data-free Gaussian prior along the straight chord $\linpath{\gamma_{\mathrm{lin}}(t) = x_0 + t(x_1 - x_0)}$, we define \textbf{Expected Gaussian IG} and its $M$-sample Monte Carlo estimator ($x_0^{(1)}, \dots, x_0^{(M)} \stackrel{\text{i.i.d.}}{\sim} p_0^{\mathrm{G}}$, $M \ge 1$) as:
\begin{equation}
\begin{aligned}
    \Phi_i^{\mathrm{EG\text{-}G}}(f_c; x_1)
    &:= \mathbb{E}_{x_0 \sim p_0^{\mathrm{G}}}\!\left[\mathrm{IG}_i(f_c; x_0, x_1)\right] \\
    &\approx \frac{1}{M} \sum_{m=1}^M \mathrm{IG}_i\bigl(f_c; x_0^{(m)}, x_1\bigr),
    \label{eq:expected_gaussian_mc}
\end{aligned}
\end{equation}
requiring zero generator parameters, zero dataset access, and zero ODE inversions. Setting $M = 1$ yields \textbf{Gaussian Linear IG} ($G$), matching the $100$-interval gradient budget of single-path IG and Flow-IG (Hypothesis~\circnumC{C}), while $M = 16$ ($G_{16}$) reduces single-seed variance and probes feature interactions across independent random baselines without bending any trajectory. By linearity of expectation and Eq.~\eqref{eq:completeness}, Expected Gaussian IG preserves linearity, implementation invariance~\cite{sundararajan2017axiomatic}, dummy sensitivity, and expected completeness ($\sum_{i=1}^d \Phi_i^{\mathrm{EG\text{-}G}}(f_c; x_1) = f_c(x_1) - \mathbb{E}_{x_0 \sim p_0^{\mathrm{G}}}[f_c(x_0)]$, with sample-mean counterpart $f_c(x_1) - M^{-1}\sum_{m=1}^M f_c(x_0^{(m)})$). Unlike dataset-sampled Expected Gradients~\cite{erion2021improving} and SHAP~\cite{lundberg2017shap}, Expected Gaussian IG is completely data-free.

\paragraph{Gaussian SPI control (for Hypothesis~\circnumD{D}).}
To test whether extra compute is better spent averaging independent Gaussian baselines (Eq.~\eqref{eq:expected_gaussian_mc}) or averaging curved paths from a fixed baseline, \textbf{Gaussian SPI} ($S$) holds one draw $x_0 \sim p_0^{\mathrm{G}}$ fixed and averages $M$ coordinate-monotone deviations $\excurs{\xi_t^{\mathrm{SPI}}}$ sampled via randomized Dirichlet schedules following Stick-Breaking Path Integration~\cite{jeon2023beyond}, preserving completeness and the Zero-Difference Dummy property. Full schedule equations and budgets ($9 \times 11$ and $16 \times 100$) are in Appendices~\ref{app:intervention_sequence} and~\ref{app:explainers}.

\section{Experiments}
\label{sec:experiments}
To evaluate our baseline--path decomposition and test Hypotheses~\circnumA{A}, \circnumB{B}, \circnumC{C}, and~\circnumD{D}, we conduct a systematic empirical study across three vision benchmarks, two classifier architectures (six dataset--backbone settings), fifteen attribution methods, and nine evaluation metrics. Full preprocessing, hyperparameter, and protocol specifications are provided in Appendix~\ref{app:exp_details}.

\subsection{Experimental Setup}
\label{sec:exp_setup}

\paragraph{Datasets and classifier backbones.}
All inputs are evaluated in a unified $[-1, 1]^d$ RGB pixel space ($224 \times 224$) wrapped around each classifier's native ImageNet normalization, explaining the ground-truth target logit on stratified correctly predicted samples. We span three vision benchmarks:
(i)~\textbf{ImageNet}~\cite{deng2009imagenet} validation split ($n=1{,}000$; $1$ image/class across $1{,}000$ classes, with bounding-box masks),
(ii)~\textbf{Oxford-IIIT Pet}~\cite{parkhi2012pets} test split ($n=1{,}480$; $40$ images/class across $37$ breeds, with pixel segmentation masks), and
(iii)~\textbf{Oxford Flowers-102}~\cite{nilsback2008flowers} test split ($n=1{,}020$; $10$ images/class across $102$ classes, without localization masks).
We benchmark both a convolutional backbone (\textbf{ResNet-50}~\cite{he2016resnet}) and a hierarchical Vision Transformer (\textbf{PVT-v2-B2}~\cite{wang2022pvtv2}), using ImageNet-pretrained weights on ImageNet and fine-tuned checkpoints on Pets and Flowers-102 (Appendix~\ref{app:datasets_models}).

\paragraph{Evaluated explainers and generator-free baselines.}
We compare $15$ explainers across five families in native RGB space under matched nominal step budgets ($100$ steps/intervals where applicable):
(1)~\emph{Sanity checks} (ResNet-50 only): Deconvolution~\cite{zeiler2014deconv} and Guided Backprop~\cite{springenberg2014guided};
(2)~\emph{Non-integrated baselines}: Absolute and Signed Gradient~\cite{simonyan2013saliency}, Input$\times$Gradient~\cite{shrikumar2017inputxgrad}, and Guided Grad-CAM~\cite{selvaraju2017gradcam} (ResNet-50 only);
(3)~\emph{Classical and adaptive IG} (black baseline $x_{\mathrm{blk}} = -\mathbf{1}$ unless noted): Integrated Gradients~\cite{sundararajan2017axiomatic}, Expected Gradients~\cite{erion2021improving}, Guided IG~\cite{kapishnikov2021guided}, and Blur IG~\cite{xu2020attribution};
(4)~\emph{Generative Flow-IG baselines}: covering diverse path geometries, \textbf{Flow-IG (PixelDiT)} uses the unconditional $256\text{px}$ PixelDiT pixel-space flow-matching model~\cite{yu2026pixeldit,liu2022flow} for straighter OT-style paths, whereas \textbf{Flow-IG (Diffusion)} uses OpenAI's unconditional $256\text{px}$ diffusion model~\cite{dhariwal2021diffusion} trained with DDPM~\cite{ho2020denoising} and inverted via deterministic DDIM~\cite{song2021denoising} as a naturally curved-path alternative; because both generators operate at $256 \times 256$, waypoints are bicubic-downsampled to $224 \times 224$ classifier space and the $N$-th (final) waypoint is pinned to $x_N \equiv x_1$ to eliminate inversion reconstruction drift; and
(5)~\emph{Our generator-free methods and controls} (using budget notation $M \times N$ for $M$ references/paths of $N$ integration intervals): \textbf{Expected Gaussian IG}, evaluated both as a single-draw straight-chord control ($1 \times 100$, denoted $G$ or Gaussian Linear IG) and with $16$-reference baseline averaging ($16 \times 100$, $G_{16}$), alongside the \textbf{Gaussian SPI} path-design ablation ($S$) at compute-matched ($9 \times 11$) and full-precision ($16 \times 100$) budgets.
For full implementation details, see Appendix~\ref{app:explainers}.

\paragraph{Multi-axis evaluation metrics.}
A good attribution map should satisfy three distinct requirements: it should identify pixels the classifier actually relies on (\emph{faithfulness}), place mass on the object rather than background artifacts (\emph{localization}), and change when asked to explain a different class (\emph{class sensitivity}). Because any single metric alone can be misleading, for instance, an unsigned edge detector can score well on object localization without reflecting the model's decision, we evaluate nine metrics across all three axes (full definitions and hyperparameters in Appendix~\ref{app:metrics}):
\begin{itemize}
  \item \emph{Faithfulness:} Deletion AUC ($\downarrow$) and Insertion AUC ($\uparrow$) trace the target softmax probability across $100$-step spatial perturbation curves~\cite{petsiuk2018rise}, checking whether removing top-ranked pixels quickly destroys the prediction and inserting them quickly restores it. To test whether signed attribution magnitudes also track prediction changes proportionally, we include Faithfulness Correlation ($\mathrm{FC}\uparrow$~\cite{bhatt2020evaluating}, correlation between random-subset attribution sums and output drops), Monotonicity Correlation ($\mathrm{MC}\uparrow$~\cite{nguyen2020quantitative}), and Faithfulness Estimate ($\mathrm{FE}\uparrow$~\cite{alvarez2018robust}), which measure rank and linear correlation across progressively perturbed feature groups~\cite{hedstrom2023quantus}. All perturbation metrics replace masked pixels with the dataset training-split RGB mean, which is common practice in benchmarks, avoiding OOD artifacts \cite{hooker2019benchmark,carter2021overinterpretation}.
  \item \emph{Spatial localization} (ImageNet and Pets): Pointing Game ($\mathrm{PG}\uparrow$~\cite{zhang2018top}, whether the highest-scoring pixel falls inside the ground-truth mask), Top-$K$ Intersection ($\mathrm{TopK}\uparrow$~\cite{theiner2022interpretable}, mask overlap of the top-$K$ pixels), and Relevance Rank Accuracy ($\mathrm{RRA}\uparrow$~\cite{arras2022clevr}, fraction of top-$|\text{mask}|$ pixels inside the mask). These measure \emph{where} an explainer looks, not whether those pixels drive the logit.
  \item \emph{Class sensitivity:} RandomLogit ($\downarrow$)~\cite{pmlr-v119-sixt20a} computes the 1D SSIM between attributions for the target class~$c$ and a random off-target class~$c' \neq c$. Scores near $1.0$ mean the map barely changes with the explained class: the signature of a generic feature or edge detector rather than a class-specific explanation~\cite{adebayo2018sanity}.
\end{itemize}

\paragraph{Controlled ablation and paired-testing protocol.}
To test Hypotheses~\circnumA{A}--\circnumC{C} on both generative backends (\textbf{PixelDiT} and \textbf{Diffusion}), we compare four matched configurations: the full learned trajectory ($F$), its straight-line chord ($C$, keeping the same inverted baseline $\hat{x}_0 = h_\theta(x_1)$ and pairing $\coupl{\hat{\pi}_{h_\theta}}$ with $\excurs{\xi_t} \equiv \mathbf{0}$), a donor-permuted chord ($P$, shuffling inverted baselines across evaluation images without replacement so marginal distributions match while per-image pairing is broken), and generator-free \textbf{Gaussian Linear IG} ($G$, a single independent draw $x_0 \sim p_0^{\mathrm{G}}$ along a straight chord). For Hypothesis~\circnumD{D}, we compare two generator-free $16 \times 100$ configurations under $p_0^{\mathrm{G}}$: \textbf{Expected Gaussian IG} ($G_{16}$, averaging straight chords over $16$ independent baselines) versus \textbf{Gaussian SPI} ($S$, averaging $16$ Stick-Breaking paths from one fixed baseline).

Table~\ref{tab:main-benchmark-summary} reports equal-weight means across all six dataset--backbone settings (ImageNet and Oxford-Pets have ground-truth masks for localization, Flowers-102 does not, thus no localization metrics for this dataset). Table~\ref{tab:ablation-summary} summarizes sample-paired statistical tests across those same settings: for each $(\text{dataset} \times \text{backbone}, \text{metric})$ cell, we test the per-image score difference $\text{Left} - \text{Right}$ using a two-sided paired Student $t$-test with Holm step-down correction~\cite{holm1979simple} per contrast ($p_{\mathrm{Holm}} < 0.05$). In each $\mathrm{L}/\mathrm{ND}/\mathrm{R}$ triplet, $\mathrm{L}$ (\emph{Favors left}) and $\mathrm{R}$ (\emph{Favors right}) count cells with a significant difference in the metric's improvement direction, while $\mathrm{ND}$ (\emph{Not detected}) counts cells without a significant difference. Complete per-cell tables, the composite $P - F$ contrast, and full test statistics appear in Appendices~\ref{app:full_tables}--\ref{app:paired_tests}.

\begin{table*}[t]
  \centering
  \caption{Global benchmark summary across $3$ datasets $\times$ $2$ classifier backbones ($6$-cell equal-weight mean for faithfulness and class sensitivity; $4$-cell mean across ImageNet and Oxford Pets for localization). $\dagger$~Evaluated on the $3$ ResNet-50 cells only ($2$ for localization), as PVT-v2-B2 lacks ReLU backward hooks and a standard final convolutional layer; $\dagger$ rows are excluded from $6$-cell \textbf{best} and \underline{second-best} highlighting. Framed \textsf{RandomLogit} scores indicate class-insensitive failure. Per-cell tables ($n$ and $\pm\,\mathrm{std}$) appear in Appendix~\ref{app:full_tables}.}
  \label{tab:main-benchmark-summary}
  \small
  \FlowerTableLayout
  \resizebox{\textwidth}{!}{%
  \begin{tabular}{p{1.6em}p{1.2em} l *{5}{c}!{\textcolor{FlowerDivider}{\vrule width 0.25pt}}*{3}{c}!{\textcolor{FlowerDivider}{\vrule width 0.25pt}}c}
    \toprule
    \multicolumn{3}{l}{}
    & \multicolumn{5}{c}{\scriptsize\textsf{Faithfulness (6-cell mean)}}
    & \multicolumn{3}{c}{\scriptsize\textsf{Localization (4-cell mean)}}
    & \multicolumn{1}{c}{\scriptsize\textsf{Class sensitivity}} \\
    \cmidrule(lr){4-8}\cmidrule(lr){9-11}\cmidrule(lr){12-12}
    \FlowerHeaders
    \midrule
    \rowcolor{FlowerSanityTint}
     &  & \textcolor{FlowerSanityInk}{Deconvolution $\dagger$} & \FlowerScore{0.1579} & \FlowerScore{0.1172} & \FlowerScore{-0.0027} & \FlowerScore{0.1936} & \FlowerScore{0.2049} & \FlowerScore{0.7849} & \FlowerScore{0.6873} & \FlowerScore{0.5884} & \FlowerWarning{0.9999} \\
    \rowcolor{FlowerSanityTint}
    \FlowerSanityLabel &  & \textcolor{FlowerSanityInk}{Guided Backprop $\dagger$} & \FlowerScore{0.1708} & \FlowerScore{0.1544} & \FlowerScore{-0.0001} & \FlowerScore{0.2559} & \FlowerScore{0.1500} & \FlowerScore{0.8370} & \FlowerScore{0.8230} & \FlowerScore{0.6753} & \FlowerWarning{0.9796} \\
    \midrule
     &  & Absolute Gradient & \FlowerScore{0.2801} & \FlowerScore{0.3457} & \FlowerScore{0.0061} & \FlowerScore{0.2522} & \FlowerScore{0.1882} & \FlowerScore{0.6696} & \FlowerScore{0.6790} & \FlowerScore{0.6344} & \FlowerScore{0.5615} \\
     &  & Signed Gradient & \FlowerScore{0.2535} & \FlowerScore{0.2443} & \FlowerScore{-0.0082} & \FlowerScore{0.2522} & \FlowerScore{0.1882} & \FlowerScore{0.6781} & \FlowerScore{0.6788} & \FlowerScore{0.6135} & \FlowerScore{0.4835} \\
     &  & InputXGradient & \FlowerScore{0.1989} & \FlowerScore{0.2926} & \FlowerScore{0.0673} & \FlowerBest{0.3747} & \FlowerScore{0.2544} & \FlowerScore{0.6900} & \FlowerScore{0.6671} & \FlowerScore{0.6010} & \FlowerScore{0.6196} \\
     &  & Guided Grad-CAM $\dagger$ & \FlowerScore{0.2177} & \FlowerScore{0.2144} & \FlowerScore{-0.0019} & \FlowerScore{0.3591} & \FlowerScore{0.1681} & \FlowerScore{0.9086} & \FlowerScore{0.8601} & \FlowerScore{0.6834} & \FlowerScore{0.8600} \\
    \midrule
     &  & Integrated Gradients & \FlowerScore{0.2636} & \FlowerScore{0.2769} & \FlowerScore{0.0244} & \FlowerScore{0.2027} & \FlowerScore{0.1583} & \FlowerScore{0.7088} & \FlowerScore{0.6829} & \FlowerScore{0.6064} & \FlowerScore{0.6116} \\
     &  & Expected Gradients & \FlowerScore{0.1744} & \FlowerScore{0.3722} & \FlowerScore{0.0558} & \FlowerScore{0.2951} & \FlowerScore{0.3301} & \FlowerScore{0.7736} & \FlowerScore{0.7214} & \FlowerScore{0.6065} & \FlowerSecond{0.4775} \\
     &  & Guided IG & \FlowerScore{0.2266} & \FlowerScore{0.2613} & \FlowerScore{0.0621} & \FlowerScore{0.3017} & \FlowerScore{0.3067} & \FlowerScore{0.8197} & \FlowerScore{0.7995} & \FlowerBest{0.6623} & \FlowerScore{0.6119} \\
     &  & Blur IG & \FlowerScore{0.2028} & \FlowerScore{0.3022} & \FlowerScore{0.0198} & \FlowerScore{0.2198} & \FlowerScore{0.2150} & \FlowerScore{0.8404} & \FlowerScore{0.7228} & \FlowerScore{0.6026} & \FlowerScore{0.5300} \\
     &  & Flow-IG (PixelDiT) & \FlowerScore{0.1364} & \FlowerScore{0.3351} & \FlowerScore{0.0740} & \FlowerScore{0.2768} & \FlowerScore{0.3687} & \FlowerScore{0.7917} & \FlowerScore{0.7498} & \FlowerScore{0.6261} & \FlowerScore{0.5849} \\
     &  & Flow-IG (Diffusion) & \FlowerScore{0.1555} & \FlowerScore{0.3203} & \FlowerScore{0.0291} & \FlowerScore{0.2311} & \FlowerScore{0.3010} & \FlowerScore{0.7531} & \FlowerScore{0.7241} & \FlowerScore{0.6209} & \FlowerScore{0.6133} \\
    \cmidrule(l){2-12}
    \rowcolor{FlowerSPITint}
     &  & Gaussian SPI ($9\!\times\!11$) & \FlowerScore{0.1144} & \FlowerScore{0.3984} & \FlowerScore{0.0523} & \FlowerScore{0.2228} & \FlowerScore{0.3439} & \FlowerScore{0.7730} & \FlowerScore{0.7565} & \FlowerScore{0.6267} & \FlowerScore{0.6133} \\
    \rowcolor{FlowerSPITint}
     & \FlowerSPILabel & Gaussian SPI ($16\!\times\!100$) & \FlowerSecond{0.0989} & \FlowerSecond{0.4174} & \FlowerSecond{0.0781} & \FlowerScore{0.2300} & \FlowerSecond{0.4077} & \FlowerSecond{0.8427} & \FlowerSecond{0.8001} & \FlowerScore{0.6377} & \FlowerScore{0.5452} \\
    \cmidrule(l){2-12}
    \rowcolor{FlowerOursTint}
     &  & \textbf{Expected Gaussian IG} ($1\!\times\!100$) & \FlowerScore{0.1269} & \FlowerScore{0.3474} & \FlowerScore{0.0735} & \FlowerScore{0.2449} & \FlowerScore{0.3730} & \FlowerScore{0.8057} & \FlowerScore{0.7654} & \FlowerScore{0.6312} & \FlowerScore{0.5268} \\
    \rowcolor{FlowerOursTint}
    \FlowerIGLabel & \FlowerOursLabel & \textbf{Expected Gaussian IG} ($16\!\times\!100$) & \FlowerBest{0.0901} & \FlowerBest{0.4578} & \FlowerBest{0.2070} & \FlowerSecond{0.3583} & \FlowerBest{0.5503} & \FlowerBest{0.8513} & \FlowerBest{0.8011} & \FlowerSecond{0.6508} & \FlowerBest{0.4000} \\
    \bottomrule
  \end{tabular}%
  }
\end{table*}

\begin{table*}[t]
  \centering
  \caption{Paired hypothesis test outcomes across $6$ dataset--backbone cells ($4$ masked cells for localization). Triplets $\mathrm{L}/\mathrm{ND}/\mathrm{R}$ (\emph{Favors left} / \emph{Not detected} / \emph{Favors right}) count cell-level paired Student $t$-test outcomes at Holm-adjusted $p_{\mathrm{Holm}} < 0.05$ for Hypotheses~\circnumA{A} ($C - F$, $m=96$), \circnumB{B} ($P - C$, $m=96$), \circnumC{C} ($G - F$, $m=96$), and~\circnumD{D} ($G_{16} - S$, $m=48$), giving $48$ paired tests per row. \textbf{Bold} on $\mathrm{L}$ marks $\mathrm{L} > \mathrm{R}$; \underline{underline} on $\mathrm{R}$ marks $\mathrm{R} > \mathrm{L}$. Cell means and full paired statistics appear in Appendices~\ref{app:full_tables}--\ref{app:paired_tests}.}
  \label{tab:ablation-summary}
  \small
  \setlength{\tabcolsep}{3.5pt}
  \renewcommand{\arraystretch}{1.12}
  \resizebox{\textwidth}{!}{%
  \begin{tabular}{ll *{5}{c}!{\color{gray!50}\vrule width 0.25pt}*{3}{c}!{\color{gray!50}\vrule width 0.25pt}c!{\color{gray!50}\vrule width 0.25pt}c}
    \toprule
    & & \multicolumn{5}{c}{\scriptsize\textsf{Faithfulness (6 cells)}} & \multicolumn{3}{c}{\scriptsize\textsf{Localization (4 cells)}} & \multicolumn{1}{c}{\scriptsize\textsf{Sensitivity (6 cells)}} & \multicolumn{1}{c}{\scriptsize\textsf{48-Test Summary}} \\
    \cmidrule(lr){3-7}\cmidrule(lr){8-10}\cmidrule(lr){11-11}\cmidrule(lr){12-12}
    \textbf{Backend} & \textbf{Paired Contrast} & Del.\ AUC\,$\downarrow$ & Ins.\ AUC\,$\uparrow$ & FC\,$\uparrow$ & MC\,$\uparrow$ & FE\,$\uparrow$ & PG\,$\uparrow$ & TopK\,$\uparrow$ & RRA\,$\uparrow$ & RandomLogit\,$\downarrow$ & $\mathrm{L} / \mathrm{ND} / \mathrm{R}$ \\
    \midrule
    \multirow{3}{*}{\rotatebox[origin=c]{90}{\scriptsize\textsf{PixelDiT}}}
      & \circnumA{A}\ $C - F$ (Chord vs.\ Flow) & $0/4/\underline{2}$ & $0/4/\underline{2}$ & $0/5/\underline{1}$ & $0/0/\underline{6}$ & $0/6/0$ & $0/4/0$ & $\mathbf{2}/1/1$ & $\mathbf{2}/1/1$ & $\mathbf{6}/0/0$ & $10 / 25 / \underline{13}$ \\
      & \circnumB{B}\ $P - C$ (Permuted vs.\ Chord) & $\mathbf{6}/0/0$ & $\mathbf{3}/3/0$ & $0/5/\underline{1}$ & $\mathbf{1}/5/0$ & $1/4/1$ & $0/4/0$ & $0/4/0$ & $0/2/\underline{2}$ & $\mathbf{5}/1/0$ & $\mathbf{16} / 28 / 4$ \\
      & \circnumC{C}\ $G - F$ (Gaussian vs.\ Flow) & $\mathbf{4}/2/0$ & $\mathbf{6}/0/0$ & $0/5/\underline{1}$ & $0/2/\underline{4}$ & $\mathbf{2}/4/0$ & $0/4/0$ & $\mathbf{3}/1/0$ & $\mathbf{2}/2/0$ & $\mathbf{6}/0/0$ & $\mathbf{23} / 20 / 5$ \\
    \midrule
    \multirow{3}{*}{\rotatebox[origin=c]{90}{\scriptsize\textsf{Diffusion}}}
      & \circnumA{A}\ $C - F$ (Chord vs.\ Flow) & $\mathbf{6}/0/0$ & $\mathbf{6}/0/0$ & $\mathbf{6}/0/0$ & $1/4/1$ & $\mathbf{5}/1/0$ & $0/4/0$ & $\mathbf{2}/1/1$ & $\mathbf{2}/1/1$ & $\mathbf{5}/1/0$ & $\mathbf{33} / 12 / 3$ \\
      & \circnumB{B}\ $P - C$ (Permuted vs.\ Chord) & $\mathbf{6}/0/0$ & $\mathbf{3}/3/0$ & $0/5/\underline{1}$ & $\mathbf{2}/4/0$ & $1/4/1$ & $0/4/0$ & $\mathbf{1}/3/0$ & $0/3/\underline{1}$ & $\mathbf{4}/2/0$ & $\mathbf{17} / 28 / 3$ \\
      & \circnumC{C}\ $G - F$ (Gaussian vs.\ Flow) & $\mathbf{6}/0/0$ & $\mathbf{6}/0/0$ & $\mathbf{6}/0/0$ & $\mathbf{1}/5/0$ & $\mathbf{6}/0/0$ & $\mathbf{3}/1/0$ & $\mathbf{3}/1/0$ & $\mathbf{3}/1/0$ & $\mathbf{6}/0/0$ & $\mathbf{40} / 8 / 0$ \\
    \midrule
    \scriptsize\textsf{Gen.-free}
      & \circnumD{D}\ $G_{16} - S$ (Refs vs.\ SPI Paths) & $\mathbf{5}/1/0$ & $\mathbf{6}/0/0$ & $\mathbf{6}/0/0$ & $\mathbf{6}/0/0$ & $\mathbf{6}/0/0$ & $\mathbf{1}/3/0$ & $\mathbf{2}/1/1$ & $\mathbf{2}/2/0$ & $\mathbf{6}/0/0$ & $\mathbf{40} / 7 / 1$ \\
    \bottomrule
  \end{tabular}%
  }
\end{table*}

\subsection{Main Results and Hypothesis Evaluation}
\label{sec:exp_results}

\paragraph{Global benchmark and sanity checks (Table~\ref{tab:main-benchmark-summary}).}
Our sanity controls show why all three metric axes matter together. On ResNet-50, Deconvolution~$\dagger$ ($\mathrm{RandomLogit}=0.9999$) and Guided Backprop~$\dagger$ ($0.9796$) produce virtually identical maps regardless of the target logit~\cite{adebayo2018sanity}, and Guided Grad-CAM~$\dagger$ inherits much of this insensitivity ($0.8600$). Acting like class-agnostic edge and foreground detectors, these methods still score high on spatial localization ($\mathrm{PG}=0.785\text{--}0.909$, $\mathrm{TopK}=0.687\text{--}0.860$) while collapsing to near-zero Faithfulness Correlation ($\mathrm{FC}\approx 0$): their maps trace object contours without tracking what moves the classifier's prediction. Conversely, classical black-baseline IG is class-sensitive ($\mathrm{RandomLogit}=0.6116$) yet suffers from severe baseline blindness ($\mathrm{FC}=0.0244$, collapsing to $0.0013$ on ImageNet/PVT-v2-B2; Appendix Table~\ref{tab:imagenet-pvt-explainers}). Dataset-sampled Expected Gradients partially mitigates this, whereas \textbf{Expected Gaussian IG} ($16 \times 100$) wins across all three axes at once: best on $4/5$ faithfulness metrics, top overall localization ($\mathrm{PG}$ and $\mathrm{TopK}$), and lowest $\mathrm{RandomLogit}$ ($0.4000$) without a reference dataset or generator.

\paragraph{Hypothesis~\circnumA{A}: Generative path deviations are unnecessary ($C - F$, Table~\ref{tab:ablation-summary}).}
Flattening learned trajectories to straight chords ($\excurs{\xi_t} \equiv \mathbf{0}$) strongly improves \textbf{Diffusion} ($\mathbf{33}/12/3$, winning all $6/6$ cells on Del.\ AUC, Ins.\ AUC, and $\mathrm{FC}$). On \textbf{PixelDiT}, $35/48$ tests either detect no difference or favor the straight chord ($10/25/13$; $C$ wins $6/6$ on RandomLogit, while Flow-IG consistently leads only on $\mathrm{MC}$, $0/0/6$). Curved flow paths thus offer no consistent gain over straight chords.

\paragraph{Hypothesis~\circnumB{B}: Input-specific coupling is unnecessary ($P - C$, Table~\ref{tab:ablation-summary}).}
Permuting inverted baselines across images along straight chords outperforms own-image pairing in $\mathbf{33}/96$ paired tests vs.\ only $7/96$ favoring own-image coupling, and lowers Deletion AUC in \textbf{all $12/12$ settings}—directly refuting the coupling hypothesis of~\citet{zhang2026otflowshap}.

\paragraph{Hypothesis~\circnumC{C}: Generator-free Gaussian Linear IG matches or beats flows ($G - F$, Tables~\ref{tab:main-benchmark-summary}--\ref{tab:ablation-summary}).}
A single independent clamped Gaussian draw along a straight chord ($G$) attains the top single-path mean on $6/9$ metrics (Appendix Table~\ref{tab:ablation-means-summary}), dominates Diffusion ($\mathbf{40}/8/0$, zero losses), and outperforms PixelDiT ($\mathbf{23}/20/5$, winning $6/6$ on Ins.\ AUC and RandomLogit and $4/6$ on Del.\ AUC).

\paragraph{Hypothesis~\circnumD{D}: Baseline expectation outweighs path design ($G_{16} - S$, Tables~\ref{tab:main-benchmark-summary}--\ref{tab:ablation-summary}).}
At matched $16 \times 100$ compute, averaging straight chords over $16$ independent Gaussian baselines ($G_{16}$) beats averaging $16$ Stick-Breaking paths from one baseline ($S$) on \textbf{all $9/9$ macro-means} and $\mathbf{40}/48$ paired tests ($40/7/1$), ranking first overall among all $15$ methods on $7/9$ metrics (second on $\mathrm{MC}$ and $\mathrm{RRA}$). Per-cell tables and complete paired-test statistics appear in Appendices~\ref{app:full_tables}--\ref{app:paired_tests}.

\paragraph{Qualitative comparison.}
Appendix Figures~\ref{fig:qual_resnet_signed}--\ref{fig:qual_pvt_mag} visualize signed ($\sum_c A_c$) and magnitude ($\lvert\sum_c A_c\rvert$) attribution grids across six shared samples on both ResNet-50 and PVT-v2-B2. Directly echoing our quantitative results, Black IG exhibits severe baseline blindness on dark foregrounds such as \textit{ski mask}. Single-draw Gaussian IG ($1\!\times\!100$) closely matches Flow-IG (PixelDiT) without any generator, and Expected Gaussian IG ($16\!\times\!100$) visibly reduces background speckle while sharpening object focus across both backbones. Visualization details are listed in Appendix~\ref{app:qualitative_vis}.

\section{Conclusion}
\label{sec:conclusion}

Do feature attributions need generative flows? Across our theoretical analysis and controlled evaluations, the answer is no. By separating path-based explanation into \coupl{reference selection} and \excurs{trajectory geometry}, we show that much of the apparent appeal of recent flow- and diffusion-based explainers comes from conflating two distinct mechanisms: moving away from a degenerate all-black baseline, which changes the explained prediction gap itself, and curving the interior path, which only redistributes credit along a fixed gap and vanishes altogether under exact quadratic optimal transport. Once these factors are isolated, neither learned curved trajectories nor input-specific inversion pairings deliver consistent gains. We conclude that heavy generative pipelines were largely serving as expensive samplers for a zero-mean, full-support reference.

When that role is recognized, the practical remedy becomes strikingly simple. Replacing the generator with straight-line integration from an independent clamped Gaussian baseline matches flow-based explainers at zero generative overhead. Furthermore, averaging straight chords across multiple independent Gaussian draws (\textbf{Expected Gaussian IG}) resolves baseline blindness and outperforms all tested methods on multiple metrics, while preserving Aumann--Shapley axioms without reference datasets, ODE inversions, or metric-tuned path guidance.

Finally, generative models may still prove valuable when explanations genuinely require structured, semantic counterfactuals or non-Euclidean latent trajectories. But until new methods are sequentially ablated against matched straight-chord and independent-reference controls, apparent breakthroughs from generative complexity should be treated with healthy skepticism.

\paragraph{Scope and limitations.}
Our theoretical collapse result and empirical evaluation target pixel-space flow-matching and diffusion-based explainers. Latent-manifold methods that integrate along non-Euclidean VAE geodesics or adaptive latent paths such as MIG~\cite{zaher2024manifold} and MA-GIG~\cite{kim2026manifoldaligned} are not evaluated here. Isolating latent curvature, decoder artifacts, and reference-baseline shifts in VAE-based explainers via the same sequential ablation protocol remains an open direction for future work.

\section*{LLM Usage Statement}
The authors used LLM-based tools (Gemini, ChatGPT, and Grammarly) for copyediting, grammar correction, phrasing refinement, and iterative structural polishing of the manuscript text. In addition, the authors used Gemini and ChatGPT as agentic coding assistants throughout the software development cycle: the authors specified implementation requirements and extensions, used the agents to draft and refine code, and manually reviewed, modified, and verified the resulting implementations. No LLMs or generative AI tools were used for research ideation or formulating scientific claims. The authors reviewed all outputs and take full responsibility for all code, experiments, analyses, and text in this submission.

\section*{Acknowledgments}
This paper is supported by the DAAD programme Konrad Zuse Schools of Excellence in Artificial Intelligence, sponsored by the German Federal Ministry of Research, Technology and Space. The authors gratefully acknowledge the scientific support and HPC resources provided by the Erlangen National High Performance Computing Center (NHR@FAU) of Fried\-rich-Alex\-an\-der-Uni\-ver\-si\-tät Er\-lan\-gen-Nürn\-berg (FAU), as well as the computational resources of the Leibniz Supercomputing Centre (LRZ).

{
    \small
    \bibliographystyle{ieeenat_fullname}
    \bibliography{main}
}

\clearpage
\setcounter{page}{1}
\maketitlesupplementary

\section{Supporting analysis}

\subsection{Path dependence and interaction allocation}
\label{app:path_dependence}
For any two admissible paths with the same endpoints, subtracting Eq.~\eqref{eq:completeness} proves that the attribution difference sums to zero. For additive scores, the one-dimensional chain rule proves the coordinate identity stated in the main text. These are standard consequences of path integration; the additive result also appears in Zhang et al.~\cite[Prop.~B.3]{zhang2026otflowshap}.

The interaction interpretation can be made explicit for $f_c\in C^2$. Consider a smooth deformation $H(s,t)$ between two paths, with endpoints fixed for all $s$. Differentiation under the integral and integration by parts give
\begin{equation}
\begin{aligned}
&\frac{d}{ds}\Phi_i(f_c,H(s,\cdot)) \\
&\quad = \sum_{j\ne i}\int_0^1 \partial_{ij}f_c(H)\bigl(
\partial_sH_j\,\partial_tH_i
-\partial_tH_j\,\partial_sH_i\bigr)\,dt.
\end{aligned}
\label{eq:interaction}
\end{equation}
To obtain this identity, expand the derivative of $\partial_i f_c(H)\partial_tH_i$. Integrating its $\partial_{st}H_i$ term by parts produces a zero boundary term because $\partial_sH_i(s,0)=\partial_sH_i(s,1)=0$. The $j=i$ terms cancel. Thus changes in coordinate attribution are governed by mixed partial derivatives and the deformation of the path. Summing over $i$ cancels the remaining terms pairwise. On $\mathbb{R}^d$, vanishing mixed partials imply additive structure and path-independent coordinate attribution. For example, for $f(x) = x_1 x_2$ between $(0, 0)$ and $(1, 1)$, the straight chord assigns $(1/2, 1/2)$, whereas advancing $x_1$ before $x_2$ assigns $(0, 1)$.

\paragraph{Geometric classes of path deviations.}
Path-based explainers fall into three geometric classes (mapped in Table~\ref{tab:path_attribution_taxonomy}):
(1)~\emph{Straight and reparameterized chords} ($\excurs{\xi_t} \equiv \mathbf{0}$, or $\excurs{\xi_t} = (s(t)-t)(x_1 - x_0)$ for a smooth increasing bijection $s:[0,1]\to[0,1]$), which have $\Delta_i(\excurs{\xi}) = 0$ identically by reparameterization invariance;
(2)~\emph{Coordinate-monotone paths} ($\excurs{\xi_t} = (\lambda(t) - t\mathbf{1}) \odot (x_1 - x_0)$ with monotone schedules $\lambda_i(t) \in [0, 1]$), which stay inside $\prod_{i=1}^d [\min(x_{0,i}, x_{1,i}), \max(x_{0,i}, x_{1,i})]$ and preserve the zero-difference property ($x_{1,i} = x_{0,i} \Rightarrow \Phi_i = 0$), covering Guided IG~\cite{kapishnikov2021guided} and SPI~\cite{jeon2023beyond}; and
(3)~\emph{Unconstrained curved paths} (including Blur IG~\cite{xu2020attribution} and flow- or diffusion-based IG~\cite{zhang2026otflowshap,kim2026diffusion}), which can wander away from $x_{0,i}$ at $t \in (0, 1)$ and assign non-zero credit even when $x_{1,i} = x_{0,i}$.

\subsection{Transport energy and timing}
\label{app:energy_timing}
The constant-speed chord also provides a useful energy decomposition. Let $(X_0,X_1)\sim\coupl{\pi}$, let $D=X_1-X_0$, and suppose the random absolutely continuous path $\gamma(t)=\linpath{X_0+tD}+\excurs{\xi_t}$ has finite expected squared velocity, $\mathbb{E}\int_0^1\|\dot\gamma(t)\|^2\,dt < \infty$. Since $\excurs{\xi_0}=\excurs{\xi_1}=0$,
\begin{equation}
\int_0^1\langle D,\excurs{\dot\xi_t}\rangle\,dt
=\langle D,\excurs{\xi_1}-\excurs{\xi_0}\rangle=0.
\end{equation}
Expanding the squared velocity first gives $\mathbb{E}\int_0^1\|\dot\gamma(t)\|^2\,dt = \mathbb{E}_{\coupl{\pi}}\|D\|^2 + \mathbb{E}\int_0^1\|\excurs{\dot\xi_t}\|^2\,dt$, so both terms on the right are finite and $W_2^2(p_0,p_1) \le \mathbb{E}_{\coupl{\pi}}\|D\|^2 < \infty$. Adding and subtracting $W_2^2(p_0,p_1)$ therefore yields
\begin{equation}
\begin{aligned}
&\mathbb{E}\int_0^1\|\dot\gamma(t)\|^2\,dt \\
&\quad = W_2^2(p_0,p_1)
+\bigl(\mathbb{E}_{\coupl{\pi}}\|X_1-X_0\|^2-W_2^2(p_0,p_1)\bigr) \\
&\qquad +\mathbb{E}\int_0^1\|\excurs{\dot\xi_t}\|^2\,dt.
\end{aligned}
\label{eq:energy}
\end{equation}
The second term measures excess endpoint transport cost, and the third measures path-deviation velocity energy. The latter includes pure traversal timing: for any smooth increasing bijection $s:[0,1]\to[0,1]$ with $s(0)=0$ and $s(1)=1$, the collinearly reparameterized path $\gamma_s(t)=X_0+s(t)D$ (corresponding to $\excurs{\xi_t}=(s(t)-t)D$) has the exact same continuous attribution as the constant-speed chord by reparameterization invariance ($\Delta_i(\excurs{\xi}) = 0$), while its energy is
\begin{equation}
\begin{aligned}
\int_0^1\|\dot\gamma_s(t)\|^2\,dt
&= \|D\|^2\int_0^1 s'(t)^2\,dt \\
&= \|D\|^2\!\left(1 + \int_0^1 (s'(t)-1)^2\,dt\right) \ge \|D\|^2,
\end{aligned}
\end{equation}
which can be made arbitrarily large (e.g., via $s(t)=t^p$ as $p\to\infty$) without changing $\Phi_i(f_c,\gamma_s)$ at all.

Equation~\eqref{eq:energy} concerns particle energy. For the conditional-mean velocity $v_t(x)=\mathbb{E}[\dot\gamma(t)\mid\gamma(t)=x]$, Jensen's inequality gives
\begin{equation}
\int_0^1\mathbb{E}\|v_t(\gamma(t))\|^2\,dt
\leq \mathbb{E}\int_0^1\|\dot\gamma(t)\|^2\,dt,
\end{equation}
with equality when the velocity is determined by $(x,t)$, as for a deterministic ODE flow.

\paragraph{Consequences for kinetic-action diagnostics.}
Equation~\eqref{eq:energy} carries two immediate implications for transport-based attribution claims. First, because scalar kinetic action conflates coupling suboptimality (which alters the explained score gap via $x_0$), spatial path curvature (which reallocates interactions via $\partial_{ij}^2 f_c$), and pure traversal timing (which leaves continuous attributions invariant), observing lower action does not establish why—or even whether—continuous attributions changed. Second, while convergence toward a unique optimal coupling can improve cross-seed consistency, minimizing path-deviation velocity energy is not a property that requires a learned flow trajectory: for any endpoint pairing, the zero-deviation chord $\linpath{\gamma_{\text{lin}}}$ already achieves $\mathbb{E}\int_0^1\|\excurs{\dot\xi_t}\|^2\,dt = 0$ exactly.

\subsection{Interpreting published diffusion and flow estimators}
\label{app:diffusion_estimators}

\paragraph{DDPath.}
Lei et al.~\cite{lei2024denoising} do not release an official implementation (the NeurIPS checklist notes that code is withheld, requiring re-implementation in later work~\cite[App.~E]{kim2026diffusion}), and the published formulation leaves two foundational gaps:
\begin{enumerate}
    \item \emph{Non-standard diffusion and lost endpoint matching.} Standard classifier-guided reverse diffusion~\cite{dhariwal2021diffusion} conditions the denoiser mean $\mu_\theta(x_t, t)$ on the current noisy state $x_t$. DDPath~\cite[Def.~2, Sec.~4.1, Alg.~1]{lei2024denoising} instead anchors $\mu_\theta(x)$ and $\Sigma_\theta(x)$ directly to the target image $x$ and draws $x'_t \sim \mathcal{N}_t(\hat{\mu}_\theta(x), \Sigma_\theta(x))$ around $\hat{\mu}_\theta(x) = (1 - t/T)\,\mu_\theta(x) + (t/T)\,\Sigma\nabla_{x_t}\log p_\phi(y \mid x_t)$. Because sampling relies on stochastic draws steered by an increasingly strong hard classifier gradient with no boundary pinning at $x_1$, trajectories have no formal guarantee of reaching $x_1$ (even if target-centering promotes empirical convergence in practice), forfeiting exact endpoint matching and completeness.
    \item \emph{Omitted step displacements and broken completeness.} While Definition~2 writes the continuous path integral and Sec.~4.2 claims completeness, Eq.~(6) replaces the coordinate increments $\Delta x_t$ with $(t/T)x_t$, and Algorithm~1 drops displacements altogether to return the unweighted mean $T^{-1}\sum_{t=0}^{T-1}\nabla_{x_t} F(x_t)$---reducing the estimator to a noisy-trajectory analog of SmoothGrad~\cite{smilkov2017smoothgrad}. On the scalar linear score $f(x)=x$, Algorithm~1 returns $1$ for every sequence $(x_t)$, violating the required total attribution $x_1 - x_0$ (e.g., $1/2$ for $x_0=0, x_1=1/2$).
\end{enumerate}

\paragraph{DiffIG.}
DiffIG~\cite{kim2026diffusion} requires a pretrained $\beta$-VAE, an unconditional DiT1D diffusion model over Stick-Breaking Process (SBP) paths in latent space, and two DiT1D guidance predictors trained per $(\text{dataset}, \text{model})$ pair, followed by multi-path DDPM sampling~\cite[App.~G, Table~10]{kim2026diffusion}. Kim et al.'s own results point to three reasons this overhead is not isolated as the driver of faithfulness:
\begin{enumerate}
    \item \emph{Guidance trained on the evaluation objective.} The faithfulness regressor $\mathcal{J}_{\phi_1}$ is trained directly on $\mathbb{E}_{r\in\mathcal{R}}[\mathrm{Conf}^{\mathrm{ins}}(r) - \mathrm{Conf}^{\mathrm{del}}(r)]$~\cite[Sec.~3.3, Eq.~(15)]{kim2026diffusion}, the same Insertion-minus-Deletion confidence gap measured by the \texttt{DiffID} test metric (differing only in noisy-latent regression vs.\ pixel-space evaluation).
    \item \emph{Multi-path averaging, not guidance, drives performance.} Across all six $(\text{dataset}, \text{backbone})$ heatmaps in~\cite[App.~Q.3, Fig.~14]{kim2026diffusion}, zeroing both guidance weights ($\lambda_{\mathrm{faith}}=\lambda_{\mathrm{comp}}=0$) retains $96\text{--}99\%$ of peak \texttt{DiffID} (for instance, dropping only $0.94\%$ on Oxford-IIIT Pet/VGG16, from $0.639$ to $0.633$). By contrast, increasing sampled path counts yields substantial gains~\cite[App.~G Table~5, App.~Q.1 Fig.~13]{kim2026diffusion}, while \texttt{Best-of-}$N$ selection---which chooses the single path maximizing the learned guidance objective---consistently trails multi-path averaging by a wide margin~\cite[Fig.~6, App.~Q.2 Figs.~15--16]{kim2026diffusion}. Consequently, DiffIG's gains point to averaging multiple randomized SBP-like schedules rather than diffusion guidance, motivating our direct generator-free Gaussian SPI control (Hypothesis~\circnumD{D}).
    \item \emph{Independent-metric sensitivity.} On the independent Quantus Faithfulness Correlation metric (computed on absolute attributions for Oxford-IIIT Pet/ResNet18, $n=370$~\cite[App.~M]{kim2026diffusion}), DiffIG, AGI, and Guided IG all score near zero ($-0.0025$, $-0.0042$, and $-0.0093$). This motivates evaluating signed attributions across a broader suite of independent faithfulness, localization, and sensitivity criteria on our own datasets and backbones.
\end{enumerate}

\paragraph{Flow diagnostics.}
The Flow Consistency Error in Zhang et al.~\cite[App.~E.2]{zhang2026otflowshap} compares finite-difference waypoint velocities to the generator's learned vector field. As their revised paper explicitly notes, it is an internal dynamics diagnostic rather than a measure of classifier faithfulness; together with Eq.~\eqref{eq:energy} (Appendix~\ref{app:energy_timing}), this leaves the classifier-faithfulness case for curved flow trajectories unverified.

\subsection{Realizing the Controlled Ablation Configurations}
\label{app:intervention_sequence}
For evaluation inputs $x_1^{(j)}$, let $b_j=h_\theta(x_1^{(j)})$ be the actual classifier-space reference and let the corresponding generated waypoint path ($F$) be pinned to $b_j$ and $x_1^{(j)}$. The hypothesis tests compare the following controlled configurations:
\begin{description}
\item[Hypothesis~\circnumA{A} ($C$ vs.\ $F$).] Replace the generated path ($F$) with the straight chord ($C$) from $b_j$ to $x_1^{(j)}$. Both endpoints and their empirical coupling are retained.
\item[Hypothesis~\circnumB{B} ($P$ vs.\ $C$).] Draw a permutation $\sigma$ of the input indices and connect $b_{\sigma(j)}$ to $x_1^{(j)}$ along a straight chord ($P$). Both empirical marginals are retained and the assignment changes.
\item[Hypothesis~\circnumC{C} ($G$ vs.\ $F$).] Replace the generative pipeline ($F$) with the $M=1$ special case of Expected Gaussian IG (Eq.~\eqref{eq:expected_gaussian_mc}): a single independent clamped Gaussian draw $x_0^{(j)} = \operatorname{clip}(Z^{(j)}, -1, 1)$ ($Z^{(j)} \sim \mathcal{N}(0, I_d)$) connected to $x_1^{(j)}$ by a straight chord ($G$).
\item[Hypothesis~\circnumD{D} ($G_{16}$ vs.\ $S$).] Under the same clamped Gaussian reference law $p_0^{\mathrm{G}}$ and matched compute ($16 \times 100$), compare \textbf{Expected Gaussian IG} ($G_{16}$, averaging straight chords over $M=16$ i.i.d.\ draws $x_0^{(1)}, \dots, x_0^{(M)} \sim p_0^{\mathrm{G}}$ via Eq.~\eqref{eq:expected_gaussian_mc}) against \textbf{Gaussian SPI} ($S$), which holds a single draw $x_0 \sim p_0^{\mathrm{G}}$ fixed and averages $M$ finite-grid Stick-Breaking Path Integration (SPI)~\cite{jeon2023beyond} trajectories.
\end{description}

\paragraph{Finite-grid Gaussian SPI formulation.}
Along a single straight chord $\linpath{\gamma_{\mathrm{lin}}(t)}$, all features advance synchronously at the uniform rate $\lambda_i(t) = t$. When $f_c$ exhibits cross-feature interactions ($\partial_{ij}^2 f_c \neq 0$), however, coordinate credit depends on the relative order in which features transition from $x_0$ to $x_1$ (Appendix~\ref{app:path_dependence}). To isolate multi-path coordinate diversity at a fixed endpoint pair $(x_0, x_1)$ with $x_0 \sim p_0^{\mathrm{G}}$, each of $M$ SPI paths independently samples coordinate progress increments on a uniform $N$-step grid:
\begin{equation}
\begin{aligned}
    \bigl(u_{i,1}^{(m)}, \dots, u_{i,N}^{(m)}\bigr) &\sim \mathrm{Dirichlet}\!\left(\frac{\alpha}{N}, \dots, \frac{\alpha}{N}\right), \\
    \lambda_{i,k}^{(m)} &= \sum_{j=1}^k u_{i,j}^{(m)},
\end{aligned}
\label{eq:dirichlet}
\end{equation}
with $\lambda_{i,0}^{(m)} = 0$, and linearly interpolates $\lambda^{(m)}(t)$ between grid times $t_k = k/N$ so each trajectory forms a continuous, piecewise-linear box-constrained Hadamard path deviation $\excurs{\xi_t^{(m)}} = (\lambda^{(m)}(t) - t\mathbf{1}) \odot (x_1 - x_0)$:
\begin{equation}
\begin{aligned}
    \gamma^{(m)}(t) &= x_0 + \lambda^{(m)}(t) \odot (x_1 - x_0), \\
    \bar{\Phi}_i &= \frac{1}{M} \sum_{m=1}^M \Phi_i(f_c, \gamma^{(m)}).
\end{aligned}
\label{eq:gspi}
\end{equation}
Because different coordinates advance at different random rates within $[x_0, x_1]$, averaging over $M$ paths performs Monte Carlo integration over asynchronous feature arrival orders while strictly preserving completeness, linearity, and the Zero-Difference Dummy property ($x_{1,i} = x_{0,i} \Rightarrow \bar{\Phi}_i = 0$).

A realized permutation is a finite assignment without replacement. For a uniform permutation, its expected joint measure is the product of the empirical marginals, while any individual realization pairs each input with exactly one donor. Comparisons between Expected Gaussian IG ($G_{16}$) and Gaussian SPI ($S$) share the same reference-selection law $p_0^{\mathrm{G}}$ and total gradient budget ($16 \times 100$); for exact $1$-sample endpoint matching in single-baseline controls, the realized Gaussian reference sample is reused.

Finite-grid comparisons evaluate the combined effect of trajectory choice and its numerical approximation. Identical attribution rules and endpoint handling support interpretation, and residual checks quantify integration accuracy. Gradient counts, generator evaluations, and runtime characterize distinct costs.

\section{Extended Experimental Details}
\label{app:exp_details}

\subsection{Datasets, Preprocessing, and Classifier Backbones}
\label{app:datasets_models}

\paragraph{Pixel-space contract.}
Every image undergoes deterministic evaluation geometry consisting of an aspect-ratio-preserving bicubic resize of the shorter edge to $256$ pixels followed by a $224 \times 224$ center crop. Raw tensors in $[0, 1]^d$ are mapped linearly to the $[-1, 1]^d$ RGB pixel space ($d = 3 \times 224 \times 224 = 150{,}528$). Each classifier is wrapped internally by a normalization layer applying standard ImageNet statistics ($\mu = [0.485, 0.456, 0.406]$, $\sigma = [0.229, 0.224, 0.225]$), so all explainers operate directly in $[-1, 1]^d$ and classifier gradients backpropagate to pixel space via the affine chain rule. Attribution and metric evaluation run at sample-level batch size $1$.

\paragraph{Datasets and stratified sample manifests.}
For each dataset and backbone, we evaluate on deterministic, backbone-specific stratified manifests ($\mathrm{seed}=42$) of correctly predicted samples from the designated evaluation split, explaining the ground-truth target class pre-softmax logit $f_c(x_1)$:
\begin{itemize}
  \item \textbf{ImageNet (ILSVRC2012)}~\cite{deng2009imagenet}: official validation split ($50{,}000$ images; evaluation-only), stratified to $n = 1{,}000$ ($1$ image/class across $1{,}000$ classes). Ground-truth masks are rasterized at $224 \times 224$ from PASCAL-VOC-style XML bounding boxes using the identical resize/crop geometry ($n = 1{,}000$ masks).
  \item \textbf{Oxford-IIIT Pet}~\cite{parkhi2012pets}: \texttt{trainval} split deterministically partitioned $80/20$ ($\mathrm{seed}=42$) for training/validation and official \texttt{test} split for XAI evaluation, stratified to $n = 1{,}480$ ($40$ images/class across $37$ breeds). Pixel segmentation masks are available for $n = 1{,}475$ PVT-v2-B2 and $n = 1{,}477$ ResNet-50 samples (unannotated images are skipped per-sample in localization metrics).
  \item \textbf{Oxford Flowers-102}~\cite{nilsback2008flowers}: official \texttt{train}/\texttt{val}/\texttt{test} splits, stratified on \texttt{test} to $n = 1{,}020$ ($10$ images/class across $102$ classes). No localization masks are available (marked \textnormal{n/a}).
\end{itemize}

\paragraph{Classifier backbones and fine-tuning.}
We benchmark a convolutional network (\textbf{ResNet-50}~\cite{he2016resnet}) and a hierarchical Vision Transformer (\textbf{PVT-v2-B2}~\cite{wang2022pvtv2}). ImageNet uses standard ImageNet-1k pretrained weights without fine-tuning. On Oxford-IIIT Pet and Oxford Flowers-102, we replace the classification head and fine-tune from ImageNet-pretrained weights using AdamW, a cosine learning-rate schedule, differential learning rates (default backbone LR $10^{-4}$, head LR $10^{-3}$), and $5$ frozen-backbone warmup epochs followed by $45$ unfrozen epochs. Training uses a shared augmentation pipeline (\texttt{RandomResizedCrop(224)} with bicubic interpolation, random horizontal flip, and color jitter), while validation uses deterministic \texttt{Resize(256)+CenterCrop(224)} matching XAI evaluation; checkpoints are selected by validation Top-1 accuracy.

\subsection{Explainer Configurations and Compute Budgets}
\label{app:explainers}

All $15$ explainers return native $(1, 3, 224, 224)$ RGB attributions in $[-1, 1]^d$ space. Flow-IG (including its linear-chord, donor-permuted, and subdivided ablations), Expected Gaussian IG, and Gaussian SPI evaluate the coordinate-weighted trapezoidal rule $\sum_k \frac{1}{2}(\nabla f_c(x_{k-1}) + \nabla f_c(x_k)) \odot (x_k - x_{k-1})$ via a shared evaluator, and Blur IG applies the same trapezoidal formula directly across blurred waypoints; waypoint/sigma-step gradients are evaluated in chunks of at most $8$ states (\texttt{eval\_chunk\_size=8}) for Flow-IG and Blur IG. Classical non-trapezoidal baselines follow their standard reference formulations noted below.

\begin{enumerate}
  \item \emph{Sanity checks} (ResNet-50 only; marked \textnormal{n/a} on PVT-v2-B2 because its GELU activations violate the original ReLU backward-hook assumptions): \textbf{Deconvolution}~\cite{zeiler2014deconv} and \textbf{Guided Backprop}~\cite{springenberg2014guided}.
  \item \emph{Non-integrated baselines}: \textbf{Absolute Gradient} (Saliency)~\cite{simonyan2013saliency}, \textbf{Signed Gradient}, \textbf{Input$\times$Gradient}~\cite{shrikumar2017inputxgrad}, and \textbf{Guided Grad-CAM}~\cite{selvaraju2017gradcam} (ResNet-50 only).
  \item \emph{Classical and adaptive IG} (black baseline $x_{\mathrm{blk}} = -\mathbf{1}$ unless noted):
  \begin{itemize}
    \item \textbf{Integrated Gradients}~\cite{sundararajan2017axiomatic}: Captum implementation with $100$ Gauss--Legendre quadrature steps.
    \item \textbf{Expected Gradients}~\cite{erion2021improving}: Captum \texttt{GradientShap} Monte Carlo single-point sampling over dataset-sampled reference baselines.
    \item \textbf{Guided IG}~\cite{kapishnikov2021guided}: PyTorch port of PAIR-code's bounded $L_1$-budget adaptive path ($100$ steps, $\mathrm{fraction} = 0.25$, $\mathrm{max\_dist} = 0.02$, \texttt{float64} path bookkeeping), accumulating step-start gradients $(x_k - x_{k-1}) \odot \nabla f_c(x_{k-1})$.
    \item \textbf{Blur IG}~\cite{xu2020attribution}: $100$ steps from $\sigma_{\max} = 50$ down to $0$ (reusing the input when $\sigma < 0.01$), using explicit black ($-1$) convolution padding.
  \end{itemize}
  \item \emph{Generative Flow-IG baselines}: Both pixel-space generators operate natively at $256 \times 256$. Using the \texttt{crop\_upscale} delivery protocol, each $224 \times 224$ input is bicubic-upscaled to $256 \times 256$ and clamped to $[-1, 1]$, inverted/sampled in pixel space, bicubic-downsampled back to $224 \times 224$ and clamped to $[-1, 1]$, with the final waypoint pinned identically to the input ($x_N \equiv x_1$) to eliminate reconstruction drift. Generators and classifiers are strictly decoupled:
  \begin{itemize}
    \item \textbf{Flow-IG (PixelDiT)}: unconditional PixelDiT-ImageNet ($256\text{px}$, null class token $1000$)~\cite{yu2026pixeldit}, $100$-step Heun ODE inversion ($= 200$ generator NFE), yielding $101$ waypoints ($100$ intervals).
    \item \textbf{Flow-IG (Diffusion)}: OpenAI unconditional $256\text{px}$ diffusion checkpoint~\cite{dhariwal2021diffusion} (DDPM-trained~\cite{ho2020denoising}), $1000$-step linear schedule respaced uniformly to $100$ timesteps ($\eta = 0$ deterministic reverse DDIM~\cite{song2021denoising}, $\mathrm{clip\_denoised}=\mathrm{false}$, stopping at the last real timestep), yielding $100$ waypoints ($99$ intervals).
  \end{itemize}
  We note that these $100$-step presets match nominal temporal resolution rather than identical generator NFE ($200$ vs.\ $100$) or exact interval count ($100$ vs.\ $99$).
  \item \emph{Our generator-free method and SPI ablation}:
  \begin{itemize}
    \item \textbf{Expected Gaussian IG ($1 \times 100$ and $16 \times 100$)}: i.i.d.\ draws $Z^{(m)} \sim \mathcal{N}(0, I_d)$ ($\sigma = 1.0$) clamped via $x_0^{(m)} = \operatorname{clip}(Z^{(m)}, -1, 1)$ with zero path deviation ($\excurs{\xi_t} \equiv \mathbf{0}$), evaluated along straight chords of $100$ trapezoidal intervals per reference ($0$ generator evaluations). The $1 \times 100$ setting ($M=1$, $\mathrm{seed}=42$; $101$ classifier gradient passes, denoted \emph{Gaussian Linear IG} / $G$ in single-path ablations) matches the single-path gradient budget of Flow-IG, while $16 \times 100$ ($M=16$, $G_{16}$) evaluates multi-reference expectation.
    \item \textbf{Gaussian SPI ($9 \times 11$)} and \textbf{Gaussian SPI ($16 \times 100$)}: path-design ablation applying finite-grid Dirichlet coordinate-schedule path deviations (Eqs.~\eqref{eq:dirichlet}--\eqref{eq:gspi}, concentration $\alpha = 1.0$) over a single clamped Gaussian baseline ($\sigma = 1.0$) with independent per-sample/per-path CPU RNG streams. The $9 \times 11$ preset ($99$ total intervals / $108$ gradient passes) closely matches the classifier gradient budget of $100$-interval single-draw IG, while $16 \times 100$ ($S$) matches the compute budget of Expected Gaussian IG ($16 \times 100$).
  \end{itemize}
\end{enumerate}

\subsection{Evaluation Metrics and Perturbation Protocol}
\label{app:metrics}

All nine metrics are implemented GPU-natively in PyTorch (\texttt{fwd\_chunk\_size=128}), maintaining numerical parity with Quantus~0.6.0~\cite{hedstrom2023quantus} and RISE~\cite{petsiuk2018rise} protocols. Feature-coordinate metrics and RandomLogit consume native signed $(1, 3, 224, 224)$ RGB attributions, whereas spatial-ranking and localization metrics reduce RGB channels via the signed channel sum (row-major tie-breaking for AUC curves).

\paragraph{Perturbation replacement baseline.}
Perturbation-based faithfulness metrics replace removed/uninserted pixels across all three RGB channels simultaneously with the dataset training-split RGB mean converted to $[-1, 1]$: computed from the official training split for Flowers-102, the deterministic $\mathrm{seed}=42$ training split for Pets, and the standard ImageNet RGB mean $[0.485, 0.456, 0.406]$ for validation-only ImageNet. This avoids coincidence with the black baseline ($x_{\mathrm{blk}} = -\mathbf{1}$) used by classical IG explainers.

\paragraph{Metric definitions.}
\begin{itemize}
  \item \textbf{Deletion AUC ($\downarrow$) \& Insertion AUC ($\uparrow$)}: $100$-step cumulative spatial perturbation curves integrating target softmax probability (including $0\%$ and $100\%$ endpoints), using the dataset RGB mean for both deletion and insertion.
  \item \textbf{Faithfulness Correlation ($\mathrm{FC}\uparrow$)}: Pearson correlation between subset attribution sums and prediction drops across $100$ random subset perturbations of $7{,}526$ features ($\approx 5\%$ of $d$), evaluated on signed RGB attributions.
  \item \textbf{Monotonicity Correlation ($\mathrm{MC}\uparrow$) \& Faithfulness Estimate ($\mathrm{FE}\uparrow$)}: computed over $48$ sorted feature groups of size $3{,}136$. On Pets/PVT-v2-B2, $\mathrm{MC}$ has $n = 1{,}479$ ($n = 1{,}478$ for IG); on ImageNet/ResNet-50, Guided Grad-CAM has $n = 960$ for $\mathrm{FC}/\mathrm{MC}/\mathrm{FE}$. (Infidelity and MaxSensitivity are excluded from this benchmark.)
  \item \textbf{Pointing Game ($\mathrm{PG}\uparrow$), Top-$K$ Intersection ($\mathrm{TopK}\uparrow$, $k = 1{,}000$), \& Relevance Rank Accuracy ($\mathrm{RRA}\uparrow$)}: evaluated against ImageNet rasterized bounding boxes and Pets segmentation masks (NaN-skipped when a mask is absent; \textnormal{n/a} on Flowers-102).
  \item \textbf{RandomLogit ($\downarrow$)}: 1D SSIM between target-class $c$ attributions and attributions re-computed in memory for a uniformly sampled off-target class $c' \neq c$ (reusing target-independent Flow-IG waypoints where applicable). Scores near $1.0$ indicate class insensitivity.
\end{itemize}

\section{Complete Benchmark and Ablation Tables}
\label{app:full_tables}

Tables~\ref{tab:flowers-pvt-explainers}--\ref{tab:pets-resnet-explainers} report all six dataset--backbone cells of the $15$-explainer benchmark (summarized globally in Table~\ref{tab:main-benchmark-summary}). Table~\ref{tab:ablation-means-summary} summarizes the global equal-weight cell means across all controlled ablation configurations ($F, C, P, G, S, G_{16}$), complementing the paired hypothesis-test verdicts in Table~\ref{tab:ablation-summary}. Tables~\ref{tab:flowers-pvt-pixeldit-ablation}--\ref{tab:pets-resnet-pixeldit-ablation} and Tables~\ref{tab:flowers-pvt-diffusion-ablation}--\ref{tab:pets-resnet-diffusion-ablation} report the complete per-cell ablation results ($F, C, P, G$) for PixelDiT and Diffusion, respectively, and Table~\ref{tab:path_attribution_taxonomy} summarizes the unified known-endpoint bridge taxonomy. Section~\ref{app:paired_tests} then presents the complete paired statistical testing protocol and all $432$ cell-level contrast rows.

\begin{table*}[t]
  \centering
  \caption{Flowers102 attribution results for PVT-v2-B2. Sanity-check controls appear first, followed by non-integrated baselines and the IG family; Gaussian SPI ablations are shaded separately and \textbf{Expected Gaussian IG} forms a nested \textbf{Ours} block.}
  \label{tab:flowers-pvt-explainers}
  \small
  \FlowerTableLayout
  \resizebox{\textwidth}{!}{%
\begin{tabular}{p{1.6em}p{1.2em} l *{5}{c}!{\textcolor{FlowerDivider}{\vrule width 0.25pt}}*{3}{c}!{\textcolor{FlowerDivider}{\vrule width 0.25pt}}c}
    \toprule
    \FlowerGroupHeaders
    \FlowerHeaders
    \midrule
    \rowcolor{FlowerSanityTint}
     &  & \textcolor{FlowerSanityInk}{Deconvolution $\dagger$} & \FlowerUnavailable & \FlowerUnavailable & \FlowerUnavailable & \FlowerUnavailable & \FlowerUnavailable & \FlowerUnavailable & \FlowerUnavailable & \FlowerUnavailable & \FlowerUnavailable \\
    \rowcolor{FlowerSanityTint}
    \FlowerSanityLabel &  & \textcolor{FlowerSanityInk}{Guided Backprop $\dagger$} & \FlowerUnavailable & \FlowerUnavailable & \FlowerUnavailable & \FlowerUnavailable & \FlowerUnavailable & \FlowerUnavailable & \FlowerUnavailable & \FlowerUnavailable & \FlowerUnavailable \\
    \midrule
     &  & Absolute Gradient & \FlowerScore{0.4165} & \FlowerScore{0.3533} & \FlowerScore{0.0005} & \FlowerScore{0.2306} & \FlowerScore{0.1496} & \FlowerUnavailable & \FlowerUnavailable & \FlowerUnavailable & \FlowerScore{0.5569} \\
     &  & Signed Gradient & \FlowerScore{0.2784} & \FlowerScore{0.2866} & \FlowerScore{0.0085} & \FlowerScore{0.2306} & \FlowerScore{0.1496} & \FlowerUnavailable & \FlowerUnavailable & \FlowerUnavailable & \FlowerScore{0.4462} \\
     &  & InputXGradient & \FlowerScore{0.2217} & \FlowerScore{0.3754} & \FlowerScore{0.0936} & \FlowerScore{0.3214} & \FlowerScore{0.1923} & \FlowerUnavailable & \FlowerUnavailable & \FlowerUnavailable & \FlowerScore{0.5628} \\
     &  & Guided Grad-CAM & \FlowerUnavailable & \FlowerUnavailable & \FlowerUnavailable & \FlowerUnavailable & \FlowerUnavailable & \FlowerUnavailable & \FlowerUnavailable & \FlowerUnavailable & \FlowerUnavailable \\
    \midrule
     &  & Integrated Gradients & \FlowerScore{0.2724} & \FlowerScore{0.3215} & \FlowerScore{0.0495} & \FlowerScore{0.2928} & \FlowerScore{0.2923} & \FlowerUnavailable & \FlowerUnavailable & \FlowerUnavailable & \FlowerScore{0.5910} \\
\bottomrule
  \end{tabular}%
  }
\end{table*}

\begin{table*}[t]
  \centering
  \caption{Flowers102 attribution results for ResNet-50. Sanity-check controls appear first, followed by non-integrated baselines and the IG family; Gaussian SPI ablations are shaded separately and \textbf{Expected Gaussian IG} forms a nested \textbf{Ours} block.}
  \label{tab:flowers-resnet-explainers}
  \small
  \FlowerTableLayout
  \resizebox{\textwidth}{!}{%
  \begin{tabular}{p{1.6em}p{1.2em} l *{5}{c}!{\textcolor{FlowerDivider}{\vrule width 0.25pt}}*{3}{c}!{\textcolor{FlowerDivider}{\vrule width 0.25pt}}c}
    \toprule
    \FlowerGroupHeaders
    \FlowerHeaders
    \midrule
    \rowcolor{FlowerSanityTint}
     &  & \textcolor{FlowerSanityInk}{Deconvolution $\dagger$} & \FlowerScore{0.1511} & \FlowerScore{0.1180} & \FlowerScore{-0.0024} & \FlowerScore{0.2953} & \FlowerScore{0.1819} & \FlowerUnavailable & \FlowerUnavailable & \FlowerUnavailable & \FlowerWarning{0.9999} \\
    \rowcolor{FlowerSanityTint}
    \FlowerSanityLabel &  & \textcolor{FlowerSanityInk}{Guided Backprop $\dagger$} & \FlowerScore{0.1437} & \FlowerScore{0.1444} & \FlowerScore{0.0046} & \FlowerScore{0.2504} & \FlowerScore{0.0433} & \FlowerUnavailable & \FlowerUnavailable & \FlowerUnavailable & \FlowerWarning{0.9936} \\
    \midrule
     &  & Absolute Gradient & \FlowerScore{0.1210} & \FlowerBest{0.2210} & \FlowerScore{0.0095} & \FlowerScore{0.5030} & \FlowerScore{0.3393} & \FlowerUnavailable & \FlowerUnavailable & \FlowerUnavailable & \FlowerScore{0.5670} \\
     &  & Signed Gradient & \FlowerScore{0.0925} & \FlowerScore{0.0932} & \FlowerScore{0.0003} & \FlowerScore{0.5030} & \FlowerScore{0.3393} & \FlowerUnavailable & \FlowerUnavailable & \FlowerUnavailable & \FlowerScore{0.5383} \\
     &  & InputXGradient & \FlowerScore{0.0869} & \FlowerScore{0.0985} & \FlowerScore{0.0233} & \FlowerBest{0.6057} & \FlowerScore{0.4543} & \FlowerUnavailable & \FlowerUnavailable & \FlowerUnavailable & \FlowerScore{0.6458} \\
     &  & Guided Grad-CAM & \FlowerScore{0.1801} & \FlowerSecond{0.1834} & \FlowerScore{0.0047} & \FlowerScore{0.3618} & \FlowerScore{0.1058} & \FlowerUnavailable & \FlowerUnavailable & \FlowerUnavailable & \FlowerScore{0.8663} \\
\bottomrule
  \end{tabular}%
  }
\end{table*}

\begin{table*}[t]
  \centering
  \caption{ImageNet attribution results for PVT-v2-B2. Sanity-check controls appear first, followed by non-integrated baselines and the IG family; Gaussian SPI ablations are shaded separately and \textbf{Expected Gaussian IG} forms a nested \textbf{Ours} block.}
  \label{tab:imagenet-pvt-explainers}
  \small
  \FlowerTableLayout
  \resizebox{\textwidth}{!}{%
  \begin{tabular}{p{1.6em}p{1.2em} l *{5}{c}!{\textcolor{FlowerDivider}{\vrule width 0.25pt}}*{3}{c}!{\textcolor{FlowerDivider}{\vrule width 0.25pt}}c}
    \toprule
    \FlowerGroupHeaders
    \FlowerHeaders
    \midrule
    \rowcolor{FlowerSanityTint}
     &  & \textcolor{FlowerSanityInk}{Deconvolution $\dagger$} & \FlowerUnavailable & \FlowerUnavailable & \FlowerUnavailable & \FlowerUnavailable & \FlowerUnavailable & \FlowerUnavailable & \FlowerUnavailable & \FlowerUnavailable & \FlowerUnavailable \\
    \rowcolor{FlowerSanityTint}
    \FlowerSanityLabel &  & \textcolor{FlowerSanityInk}{Guided Backprop $\dagger$} & \FlowerUnavailable & \FlowerUnavailable & \FlowerUnavailable & \FlowerUnavailable & \FlowerUnavailable & \FlowerUnavailable & \FlowerUnavailable & \FlowerUnavailable & \FlowerUnavailable \\
    \midrule
     &  & Absolute Gradient & \FlowerScore{0.3534} & \FlowerScore{0.3051} & \FlowerScore{0.0010} & \FlowerScore{0.1496} & \FlowerScore{0.1248} & \FlowerScore{0.5950} & \FlowerScore{0.6143} & \FlowerScore{0.6318} & \FlowerScore{0.6741} \\
     &  & Signed Gradient & \FlowerScore{0.3399} & \FlowerScore{0.3223} & \FlowerScore{-0.0073} & \FlowerScore{0.1496} & \FlowerScore{0.1248} & \FlowerScore{0.6210} & \FlowerScore{0.6291} & \FlowerScore{0.6330} & \FlowerScore{0.6501} \\
     &  & InputXGradient & \FlowerScore{0.2689} & \FlowerScore{0.3953} & \FlowerScore{0.0545} & \FlowerSecond{0.3463} & \FlowerScore{0.1695} & \FlowerScore{0.6460} & \FlowerScore{0.6347} & \FlowerScore{0.6305} & \FlowerScore{0.7465} \\
     &  & Guided Grad-CAM & \FlowerUnavailable & \FlowerUnavailable & \FlowerUnavailable & \FlowerUnavailable & \FlowerUnavailable & \FlowerUnavailable & \FlowerUnavailable & \FlowerUnavailable & \FlowerUnavailable \\
    \midrule
     &  & Integrated Gradients & \FlowerScore{0.3440} & \FlowerScore{0.3677} & \FlowerScore{0.0013} & \FlowerScore{0.1107} & \FlowerScore{0.0229} & \FlowerScore{0.6430} & \FlowerScore{0.6396} & \FlowerScore{0.6350} & \FlowerScore{0.7640} \\
\bottomrule
  \end{tabular}%
  }
\end{table*}

\begin{table*}[t]
  \centering
  \caption{ImageNet attribution results for ResNet-50. Sanity-check controls appear first, followed by non-integrated baselines and the IG family; Gaussian SPI ablations are shaded separately and \textbf{Expected Gaussian IG} forms a nested \textbf{Ours} block.}
  \label{tab:imagenet-resnet-explainers}
  \small
  \FlowerTableLayout
  \resizebox{\textwidth}{!}{%
  \begin{tabular}{p{1.6em}p{1.2em} l *{5}{c}!{\textcolor{FlowerDivider}{\vrule width 0.25pt}}*{3}{c}!{\textcolor{FlowerDivider}{\vrule width 0.25pt}}c}
    \toprule
    \FlowerGroupHeaders
    \FlowerHeaders
    \midrule
    \rowcolor{FlowerSanityTint}
     &  & \textcolor{FlowerSanityInk}{Deconvolution $\dagger$} & \FlowerScore{0.1296} & \FlowerScore{0.0930} & \FlowerScore{-0.0011} & \FlowerScore{0.1035} & \FlowerScore{0.0311} & \FlowerScore{0.8250} & \FlowerScore{0.7423} & \FlowerScore{0.6500} & \FlowerWarning{1.0000} \\
    \rowcolor{FlowerSanityTint}
    \FlowerSanityLabel &  & \textcolor{FlowerSanityInk}{Guided Backprop $\dagger$} & \FlowerScore{0.1339} & \FlowerScore{0.1262} & \FlowerScore{-0.0012} & \FlowerSecond{0.2060} & \FlowerScore{0.1345} & \FlowerScore{0.8940} & \FlowerBest{0.8879} & \FlowerBest{0.7426} & \FlowerWarning{0.9550} \\
    \midrule
     &  & Absolute Gradient & \FlowerScore{0.1243} & \FlowerScore{0.1615} & \FlowerScore{0.0030} & \FlowerScore{0.1557} & \FlowerScore{0.0864} & \FlowerScore{0.7190} & \FlowerScore{0.7063} & \FlowerScore{0.6622} & \FlowerScore{0.4737} \\
     &  & Signed Gradient & \FlowerScore{0.1072} & \FlowerScore{0.1055} & \FlowerScore{-0.0099} & \FlowerScore{0.1557} & \FlowerScore{0.0864} & \FlowerScore{0.7140} & \FlowerScore{0.6955} & \FlowerScore{0.6463} & \FlowerScore{0.3805} \\
     &  & InputXGradient & \FlowerScore{0.0857} & \FlowerScore{0.1049} & \FlowerScore{0.0553} & \FlowerBest{0.2785} & \FlowerScore{0.1154} & \FlowerScore{0.7000} & \FlowerScore{0.6896} & \FlowerScore{0.6405} & \FlowerScore{0.5413} \\
     &  & Guided Grad-CAM & \FlowerScore{0.1914} & \FlowerScore{0.2015} & \FlowerScore{-0.0060} & \FlowerScore{0.2051} & \FlowerScore{0.1220} & \FlowerScore{0.8990} & \FlowerScore{0.8617} & \FlowerScore{0.6878} & \FlowerScore{0.8734} \\
\bottomrule
  \end{tabular}%
  }
\end{table*}

\begin{table*}[t]
  \centering
  \caption{Oxford Pets attribution results for PVT-v2-B2. Sanity-check controls appear first, followed by non-integrated baselines and the IG family; Gaussian SPI ablations are shaded separately and \textbf{Expected Gaussian IG} forms a nested \textbf{Ours} block.}
  \label{tab:pets-pvt-explainers}
  \small
  \FlowerTableLayout
  \resizebox{\textwidth}{!}{%
  \begin{tabular}{p{1.6em}p{1.2em} l *{5}{c}!{\textcolor{FlowerDivider}{\vrule width 0.25pt}}*{3}{c}!{\textcolor{FlowerDivider}{\vrule width 0.25pt}}c}
    \toprule
    \FlowerGroupHeaders
    \FlowerHeaders
    \midrule
    \rowcolor{FlowerSanityTint}
     &  & \textcolor{FlowerSanityInk}{Deconvolution $\dagger$} & \FlowerUnavailable & \FlowerUnavailable & \FlowerUnavailable & \FlowerUnavailable & \FlowerUnavailable & \FlowerUnavailable & \FlowerUnavailable & \FlowerUnavailable & \FlowerUnavailable \\
    \rowcolor{FlowerSanityTint}
    \FlowerSanityLabel &  & \textcolor{FlowerSanityInk}{Guided Backprop $\dagger$} & \FlowerUnavailable & \FlowerUnavailable & \FlowerUnavailable & \FlowerUnavailable & \FlowerUnavailable & \FlowerUnavailable & \FlowerUnavailable & \FlowerUnavailable & \FlowerUnavailable \\
    \midrule
     &  & Absolute Gradient & \FlowerScore{0.4683} & \FlowerScore{0.6369} & \FlowerScore{0.0122} & \FlowerScore{0.2258} & \FlowerScore{0.1552} & \FlowerScore{0.6366} & \FlowerScore{0.6491} & \FlowerSecond{0.5991} & \FlowerScore{0.5624} \\
     &  & Signed Gradient & \FlowerScore{0.5239} & \FlowerScore{0.4813} & \FlowerScore{-0.0301} & \FlowerScore{0.2258} & \FlowerScore{0.1552} & \FlowerScore{0.6793} & \FlowerScore{0.6784} & \FlowerScore{0.5808} & \FlowerSecond{0.4438} \\
     &  & InputXGradient & \FlowerScore{0.3808} & \FlowerScore{0.6100} & \FlowerSecond{0.1343} & \FlowerSecond{0.3262} & \FlowerScore{0.2200} & \FlowerScore{0.7064} & \FlowerScore{0.6546} & \FlowerScore{0.5562} & \FlowerScore{0.6169} \\
     &  & Guided Grad-CAM & \FlowerUnavailable & \FlowerUnavailable & \FlowerUnavailable & \FlowerUnavailable & \FlowerUnavailable & \FlowerUnavailable & \FlowerUnavailable & \FlowerUnavailable & \FlowerUnavailable \\
    \midrule
     &  & Integrated Gradients & \FlowerScore{0.5569} & \FlowerScore{0.5624} & \FlowerScore{0.0210} & \FlowerScore{0.0731} & \FlowerScore{0.0257} & \FlowerScore{0.6536} & \FlowerScore{0.6194} & \FlowerScore{0.5391} & \FlowerScore{0.6065} \\
\bottomrule
  \end{tabular}%
  }
\end{table*}

\begin{table*}[t]
  \centering
  \caption{Oxford Pets attribution results for ResNet-50. Sanity-check controls appear first, followed by non-integrated baselines and the IG family; Gaussian SPI ablations are shaded separately and \textbf{Expected Gaussian IG} forms a nested \textbf{Ours} block.}
  \label{tab:pets-resnet-explainers}
  \small
  \FlowerTableLayout
  \resizebox{\textwidth}{!}{%
  \begin{tabular}{p{1.6em}p{1.2em} l *{5}{c}!{\textcolor{FlowerDivider}{\vrule width 0.25pt}}*{3}{c}!{\textcolor{FlowerDivider}{\vrule width 0.25pt}}c}
    \toprule
    \FlowerGroupHeaders
    \FlowerHeaders
    \midrule
    \rowcolor{FlowerSanityTint}
     &  & \textcolor{FlowerSanityInk}{Deconvolution $\dagger$} & \FlowerScore{0.1931} & \FlowerScore{0.1405} & \FlowerScore{-0.0046} & \FlowerScore{0.1820} & \FlowerScore{0.4018} & \FlowerScore{0.7448} & \FlowerScore{0.6323} & \FlowerScore{0.5267} & \FlowerWarning{0.9999} \\
    \rowcolor{FlowerSanityTint}
    \FlowerSanityLabel &  & \textcolor{FlowerSanityInk}{Guided Backprop $\dagger$} & \FlowerScore{0.2349} & \FlowerScore{0.1926} & \FlowerScore{-0.0037} & \FlowerScore{0.3113} & \FlowerScore{0.2721} & \FlowerScore{0.7800} & \FlowerScore{0.7581} & \FlowerScore{0.6079} & \FlowerWarning{0.9902} \\
    \midrule
     &  & Absolute Gradient & \FlowerScore{0.1970} & \FlowerBest{0.3962} & \FlowerScore{0.0102} & \FlowerScore{0.2484} & \FlowerScore{0.2737} & \FlowerScore{0.7278} & \FlowerScore{0.7461} & \FlowerScore{0.6445} & \FlowerScore{0.5350} \\
     &  & Signed Gradient & \FlowerScore{0.1793} & \FlowerScore{0.1766} & \FlowerScore{-0.0105} & \FlowerScore{0.2484} & \FlowerScore{0.2737} & \FlowerScore{0.6980} & \FlowerScore{0.7123} & \FlowerScore{0.5937} & \FlowerBest{0.4418} \\
     &  & InputXGradient & \FlowerScore{0.1493} & \FlowerScore{0.1716} & \FlowerScore{0.0430} & \FlowerSecond{0.3701} & \FlowerScore{0.3748} & \FlowerScore{0.7075} & \FlowerScore{0.6896} & \FlowerScore{0.5767} & \FlowerScore{0.6044} \\
     &  & Guided Grad-CAM & \FlowerScore{0.2816} & \FlowerScore{0.2583} & \FlowerScore{-0.0045} & \FlowerBest{0.5104} & \FlowerScore{0.2765} & \FlowerBest{0.9181} & \FlowerBest{0.8585} & \FlowerBest{0.6789} & \FlowerScore{0.8403} \\
\bottomrule
  \end{tabular}%
  }
\end{table*}

\clearpage

\begin{table*}[t]
  \centering
  \caption{Global equal-weight cell means across controlled ablation configurations over all $6$ dataset--backbone cells ($4$ annotated cells for localization), complementing the paired hypothesis-test verdicts in Table~\ref{tab:ablation-summary}. \textbf{Top block:} $100$-interval single-path methods ($F, C, P, G$, with best \textbf{bold} and second-best \underline{underlined} within that budget). \textbf{Bottom block:} matched $16\!\times\!100$ generator-free scaling comparison ($S$ vs.\ $G_{16}$, best \textbf{bold}). Complete per-cell ablation results appear in Tables~\ref{tab:flowers-pvt-pixeldit-ablation}--\ref{tab:pets-resnet-diffusion-ablation}.}
  \label{tab:ablation-means-summary}
  \small
  \setlength{\tabcolsep}{3.5pt}
  \renewcommand{\arraystretch}{1.12}
  \resizebox{\textwidth}{!}{%
\begin{tabular}{ll *{5}{c}!{\color{gray!50}\vrule width 0.25pt}*{3}{c}!{\color{gray!50}\vrule width 0.25pt}c}
    \toprule
    & & \multicolumn{5}{c}{\scriptsize\textsf{Faithfulness (6 cells)}} & \multicolumn{3}{c}{\scriptsize\textsf{Localization (4 cells)}} & \multicolumn{1}{c}{\scriptsize\textsf{Sensitivity (6 cells)}} \\
    \cmidrule(lr){3-7}\cmidrule(lr){8-10}\cmidrule(lr){11-11}
    \textbf{Backend} & \textbf{Method} & Del.\ AUC\,$\downarrow$ & Ins.\ AUC\,$\uparrow$ & FC\,$\uparrow$ & MC\,$\uparrow$ & FE\,$\uparrow$ & PG\,$\uparrow$ & TopK\,$\uparrow$ & RRA\,$\uparrow$ & RandomLogit\,$\downarrow$ \\
    \midrule
    \multirow{3}{*}{\rotatebox[origin=c]{90}{\scriptsize\textsf{PixelDiT}}}
      & $F$: Flow-IG (PixelDiT) & 0.1364 & 0.3351 & \textbf{0.0740} & \textbf{0.2768} & 0.3687 & 0.7917 & 0.7498 & 0.6261 & 0.5849 \\
      & $C$: Linear Chord ($\excurs{\xi_t} \equiv \mathbf{0}$) & 0.1370 & 0.3328 & 0.0683 & 0.2414 & \underline{0.3709} & 0.7944 & 0.7513 & \underline{0.6272} & 0.5463 \\
      & $P$: Permuted Baseline ($\coupl{\hat{\pi}_{\mathrm{perm}}}$) & \textbf{0.1253} & 0.3415 & 0.0676 & 0.2508 & 0.3676 & \underline{0.7956} & \underline{0.7545} & 0.6249 & \underline{0.5291} \\
    \midrule
    \multirow{3}{*}{\rotatebox[origin=c]{90}{\scriptsize\textsf{Diffusion}}}
      & $F$: Flow-IG (Diffusion) & 0.1555 & 0.3203 & 0.0291 & 0.2311 & 0.3010 & 0.7531 & 0.7241 & 0.6209 & 0.6133 \\
      & $C$: Linear Chord ($\excurs{\xi_t} \equiv \mathbf{0}$) & 0.1367 & 0.3335 & 0.0669 & 0.2296 & 0.3483 & 0.7563 & 0.7277 & 0.6233 & 0.5722 \\
      & $P$: Permuted Baseline ($\coupl{\hat{\pi}_{\mathrm{perm}}}$) & \underline{0.1265} & \underline{0.3424} & 0.0663 & \underline{0.2518} & 0.3542 & 0.7763 & 0.7416 & 0.6228 & 0.5494 \\
\bottomrule
  \end{tabular}%
  }
\end{table*}

\begin{table*}[t]
  \centering
  \caption{Flowers102 attribution ablation for PVT-v2-B2: Flow-IG (PixelDiT), its Linear Chord and Permuted Baseline controls, and Expected Gaussian IG ($1\!\times\!100$).}
  \label{tab:flowers-pvt-pixeldit-ablation}
  \small
  \FlowerTableLayout
  \resizebox{\textwidth}{!}{%
  \begin{tabular}{p{1.6em}p{1.2em} l *{5}{c}!{\textcolor{FlowerDivider}{\vrule width 0.25pt}}*{3}{c}!{\textcolor{FlowerDivider}{\vrule width 0.25pt}}c}
    \toprule
    \FlowerGroupHeaders
    \FlowerHeaders
    \midrule
     &  & Flow-IG (PixelDiT) & \FlowerScore{0.1343} & \FlowerScore{0.4393} & \FlowerBest{0.1173} & \FlowerBest{0.2649} & \FlowerSecond{0.4273} & \FlowerUnavailable & \FlowerUnavailable & \FlowerUnavailable & \FlowerScore{0.4431} \\
     &  & \quad Linear Chord & \FlowerScore{0.1403} & \FlowerScore{0.4314} & \FlowerSecond{0.1134} & \FlowerSecond{0.2394} & \FlowerBest{0.4337} & \FlowerUnavailable & \FlowerUnavailable & \FlowerUnavailable & \FlowerScore{0.4169} \\
     &  & \quad Permuted Baseline & \FlowerSecond{0.1308} & \FlowerSecond{0.4446} & \FlowerScore{0.0984} & \FlowerScore{0.2188} & \FlowerScore{0.3744} & \FlowerUnavailable & \FlowerUnavailable & \FlowerUnavailable & \FlowerBest{0.3888} \\
    \cmidrule(l){2-12}
    \rowcolor{FlowerOursTint}
    \FlowerAblationIGLabel & \FlowerAblationOursLabel & \textbf{Expected Gaussian IG} ($1\!\times\!100$) & \FlowerBest{0.1303} & \FlowerBest{0.4517} & \FlowerScore{0.1018} & \FlowerScore{0.2277} & \FlowerScore{0.4073} & \FlowerUnavailable & \FlowerUnavailable & \FlowerUnavailable & \FlowerSecond{0.3987} \\
    \bottomrule
  \end{tabular}%
  }
\end{table*}

\begin{table*}[t]
  \centering
  \caption{Flowers102 attribution ablation for ResNet-50: Flow-IG (PixelDiT), its Linear Chord and Permuted Baseline controls, and Expected Gaussian IG ($1\!\times\!100$).}
  \label{tab:flowers-resnet-pixeldit-ablation}
  \small
  \FlowerTableLayout
  \resizebox{\textwidth}{!}{%
  \begin{tabular}{p{1.6em}p{1.2em} l *{5}{c}!{\textcolor{FlowerDivider}{\vrule width 0.25pt}}*{3}{c}!{\textcolor{FlowerDivider}{\vrule width 0.25pt}}c}
    \toprule
    \FlowerGroupHeaders
    \FlowerHeaders
    \midrule
     &  & Flow-IG (PixelDiT) & \FlowerScore{0.0675} & \FlowerSecond{0.1145} & \FlowerBest{0.0476} & \FlowerBest{0.4721} & \FlowerScore{0.4058} & \FlowerUnavailable & \FlowerUnavailable & \FlowerUnavailable & \FlowerScore{0.6967} \\
     &  & \quad Linear Chord & \FlowerScore{0.0710} & \FlowerScore{0.1144} & \FlowerScore{0.0405} & \FlowerScore{0.4267} & \FlowerScore{0.4048} & \FlowerUnavailable & \FlowerUnavailable & \FlowerUnavailable & \FlowerScore{0.6563} \\
     &  & \quad Permuted Baseline & \FlowerSecond{0.0646} & \FlowerSecond{0.1145} & \FlowerScore{0.0395} & \FlowerSecond{0.4476} & \FlowerBest{0.4277} & \FlowerUnavailable & \FlowerUnavailable & \FlowerUnavailable & \FlowerSecond{0.6240} \\
    \cmidrule(l){2-12}
    \rowcolor{FlowerOursTint}
    \FlowerAblationIGLabel & \FlowerAblationOursLabel & \textbf{Expected Gaussian IG} ($1\!\times\!100$) & \FlowerBest{0.0637} & \FlowerBest{0.1171} & \FlowerSecond{0.0408} & \FlowerScore{0.4024} & \FlowerSecond{0.4067} & \FlowerUnavailable & \FlowerUnavailable & \FlowerUnavailable & \FlowerBest{0.6127} \\
    \bottomrule
  \end{tabular}%
  }
\end{table*}

\begin{table*}[t]
  \centering
  \caption{ImageNet attribution ablation for PVT-v2-B2: Flow-IG (PixelDiT), its Linear Chord and Permuted Baseline controls, and Expected Gaussian IG ($1\!\times\!100$).}
  \label{tab:imagenet-pvt-pixeldit-ablation}
  \small
  \FlowerTableLayout
  \resizebox{\textwidth}{!}{%
  \begin{tabular}{p{1.6em}p{1.2em} l *{5}{c}!{\textcolor{FlowerDivider}{\vrule width 0.25pt}}*{3}{c}!{\textcolor{FlowerDivider}{\vrule width 0.25pt}}c}
    \toprule
    \FlowerGroupHeaders
    \FlowerHeaders
    \midrule
     &  & Flow-IG (PixelDiT) & \FlowerScore{0.1857} & \FlowerScore{0.4467} & \FlowerSecond{0.0530} & \FlowerBest{0.2272} & \FlowerScore{0.3570} & \FlowerBest{0.7330} & \FlowerBest{0.7358} & \FlowerSecond{0.6684} & \FlowerScore{0.6210} \\
     &  & \quad Linear Chord & \FlowerScore{0.1825} & \FlowerScore{0.4489} & \FlowerScore{0.0476} & \FlowerScore{0.1889} & \FlowerScore{0.3483} & \FlowerSecond{0.7140} & \FlowerScore{0.7170} & \FlowerScore{0.6638} & \FlowerSecond{0.6003} \\
     &  & \quad Permuted Baseline & \FlowerSecond{0.1715} & \FlowerSecond{0.4502} & \FlowerScore{0.0463} & \FlowerScore{0.2083} & \FlowerSecond{0.3600} & \FlowerScore{0.7120} & \FlowerScore{0.7294} & \FlowerScore{0.6651} & \FlowerBest{0.5908} \\
    \cmidrule(l){2-12}
    \rowcolor{FlowerOursTint}
    \FlowerAblationIGLabel & \FlowerAblationOursLabel & \textbf{Expected Gaussian IG} ($1\!\times\!100$) & \FlowerBest{0.1670} & \FlowerBest{0.4626} & \FlowerBest{0.0628} & \FlowerSecond{0.2142} & \FlowerBest{0.3976} & \FlowerSecond{0.7140} & \FlowerSecond{0.7323} & \FlowerBest{0.6703} & \FlowerScore{0.6025} \\
    \bottomrule
  \end{tabular}%
  }
\end{table*}

\begin{table*}[t]
  \centering
  \caption{ImageNet attribution ablation for ResNet-50: Flow-IG (PixelDiT), its Linear Chord and Permuted Baseline controls, and Expected Gaussian IG ($1\!\times\!100$).}
  \label{tab:imagenet-resnet-pixeldit-ablation}
  \small
  \FlowerTableLayout
  \resizebox{\textwidth}{!}{%
  \begin{tabular}{p{1.6em}p{1.2em} l *{5}{c}!{\textcolor{FlowerDivider}{\vrule width 0.25pt}}*{3}{c}!{\textcolor{FlowerDivider}{\vrule width 0.25pt}}c}
    \toprule
    \FlowerGroupHeaders
    \FlowerHeaders
    \midrule
     &  & Flow-IG (PixelDiT) & \FlowerScore{0.0669} & \FlowerScore{0.1492} & \FlowerSecond{0.0499} & \FlowerBest{0.1937} & \FlowerScore{0.1563} & \FlowerScore{0.8830} & \FlowerScore{0.8341} & \FlowerSecond{0.6933} & \FlowerScore{0.5453} \\
     &  & \quad Linear Chord & \FlowerScore{0.0688} & \FlowerScore{0.1477} & \FlowerScore{0.0437} & \FlowerScore{0.1613} & \FlowerBest{0.1629} & \FlowerSecond{0.9000} & \FlowerScore{0.8347} & \FlowerScore{0.6926} & \FlowerScore{0.4900} \\
     &  & \quad Permuted Baseline & \FlowerBest{0.0626} & \FlowerSecond{0.1611} & \FlowerBest{0.0512} & \FlowerSecond{0.1706} & \FlowerSecond{0.1566} & \FlowerSecond{0.9000} & \FlowerSecond{0.8390} & \FlowerScore{0.6909} & \FlowerSecond{0.4751} \\
    \cmidrule(l){2-12}
    \rowcolor{FlowerOursTint}
    \FlowerAblationIGLabel & \FlowerAblationOursLabel & \textbf{Expected Gaussian IG} ($1\!\times\!100$) & \FlowerSecond{0.0643} & \FlowerBest{0.1625} & \FlowerScore{0.0492} & \FlowerScore{0.1621} & \FlowerScore{0.1490} & \FlowerBest{0.9030} & \FlowerBest{0.8490} & \FlowerBest{0.6954} & \FlowerBest{0.4634} \\
    \bottomrule
  \end{tabular}%
  }
\end{table*}

\begin{table*}[t]
  \centering
  \caption{Oxford Pets attribution ablation for PVT-v2-B2: Flow-IG (PixelDiT), its Linear Chord and Permuted Baseline controls, and Expected Gaussian IG ($1\!\times\!100$).}
  \label{tab:pets-pvt-pixeldit-ablation}
  \small
  \FlowerTableLayout
  \resizebox{\textwidth}{!}{%
  \begin{tabular}{p{1.6em}p{1.2em} l *{5}{c}!{\textcolor{FlowerDivider}{\vrule width 0.25pt}}*{3}{c}!{\textcolor{FlowerDivider}{\vrule width 0.25pt}}c}
    \toprule
    \FlowerGroupHeaders
    \FlowerHeaders
    \midrule
     &  & Flow-IG (PixelDiT) & \FlowerScore{0.2599} & \FlowerScore{0.6651} & \FlowerSecond{0.1255} & \FlowerBest{0.2524} & \FlowerScore{0.5024} & \FlowerScore{0.7207} & \FlowerScore{0.6741} & \FlowerScore{0.5640} & \FlowerScore{0.5524} \\
     &  & \quad Linear Chord & \FlowerScore{0.2569} & \FlowerScore{0.6560} & \FlowerScore{0.1149} & \FlowerScore{0.2192} & \FlowerSecond{0.5054} & \FlowerSecond{0.7281} & \FlowerScore{0.6824} & \FlowerSecond{0.5687} & \FlowerScore{0.5241} \\
     &  & \quad Permuted Baseline & \FlowerBest{0.2252} & \FlowerSecond{0.6765} & \FlowerScore{0.1192} & \FlowerScore{0.2292} & \FlowerScore{0.5031} & \FlowerScore{0.7186} & \FlowerSecond{0.6830} & \FlowerScore{0.5626} & \FlowerSecond{0.5124} \\
    \cmidrule(l){2-12}
    \rowcolor{FlowerOursTint}
    \FlowerAblationIGLabel & \FlowerAblationOursLabel & \textbf{Expected Gaussian IG} ($1\!\times\!100$) & \FlowerSecond{0.2386} & \FlowerBest{0.6835} & \FlowerBest{0.1318} & \FlowerSecond{0.2461} & \FlowerBest{0.5259} & \FlowerBest{0.7580} & \FlowerBest{0.7100} & \FlowerBest{0.5768} & \FlowerBest{0.5088} \\
    \bottomrule
  \end{tabular}%
  }
\end{table*}

\begin{table*}[t]
  \centering
  \caption{Oxford Pets attribution ablation for ResNet-50: Flow-IG (PixelDiT), its Linear Chord and Permuted Baseline controls, and Expected Gaussian IG ($1\!\times\!100$).}
  \label{tab:pets-resnet-pixeldit-ablation}
  \small
  \FlowerTableLayout
  \resizebox{\textwidth}{!}{%
  \begin{tabular}{p{1.6em}p{1.2em} l *{5}{c}!{\textcolor{FlowerDivider}{\vrule width 0.25pt}}*{3}{c}!{\textcolor{FlowerDivider}{\vrule width 0.25pt}}c}
    \toprule
    \FlowerGroupHeaders
    \FlowerHeaders
    \midrule
     &  & Flow-IG (PixelDiT) & \FlowerScore{0.1039} & \FlowerScore{0.1956} & \FlowerScore{0.0507} & \FlowerBest{0.2505} & \FlowerScore{0.3636} & \FlowerScore{0.8301} & \FlowerScore{0.7553} & \FlowerScore{0.5788} & \FlowerScore{0.6508} \\
     &  & \quad Linear Chord & \FlowerScore{0.1025} & \FlowerScore{0.1983} & \FlowerScore{0.0499} & \FlowerScore{0.2127} & \FlowerSecond{0.3702} & \FlowerScore{0.8355} & \FlowerBest{0.7711} & \FlowerBest{0.5835} & \FlowerScore{0.5900} \\
     &  & \quad Permuted Baseline & \FlowerBest{0.0973} & \FlowerSecond{0.2023} & \FlowerSecond{0.0508} & \FlowerSecond{0.2300} & \FlowerBest{0.3839} & \FlowerBest{0.8517} & \FlowerScore{0.7667} & \FlowerScore{0.5809} & \FlowerSecond{0.5832} \\
    \cmidrule(l){2-12}
    \rowcolor{FlowerOursTint}
    \FlowerAblationIGLabel & \FlowerAblationOursLabel & \textbf{Expected Gaussian IG} ($1\!\times\!100$) & \FlowerSecond{0.0977} & \FlowerBest{0.2067} & \FlowerBest{0.0547} & \FlowerScore{0.2169} & \FlowerScore{0.3517} & \FlowerSecond{0.8477} & \FlowerSecond{0.7703} & \FlowerSecond{0.5824} & \FlowerBest{0.5748} \\
    \bottomrule
  \end{tabular}%
  }
\end{table*}

\clearpage

\begin{table*}[t]
  \centering
  \caption{Flowers102 attribution ablation for PVT-v2-B2: Flow-IG (Diffusion), its Linear Chord and Permuted Baseline controls, and Expected Gaussian IG ($1\!\times\!100$).}
  \label{tab:flowers-pvt-diffusion-ablation}
  \small
  \FlowerTableLayout
  \resizebox{\textwidth}{!}{%
  \begin{tabular}{p{1.6em}p{1.2em} l *{5}{c}!{\textcolor{FlowerDivider}{\vrule width 0.25pt}}*{3}{c}!{\textcolor{FlowerDivider}{\vrule width 0.25pt}}c}
    \toprule
    \FlowerGroupHeaders
    \FlowerHeaders
    \midrule
     &  & Flow-IG (Diffusion) & \FlowerScore{0.1649} & \FlowerScore{0.4191} & \FlowerScore{0.0562} & \FlowerBest{0.2306} & \FlowerScore{0.3751} & \FlowerUnavailable & \FlowerUnavailable & \FlowerUnavailable & \FlowerScore{0.4804} \\
     &  & \quad Linear Chord & \FlowerScore{0.1425} & \FlowerScore{0.4338} & \FlowerBest{0.1101} & \FlowerScore{0.2125} & \FlowerSecond{0.4002} & \FlowerUnavailable & \FlowerUnavailable & \FlowerUnavailable & \FlowerScore{0.4535} \\
     &  & \quad Permuted Baseline & \FlowerSecond{0.1354} & \FlowerSecond{0.4448} & \FlowerScore{0.0943} & \FlowerScore{0.2196} & \FlowerScore{0.3509} & \FlowerUnavailable & \FlowerUnavailable & \FlowerUnavailable & \FlowerSecond{0.4182} \\
    \cmidrule(l){2-12}
    \rowcolor{FlowerOursTint}
    \FlowerAblationIGLabel & \FlowerAblationOursLabel & \textbf{Expected Gaussian IG} ($1\!\times\!100$) & \FlowerBest{0.1303} & \FlowerBest{0.4517} & \FlowerSecond{0.1018} & \FlowerSecond{0.2277} & \FlowerBest{0.4073} & \FlowerUnavailable & \FlowerUnavailable & \FlowerUnavailable & \FlowerBest{0.3987} \\
    \bottomrule
  \end{tabular}%
  }
\end{table*}

\begin{table*}[t]
  \centering
  \caption{Flowers102 attribution ablation for ResNet-50: Flow-IG (Diffusion), its Linear Chord and Permuted Baseline controls, and Expected Gaussian IG ($1\!\times\!100$).}
  \label{tab:flowers-resnet-diffusion-ablation}
  \small
  \FlowerTableLayout
  \resizebox{\textwidth}{!}{%
  \begin{tabular}{p{1.6em}p{1.2em} l *{5}{c}!{\textcolor{FlowerDivider}{\vrule width 0.25pt}}*{3}{c}!{\textcolor{FlowerDivider}{\vrule width 0.25pt}}c}
    \toprule
    \FlowerGroupHeaders
    \FlowerHeaders
    \midrule
     &  & Flow-IG (Diffusion) & \FlowerScore{0.0825} & \FlowerScore{0.1104} & \FlowerScore{0.0155} & \FlowerScore{0.3614} & \FlowerScore{0.2922} & \FlowerUnavailable & \FlowerUnavailable & \FlowerUnavailable & \FlowerScore{0.7186} \\
     &  & \quad Linear Chord & \FlowerScore{0.0731} & \FlowerBest{0.1174} & \FlowerSecond{0.0397} & \FlowerSecond{0.4049} & \FlowerScore{0.3912} & \FlowerUnavailable & \FlowerUnavailable & \FlowerUnavailable & \FlowerScore{0.6598} \\
     &  & \quad Permuted Baseline & \FlowerSecond{0.0657} & \FlowerScore{0.1161} & \FlowerScore{0.0395} & \FlowerBest{0.4401} & \FlowerBest{0.4141} & \FlowerUnavailable & \FlowerUnavailable & \FlowerUnavailable & \FlowerSecond{0.6277} \\
    \cmidrule(l){2-12}
    \rowcolor{FlowerOursTint}
    \FlowerAblationIGLabel & \FlowerAblationOursLabel & \textbf{Expected Gaussian IG} ($1\!\times\!100$) & \FlowerBest{0.0637} & \FlowerSecond{0.1171} & \FlowerBest{0.0408} & \FlowerScore{0.4024} & \FlowerSecond{0.4067} & \FlowerUnavailable & \FlowerUnavailable & \FlowerUnavailable & \FlowerBest{0.6127} \\
    \bottomrule
  \end{tabular}%
  }
\end{table*}

\begin{table*}[t]
  \centering
  \caption{ImageNet attribution ablation for PVT-v2-B2: Flow-IG (Diffusion), its Linear Chord and Permuted Baseline controls, and Expected Gaussian IG ($1\!\times\!100$).}
  \label{tab:imagenet-pvt-diffusion-ablation}
  \small
  \FlowerTableLayout
  \resizebox{\textwidth}{!}{%
  \begin{tabular}{p{1.6em}p{1.2em} l *{5}{c}!{\textcolor{FlowerDivider}{\vrule width 0.25pt}}*{3}{c}!{\textcolor{FlowerDivider}{\vrule width 0.25pt}}c}
    \toprule
    \FlowerGroupHeaders
    \FlowerHeaders
    \midrule
     &  & Flow-IG (Diffusion) & \FlowerScore{0.1889} & \FlowerScore{0.4320} & \FlowerScore{0.0261} & \FlowerScore{0.1981} & \FlowerScore{0.3120} & \FlowerBest{0.7290} & \FlowerBest{0.7326} & \FlowerSecond{0.6675} & \FlowerScore{0.6527} \\
     &  & \quad Linear Chord & \FlowerScore{0.1821} & \FlowerScore{0.4466} & \FlowerScore{0.0465} & \FlowerScore{0.1837} & \FlowerScore{0.3184} & \FlowerScore{0.6950} & \FlowerScore{0.6996} & \FlowerScore{0.6593} & \FlowerScore{0.6482} \\
     &  & \quad Permuted Baseline & \FlowerSecond{0.1730} & \FlowerSecond{0.4517} & \FlowerSecond{0.0489} & \FlowerSecond{0.2123} & \FlowerSecond{0.3477} & \FlowerScore{0.7090} & \FlowerScore{0.7149} & \FlowerScore{0.6629} & \FlowerSecond{0.6181} \\
    \cmidrule(l){2-12}
    \rowcolor{FlowerOursTint}
    \FlowerAblationIGLabel & \FlowerAblationOursLabel & \textbf{Expected Gaussian IG} ($1\!\times\!100$) & \FlowerBest{0.1670} & \FlowerBest{0.4626} & \FlowerBest{0.0628} & \FlowerBest{0.2142} & \FlowerBest{0.3976} & \FlowerSecond{0.7140} & \FlowerSecond{0.7323} & \FlowerBest{0.6703} & \FlowerBest{0.6025} \\
    \bottomrule
  \end{tabular}%
  }
\end{table*}

\begin{table*}[t]
  \centering
  \caption{ImageNet attribution ablation for ResNet-50: Flow-IG (Diffusion), its Linear Chord and Permuted Baseline controls, and Expected Gaussian IG ($1\!\times\!100$).}
  \label{tab:imagenet-resnet-diffusion-ablation}
  \small
  \FlowerTableLayout
  \resizebox{\textwidth}{!}{%
  \begin{tabular}{p{1.6em}p{1.2em} l *{5}{c}!{\textcolor{FlowerDivider}{\vrule width 0.25pt}}*{3}{c}!{\textcolor{FlowerDivider}{\vrule width 0.25pt}}c}
    \toprule
    \FlowerGroupHeaders
    \FlowerHeaders
    \midrule
     &  & Flow-IG (Diffusion) & \FlowerScore{0.0751} & \FlowerScore{0.1422} & \FlowerScore{0.0107} & \FlowerScore{0.1478} & \FlowerScore{0.1177} & \FlowerScore{0.8160} & \FlowerScore{0.7974} & \FlowerSecond{0.6883} & \FlowerScore{0.5702} \\
     &  & \quad Linear Chord & \FlowerScore{0.0679} & \FlowerScore{0.1498} & \FlowerScore{0.0402} & \FlowerScore{0.1581} & \FlowerSecond{0.1507} & \FlowerScore{0.8240} & \FlowerScore{0.7985} & \FlowerScore{0.6867} & \FlowerScore{0.5182} \\
     &  & \quad Permuted Baseline & \FlowerBest{0.0627} & \FlowerSecond{0.1608} & \FlowerSecond{0.0477} & \FlowerBest{0.1703} & \FlowerBest{0.1568} & \FlowerSecond{0.8540} & \FlowerSecond{0.8168} & \FlowerScore{0.6861} & \FlowerSecond{0.4938} \\
    \cmidrule(l){2-12}
    \rowcolor{FlowerOursTint}
    \FlowerAblationIGLabel & \FlowerAblationOursLabel & \textbf{Expected Gaussian IG} ($1\!\times\!100$) & \FlowerSecond{0.0643} & \FlowerBest{0.1625} & \FlowerBest{0.0492} & \FlowerSecond{0.1621} & \FlowerScore{0.1490} & \FlowerBest{0.9030} & \FlowerBest{0.8490} & \FlowerBest{0.6954} & \FlowerBest{0.4634} \\
    \bottomrule
  \end{tabular}%
  }
\end{table*}

\begin{table*}[t]
  \centering
  \caption{Oxford Pets attribution ablation for PVT-v2-B2: Flow-IG (Diffusion), its Linear Chord and Permuted Baseline controls, and Expected Gaussian IG ($1\!\times\!100$).}
  \label{tab:pets-pvt-diffusion-ablation}
  \small
  \FlowerTableLayout
  \resizebox{\textwidth}{!}{%
  \begin{tabular}{p{1.6em}p{1.2em} l *{5}{c}!{\textcolor{FlowerDivider}{\vrule width 0.25pt}}*{3}{c}!{\textcolor{FlowerDivider}{\vrule width 0.25pt}}c}
    \toprule
    \FlowerGroupHeaders
    \FlowerHeaders
    \midrule
     &  & Flow-IG (Diffusion) & \FlowerScore{0.2962} & \FlowerScore{0.6411} & \FlowerScore{0.0503} & \FlowerScore{0.2314} & \FlowerScore{0.4541} & \FlowerScore{0.6902} & \FlowerScore{0.6553} & \FlowerScore{0.5625} & \FlowerScore{0.5986} \\
     &  & \quad Linear Chord & \FlowerScore{0.2490} & \FlowerScore{0.6557} & \FlowerScore{0.1137} & \FlowerScore{0.2133} & \FlowerScore{0.4848} & \FlowerScore{0.6976} & \FlowerScore{0.6708} & \FlowerSecond{0.5689} & \FlowerScore{0.5601} \\
     &  & \quad Permuted Baseline & \FlowerBest{0.2251} & \FlowerSecond{0.6784} & \FlowerSecond{0.1179} & \FlowerSecond{0.2316} & \FlowerSecond{0.4875} & \FlowerSecond{0.7146} & \FlowerSecond{0.6801} & \FlowerScore{0.5636} & \FlowerSecond{0.5480} \\
    \cmidrule(l){2-12}
    \rowcolor{FlowerOursTint}
    \FlowerAblationIGLabel & \FlowerAblationOursLabel & \textbf{Expected Gaussian IG} ($1\!\times\!100$) & \FlowerSecond{0.2386} & \FlowerBest{0.6835} & \FlowerBest{0.1318} & \FlowerBest{0.2461} & \FlowerBest{0.5259} & \FlowerBest{0.7580} & \FlowerBest{0.7100} & \FlowerBest{0.5768} & \FlowerBest{0.5088} \\
    \bottomrule
  \end{tabular}%
  }
\end{table*}

\begin{table*}[t]
  \centering
  \caption{Oxford Pets attribution ablation for ResNet-50: Flow-IG (Diffusion), its Linear Chord and Permuted Baseline controls, and Expected Gaussian IG ($1\!\times\!100$).}
  \label{tab:pets-resnet-diffusion-ablation}
  \small
  \FlowerTableLayout
  \resizebox{\textwidth}{!}{%
  \begin{tabular}{p{1.6em}p{1.2em} l *{5}{c}!{\textcolor{FlowerDivider}{\vrule width 0.25pt}}*{3}{c}!{\textcolor{FlowerDivider}{\vrule width 0.25pt}}c}
    \toprule
    \FlowerGroupHeaders
    \FlowerHeaders
    \midrule
     &  & Flow-IG (Diffusion) & \FlowerScore{0.1256} & \FlowerScore{0.1771} & \FlowerScore{0.0155} & \FlowerSecond{0.2172} & \FlowerScore{0.2548} & \FlowerScore{0.7773} & \FlowerScore{0.7112} & \FlowerScore{0.5652} & \FlowerScore{0.6591} \\
     &  & \quad Linear Chord & \FlowerScore{0.1058} & \FlowerScore{0.1979} & \FlowerSecond{0.0511} & \FlowerScore{0.2052} & \FlowerScore{0.3446} & \FlowerScore{0.8084} & \FlowerScore{0.7420} & \FlowerScore{0.5783} & \FlowerScore{0.5936} \\
     &  & \quad Permuted Baseline & \FlowerBest{0.0971} & \FlowerSecond{0.2023} & \FlowerScore{0.0497} & \FlowerBest{0.2371} & \FlowerBest{0.3679} & \FlowerSecond{0.8274} & \FlowerSecond{0.7544} & \FlowerSecond{0.5787} & \FlowerSecond{0.5906} \\
    \cmidrule(l){2-12}
    \rowcolor{FlowerOursTint}
    \FlowerAblationIGLabel & \FlowerAblationOursLabel & \textbf{Expected Gaussian IG} ($1\!\times\!100$) & \FlowerSecond{0.0977} & \FlowerBest{0.2067} & \FlowerBest{0.0547} & \FlowerScore{0.2169} & \FlowerSecond{0.3517} & \FlowerBest{0.8477} & \FlowerBest{0.7703} & \FlowerBest{0.5824} & \FlowerBest{0.5748} \\
    \bottomrule
  \end{tabular}%
  }
\end{table*}

\begin{table*}[t]
\centering
\caption{Reference selection and path deviations in path attribution.
Given $x_0\sim \coupl{K(\cdot\mid x_1)}$, the path is
$\gamma(t)=\linpath{\gamma_{\mathrm{lin}}(t)}+\excurs{\xi_t}$, where
$\linpath{\gamma_{\mathrm{lin}}(t)=(1-t)x_0+tx_1}$ and
$\excurs{\xi_0}=\excurs{\xi_1}=\mathbf{0}$.
Linear IG has zero path deviation.}
\label{tab:path_attribution_taxonomy}
\small
\setlength{\tabcolsep}{5pt}
\renewcommand{\arraystretch}{1.10}
\begin{tabularx}{\textwidth}{
    @{}l c >{\raggedright\arraybackslash}X@{}
}
\toprule
\textbf{Method}
& \textbf{Reference selection $\coupl{K(\cdot\mid x_1)}$}
& \textbf{Path deviation $\excurs{\xi_t}$} \\
\midrule
Vanilla IG
& $\delta_c$
& $\mathbf{0}$ \\

Expected Gradients
& $p_{\mathrm{data}}$
& $\mathbf{0}$ \\

Blur IG
& $\delta_{B(x_1)}$
& $G_{\sigma(t)}*x_1-\linpath{\gamma_{\mathrm{lin}}(t)}$ \\

Guided IG
& $\delta_c$
& $\bigl(\lambda^{\mathrm{GIG}}(t;\nabla f_c)
          -t\mathbf{1}\bigr)\odot(x_1-x_0)$ \\

SPI
& $\delta_c$
& $\bigl(F_G(t;\omega)-t\mathbf{1}\bigr)
   \odot(x_1-x_0)$ \\

DiffIG
& $\delta_c$
& $\gamma^{\mathrm{DiffIG},(k)}(t)
   -\linpath{\gamma_{\mathrm{lin}}(t)}$ \\

OTFlowSHAP
& $\delta_{h_\theta(x_1)}$
& $\gamma_{\mathrm{ODE}}(t)-\linpath{\gamma_{\mathrm{lin}}(t)}$ \\

Exact $W_2$ OT geodesic
& $\delta_{(T^*)^{-1}(x_1)}$
& $\mathbf{0}$ \\

DDPath
& \multicolumn{2}{l@{}}{
  \textit{Estimator exception: reported rule omits path increments.}} \\

\midrule
\textbf{Expected Gaussian IG (ours)}
& $p_0^{\mathrm{G}}$
& $\mathbf{0}$ \\

Gaussian SPI (ablation)
& $p_0^{\mathrm{G}}$
& $\bigl(F_G(t;\omega)-t\mathbf{1}\bigr)
   \odot(x_1-x_0)$ \\
\bottomrule
\end{tabularx}
\end{table*}

\clearpage
\section{Paired Statistical Testing Protocol and Complete Results}
\label{app:paired_tests}

\subsection{Design, Multiplicity Control, and Reading Contract}
\label{app:paired_design}

To complement the $6$-cell macro-averages in Tables~\ref{tab:main-benchmark-summary} and~\ref{tab:ablation-means-summary} (and the paired-verdict summary in Table~\ref{tab:ablation-summary}), we conduct a sample-paired statistical analysis across all four paired contrasts ($\text{Left} - \text{Right}$):
\begin{itemize}
  \item $\mathbf{C - F}$ (\textbf{Linear Chord $-$ Flow-IG}, testing Hypothesis~\circnumA{A}): isolates the effect of setting the learned path deviation to zero ($\excurs{\xi_t} \equiv \mathbf{0}$) while retaining the exact generator-inverted baseline $\hat{x}_0 = h_\theta(x_1)$ and own-image coupling $\coupl{\hat{\pi}_{h_\theta}}$.
  \item $\mathbf{P - C}$ (\textbf{Permuted Baseline $-$ Linear Chord}, testing Hypothesis~\circnumB{B}): isolates the effect of breaking own-image semantic pairing ($\coupl{\hat{\pi}_{h_\theta}} \to \coupl{\hat{\pi}_{\mathrm{perm}}}$) while holding $\excurs{\xi_t} \equiv \mathbf{0}$ and both empirical marginals $(\hat{p}_0, \hat{p}_1)$ fixed.
  \item $\mathbf{P - F}$ (\textbf{Permuted Baseline $-$ Flow-IG}, composite control): directly estimates the combined effect of removing both the learned path deviation and own-image coupling ($P$ vs.\ $F$) within the generator's inverted reference pool $\hat{p}_0$ (since statistical significance is not transitive across $C - F$ and $P - C$).
  \item $\mathbf{G - F}$ (\textbf{Gaussian Linear IG $-$ Flow-IG}, testing Hypothesis~\circnumC{C}): directly tests our $100$-step generator-free clamped Gaussian linear interpolant ($G$) against the full generative path explainer ($F$).
\end{itemize}

\paragraph{Sample-level pairing and test statistic.}
Within each $(\text{dataset}, \text{backbone}, \text{generator}, \text{metric})$ cell, observations are paired on exact string sample IDs using pairwise finite intersections ($454{,}840$ included sample-pair records across $432$ planned table rows; $384$ valid applicable tests and $48$ structurally not-applicable localization rows on Flowers-102, which has no ground-truth segmentation masks). For each applicable row, we perform a two-sided paired Student $t$-test of zero population mean difference using the sample standard deviation of paired differences ($\mathrm{ddof}=1$, $\mathrm{df}=n-1$), reporting the exact paired mean difference (raw $\text{left} - \text{right}$, not polarity-flipped), pointwise unadjusted $95\%$ $t$-confidence intervals, and signed Cohen's $d_z = \overline{\Delta} / s_{\Delta}$. For $\uparrow$ metrics ($\mathrm{FC}, \mathrm{MC}, \mathrm{FE}, \mathrm{PG}, \mathrm{TopK}, \mathrm{RRA}, \text{InsertionAUC}$), positive differences and positive $d_z$ favor the left method; for $\downarrow$ metrics ($\text{RandomLogit}, \text{DeletionAUC}$), negative differences and negative $d_z$ favor the left method. Pointing Game is analyzed as paired binary hit scores under the same $t$-test procedure (a raw difference of $0.01$ corresponds to $1$ percentage point).

\paragraph{Holm step-down multiplicity families and verdict rules.}
We define four separate Holm step-down multiplicity families~\cite{holm1979simple}, one per named contrast ($\mathbf{C-F}, \mathbf{P-C}, \mathbf{P-F}, \mathbf{G-F}$), each spanning all six dataset--backbone cells, both generators, and the applicable subset of the nine metrics ($m = 96$ planned and tested hypotheses per family; family-wise error rate controlled at $\alpha = 0.05$ per contrast). Structurally not-applicable Flowers-102 localization rows are excluded from multiplicity families. A test receives the verdict \textbf{Favors left} ($\mathrm{L}$) or \underline{Favors right} ($\mathrm{R}$) if and only if $p_{\mathrm{Holm}} < 0.05$ in the corresponding improvement direction, and \textnormal{Not detected} ($\mathrm{ND}$) otherwise. Because no equivalence or non-inferiority margin is specified, $\mathrm{ND}$ indicates that no statistically significant difference was detected after multiplicity adjustment---not formal equivalence---and is kept strictly separate from $\mathrm{L}$ in all tallies (Table~\ref{tab:paired-verdict-tally}). Unadjusted $95\%$ confidence intervals are descriptive and need not agree with Holm-adjusted verdicts.

\begin{table*}[t]
  \centering
  \caption{Complete descriptive verdict tally ($\mathrm{L}/\mathrm{ND}/\mathrm{R}$ = \textbf{Favors left} / Not detected / \underline{Favors right} at $p_{\mathrm{Holm}} < 0.05$; $m=96$ applicable tests per generative contrast family, $48$ per generator, and $m=48$ for $\mathbf{G_{16}-S}$, plus $6$ structurally not-applicable Flowers-102 localization rows per block) and equal-weight mean paired differences $\overline{\Delta}$ (raw $\text{left} - \text{right}$) across all five contrasts $\mathbf{C-F}$, $\mathbf{P-C}$, $\mathbf{P-F}$, $\mathbf{G-F}$, and $\mathbf{G_{16}-S}$.}
  \label{tab:paired-verdict-tally}
  \small
  \setlength{\tabcolsep}{3.5pt}
  \renewcommand{\arraystretch}{1.12}
  \resizebox{\textwidth}{!}{%
  \begin{tabular}{ll *{5}{c}!{\color{gray!50}\vrule width 0.25pt}*{3}{c}!{\color{gray!50}\vrule width 0.25pt}c!{\color{gray!50}\vrule width 0.25pt}c}
    \toprule
    & & \multicolumn{5}{c}{\scriptsize\textsf{Faithfulness (6 cells)}} & \multicolumn{3}{c}{\scriptsize\textsf{Localization (4 cells)}} & \multicolumn{1}{c}{\scriptsize\textsf{Sensitivity (6 cells)}} & \multicolumn{1}{c}{\scriptsize\textsf{48-Test Tally}} \\
    \cmidrule(lr){3-7}\cmidrule(lr){8-10}\cmidrule(lr){11-11}\cmidrule(lr){12-12}
    \textbf{Generator} & \textbf{Contrast (Quantity)} & Del.\ AUC\,$\downarrow$ & Ins.\ AUC\,$\uparrow$ & FC\,$\uparrow$ & MC\,$\uparrow$ & FE\,$\uparrow$ & PG\,$\uparrow$ & TopK\,$\uparrow$ & RRA\,$\uparrow$ & RandomLogit\,$\downarrow$ & $\mathrm{L} / \mathrm{ND} / \mathrm{R}$ ($\mathrm{NA}$) \\
    \midrule
    \multirow{8}{*}{\rotatebox[origin=c]{90}{\scriptsize\textsf{PixelDiT ImageNet}}}
      & \circnumA{A}\ $C - F$ ($\mathrm{L}/\mathrm{ND}/\mathrm{R}$) & $0/4/\underline{2}$ & $0/4/\underline{2}$ & $0/5/\underline{1}$ & $0/0/\underline{6}$ & $0/6/0$ & $0/4/0$ & $\mathbf{2}/1/1$ & $\mathbf{2}/1/1$ & $\mathbf{6}/0/0$ & $10 / 25 / \underline{13}$ ($6$) \\
      & \quad Mean paired $\overline{\Delta}$ & $+0.0006$ & $-0.0023$ & $-0.0057$ & $-0.0354$ & $+0.0021$ & $+0.0027$ & $+0.0015$ & $+0.0010$ & $-0.0386$ & --- \\
      \cmidrule(l){2-12}
      & \circnumB{B}\ $P - C$ ($\mathrm{L}/\mathrm{ND}/\mathrm{R}$) & $\mathbf{6}/0/0$ & $\mathbf{3}/3/0$ & $0/5/\underline{1}$ & $\mathbf{1}/5/0$ & $1/4/1$ & $0/4/0$ & $0/4/0$ & $0/2/\underline{2}$ & $\mathbf{5}/1/0$ & $\mathbf{16} / 28 / 4$ ($6$) \\
      & \quad Mean paired $\overline{\Delta}$ & $-0.0117$ & $+0.0088$ & $-0.0008$ & $+0.0094$ & $-0.0032$ & $+0.0012$ & $+0.0033$ & $-0.0023$ & $-0.0172$ & --- \\
      \cmidrule(l){2-12}
      & $P - F$ ($\mathrm{L}/\mathrm{ND}/\mathrm{R}$) & $\mathbf{5}/1/0$ & $\mathbf{3}/3/0$ & $0/5/\underline{1}$ & $0/1/\underline{5}$ & $\mathbf{2}/3/1$ & $0/4/0$ & $\mathbf{2}/2/0$ & $1/1/\underline{2}$ & $\mathbf{6}/0/0$ & $\mathbf{19} / 20 / 9$ ($6$) \\
      & \quad Mean paired $\overline{\Delta}$ & $-0.0110$ & $+0.0065$ & $-0.0064$ & $-0.0260$ & $-0.0011$ & $+0.0039$ & $+0.0047$ & $-0.0012$ & $-0.0558$ & --- \\
      \cmidrule(l){2-12}
      & \circnumC{C}\ $G - F$ ($\mathrm{L}/\mathrm{ND}/\mathrm{R}$) & $\mathbf{4}/2/0$ & $\mathbf{6}/0/0$ & $0/5/\underline{1}$ & $0/2/\underline{4}$ & $\mathbf{2}/4/0$ & $0/4/0$ & $\mathbf{3}/1/0$ & $\mathbf{2}/2/0$ & $\mathbf{6}/0/0$ & $\mathbf{23} / 20 / 5$ ($6$) \\
      & \quad Mean paired $\overline{\Delta}$ & $-0.0095$ & $+0.0123$ & $-0.0005$ & $-0.0319$ & $+0.0043$ & $+0.0140$ & $+0.0156$ & $+0.0051$ & $-0.0581$ & --- \\
    \midrule
    \multirow{8}{*}{\rotatebox[origin=c]{90}{\scriptsize\textsf{Guided Diffusion}}}
      & \circnumA{A}\ $C - F$ ($\mathrm{L}/\mathrm{ND}/\mathrm{R}$) & $\mathbf{6}/0/0$ & $\mathbf{6}/0/0$ & $\mathbf{6}/0/0$ & $1/4/1$ & $\mathbf{5}/1/0$ & $0/4/0$ & $\mathbf{2}/1/1$ & $\mathbf{2}/1/1$ & $\mathbf{5}/1/0$ & $\mathbf{33} / 12 / 3$ ($6$) \\
      & \quad Mean paired $\overline{\Delta}$ & $-0.0188$ & $+0.0132$ & $+0.0379$ & $-0.0015$ & $+0.0473$ & $+0.0032$ & $+0.0036$ & $+0.0025$ & $-0.0410$ & --- \\
      \cmidrule(l){2-12}
      & \circnumB{B}\ $P - C$ ($\mathrm{L}/\mathrm{ND}/\mathrm{R}$) & $\mathbf{6}/0/0$ & $\mathbf{3}/3/0$ & $0/5/\underline{1}$ & $\mathbf{2}/4/0$ & $1/4/1$ & $0/4/0$ & $\mathbf{1}/3/0$ & $0/3/\underline{1}$ & $\mathbf{4}/2/0$ & $\mathbf{17} / 28 / 3$ ($6$) \\
      & \quad Mean paired $\overline{\Delta}$ & $-0.0102$ & $+0.0088$ & $-0.0006$ & $+0.0222$ & $+0.0058$ & $+0.0200$ & $+0.0138$ & $-0.0005$ & $-0.0229$ & --- \\
      \cmidrule(l){2-12}
      & $P - F$ ($\mathrm{L}/\mathrm{ND}/\mathrm{R}$) & $\mathbf{6}/0/0$ & $\mathbf{6}/0/0$ & $\mathbf{6}/0/0$ & $\mathbf{2}/4/0$ & $\mathbf{5}/1/0$ & $\mathbf{1}/3/0$ & $\mathbf{3}/0/1$ & $1/2/1$ & $\mathbf{6}/0/0$ & $\mathbf{36} / 10 / 2$ ($6$) \\
      & \quad Mean paired $\overline{\Delta}$ & $-0.0291$ & $+0.0220$ & $+0.0373$ & $+0.0208$ & $+0.0532$ & $+0.0231$ & $+0.0174$ & $+0.0020$ & $-0.0639$ & --- \\
      \cmidrule(l){2-12}
      & \circnumC{C}\ $G - F$ ($\mathrm{L}/\mathrm{ND}/\mathrm{R}$) & $\mathbf{6}/0/0$ & $\mathbf{6}/0/0$ & $\mathbf{6}/0/0$ & $\mathbf{1}/5/0$ & $\mathbf{6}/0/0$ & $\mathbf{3}/1/0$ & $\mathbf{3}/1/0$ & $\mathbf{3}/1/0$ & $\mathbf{6}/0/0$ & $\mathbf{40} / 8 / 0$ ($6$) \\
      & \quad Mean paired $\overline{\Delta}$ & $-0.0286$ & $+0.0270$ & $+0.0445$ & $+0.0138$ & $+0.0721$ & $+0.0526$ & $+0.0413$ & $+0.0104$ & $-0.0865$ & --- \\
    \midrule
    \multirow{2}{*}{\scriptsize\textsf{\shortstack[l]{Gen.-\\free}}}
      & \circnumD{D}\ $G_{16} - S$ ($\mathrm{L}/\mathrm{ND}/\mathrm{R}$) & $\mathbf{5}/1/0$ & $\mathbf{6}/0/0$ & $\mathbf{6}/0/0$ & $\mathbf{6}/0/0$ & $\mathbf{6}/0/0$ & $\mathbf{1}/3/0$ & $\mathbf{2}/1/1$ & $\mathbf{2}/2/0$ & $\mathbf{6}/0/0$ & $\mathbf{40} / 7 / 1$ ($6$) \\
      & \quad Mean paired $\overline{\Delta}$ & $-0.0088$ & $+0.0405$ & $+0.1289$ & $+0.1283$ & $+0.1426$ & $+0.0086$ & $+0.0010$ & $+0.0131$ & $-0.1451$ & --- \\
    \bottomrule
  \end{tabular}%
  }
\end{table*}

\subsection{Complete Cell-Level Paired Student \texorpdfstring{$t$}{t}-Test Tables}
\label{app:paired_full_tables}

Tables~\ref{tab:paired-cf-diffusion}--\ref{tab:paired-g16s-gaussian} report all $486$ paired-test rows across the $9$ contrast--backend pairs ($432$ rows for the $8$ generative contrast tables plus $54$ rows for $G_{16} - S$ across $6$ cells $\times$ $9$ metrics, including structurally not-applicable Flowers-102 localization rows).

\begin{table*}[t]
  \centering
  \caption{Complete paired Student $t$-test results for contrast $\mathbf{C-F}$ (Linear Chord $-$ Flow-IG; Path-deviation ablation ($\excurs{\xi_t} \to \mathbf{0}$, Hypothesis~\circnumA{A})) on generator \textbf{Flow-IG (Diffusion)}. Differences and Cohen's $d_z$ are raw $\text{left} - \text{right}$ (positive favors left for $\uparrow$ metrics; negative favors left for $\downarrow$ metrics). Multiplicity adjustment uses Holm's step-down procedure across the $m=96$ applicable tests of family $\mathbf{C-F}$. Tally across the $48$ applicable tests for this generator: $\mathrm{L}=33$ Favors left, $\mathrm{ND}=12$ Not detected, $\mathrm{R}=3$ Favors right ($\mathrm{NA}=6$ Flowers-102 localization exclusions).}
  \label{tab:paired-cf-diffusion}
  \scriptsize
  \setlength{\tabcolsep}{4pt}
  \renewcommand{\arraystretch}{0.96}
  \resizebox{\textwidth}{!}{%
  \begin{tabular}{ll c r r c r r r l}
    \toprule
    \textbf{Dataset $\times$ Backbone} & \textbf{Metric} & \textbf{Pol.} & $n$ & \textbf{Mean diff.\ ($\overline{\Delta}$)} & \textbf{95\% CI (unadjusted)} & \textbf{Cohen $d_z$} & $p_{\mathrm{raw}}$ & $p_{\mathrm{Holm}}$ & \textbf{Verdict} \\
    \midrule
    \multirow{9}{*}{Flowers-102 $\times$ PVT-v2-B2} & FaithfulnessCorrelation ($\mathrm{FC}$) & $\uparrow$ & $1020$ & $0.0539242$ & $[0.0488636,\, 0.0589848]$ & $0.654706$ & $4.56885 \times 10^{-81}$ & $4.06628 \times 10^{-79}$ & \textbf{Favors left} \\
     & MonotonicityCorrelation ($\mathrm{MC}$) & $\uparrow$ & $1020$ & $-0.0181243$ & $[-0.0308388,\, -0.00540983]$ & $-0.0875845$ & $0.00525143$ & $0.194303$ & Not detected \\
     & FaithfulnessEstimate ($\mathrm{FE}$) & $\uparrow$ & $1020$ & $0.0250409$ & $[0.012728,\, 0.0373538]$ & $0.124955$ & $7.05875 \times 10^{-5}$ & $0.0028235$ & \textbf{Favors left} \\
     & PointingGame ($\mathrm{PG}$) & $\uparrow$ & \textnormal{---} & \textnormal{---} & $[\textnormal{---},\, \textnormal{---}]$ & \textnormal{---} & \textnormal{---} & \textnormal{---} & \textnormal{Not applicable} \\
     & TopKIntersection ($\mathrm{TopK}$) & $\uparrow$ & \textnormal{---} & \textnormal{---} & $[\textnormal{---},\, \textnormal{---}]$ & \textnormal{---} & \textnormal{---} & \textnormal{---} & \textnormal{Not applicable} \\
     & RelevanceRankAccuracy ($\mathrm{RRA}$) & $\uparrow$ & \textnormal{---} & \textnormal{---} & $[\textnormal{---},\, \textnormal{---}]$ & \textnormal{---} & \textnormal{---} & \textnormal{---} & \textnormal{Not applicable} \\
     & RandomLogit & $\downarrow$ & $1020$ & $-0.0269039$ & $[-0.0320043,\, -0.0218035]$ & $-0.324098$ & $6.06734 \times 10^{-24}$ & $4.36848 \times 10^{-22}$ & \textbf{Favors left} \\
     & DeletionAUC & $\downarrow$ & $1020$ & $-0.0223725$ & $[-0.025285,\, -0.0194601]$ & $-0.47198$ & $1.67736 \times 10^{-46}$ & $1.34189 \times 10^{-44}$ & \textbf{Favors left} \\
     & InsertionAUC & $\uparrow$ & $1020$ & $0.0146394$ & $[0.0108957,\, 0.0183832]$ & $0.240261$ & $3.90984 \times 10^{-14}$ & $2.54139 \times 10^{-12}$ & \textbf{Favors left} \\
    \midrule
    \multirow{9}{*}{Flowers-102 $\times$ ResNet-50} & FaithfulnessCorrelation ($\mathrm{FC}$) & $\uparrow$ & $1020$ & $0.0242324$ & $[0.0178579,\, 0.0306069]$ & $0.233568$ & $1.85524 \times 10^{-13}$ & $1.1688 \times 10^{-11}$ & \textbf{Favors left} \\
     & MonotonicityCorrelation ($\mathrm{MC}$) & $\uparrow$ & $1020$ & $0.0434068$ & $[0.0314115,\, 0.055402]$ & $0.222337$ & $2.32288 \times 10^{-12}$ & $1.34727 \times 10^{-10}$ & \textbf{Favors left} \\
     & FaithfulnessEstimate ($\mathrm{FE}$) & $\uparrow$ & $1020$ & $0.0990258$ & $[0.0866857,\, 0.111366]$ & $0.493052$ & $3.60528 \times 10^{-50}$ & $2.99239 \times 10^{-48}$ & \textbf{Favors left} \\
     & PointingGame ($\mathrm{PG}$) & $\uparrow$ & \textnormal{---} & \textnormal{---} & $[\textnormal{---},\, \textnormal{---}]$ & \textnormal{---} & \textnormal{---} & \textnormal{---} & \textnormal{Not applicable} \\
     & TopKIntersection ($\mathrm{TopK}$) & $\uparrow$ & \textnormal{---} & \textnormal{---} & $[\textnormal{---},\, \textnormal{---}]$ & \textnormal{---} & \textnormal{---} & \textnormal{---} & \textnormal{Not applicable} \\
     & RelevanceRankAccuracy ($\mathrm{RRA}$) & $\uparrow$ & \textnormal{---} & \textnormal{---} & $[\textnormal{---},\, \textnormal{---}]$ & \textnormal{---} & \textnormal{---} & \textnormal{---} & \textnormal{Not applicable} \\
     & RandomLogit & $\downarrow$ & $1020$ & $-0.0587322$ & $[-0.0627844,\, -0.0546801]$ & $-0.890534$ & $1.87386 \times 10^{-131}$ & $1.74269 \times 10^{-129}$ & \textbf{Favors left} \\
     & DeletionAUC & $\downarrow$ & $1020$ & $-0.0094168$ & $[-0.0107842,\, -0.00804942]$ & $-0.423135$ & $2.12357 \times 10^{-38}$ & $1.65639 \times 10^{-36}$ & \textbf{Favors left} \\
     & InsertionAUC & $\uparrow$ & $1020$ & $0.00696578$ & $[0.00553248,\, 0.00839908]$ & $0.298605$ & $1.04545 \times 10^{-20}$ & $7.31816 \times 10^{-19}$ & \textbf{Favors left} \\
    \midrule
    \multirow{9}{*}{ImageNet $\times$ PVT-v2-B2} & FaithfulnessCorrelation ($\mathrm{FC}$) & $\uparrow$ & $1000$ & $0.0204167$ & $[0.0144245,\, 0.0264088]$ & $0.211435$ & $3.8052 \times 10^{-11}$ & $2.13091 \times 10^{-9}$ & \textbf{Favors left} \\
     & MonotonicityCorrelation ($\mathrm{MC}$) & $\uparrow$ & $1000$ & $-0.0143231$ & $[-0.0276024,\, -0.00104374]$ & $-0.0669321$ & $0.0345422$ & $1$ & Not detected \\
     & FaithfulnessEstimate ($\mathrm{FE}$) & $\uparrow$ & $1000$ & $0.00646278$ & $[-0.00710553,\, 0.0200311]$ & $0.0295575$ & $0.350173$ & $1$ & Not detected \\
     & PointingGame ($\mathrm{PG}$) & $\uparrow$ & $1000$ & $-0.034$ & $[-0.0592074,\, -0.00879258]$ & $-0.0836999$ & $0.00825299$ & $0.288855$ & Not detected \\
     & TopKIntersection ($\mathrm{TopK}$) & $\uparrow$ & $1000$ & $-0.033008$ & $[-0.03905,\, -0.026966]$ & $-0.339013$ & $1.8579 \times 10^{-25}$ & $1.35627 \times 10^{-23}$ & \underline{Favors right} \\
     & RelevanceRankAccuracy ($\mathrm{RRA}$) & $\uparrow$ & $1000$ & $-0.00817463$ & $[-0.00984333,\, -0.00650593]$ & $-0.303993$ & $5.51108 \times 10^{-21}$ & $3.91287 \times 10^{-19}$ & \underline{Favors right} \\
     & RandomLogit & $\downarrow$ & $1000$ & $-0.00451345$ & $[-0.0112223,\, 0.0021954]$ & $-0.041748$ & $0.187075$ & $1$ & Not detected \\
     & DeletionAUC & $\downarrow$ & $1000$ & $-0.00684617$ & $[-0.0105709,\, -0.00312147]$ & $-0.114059$ & $0.000325251$ & $0.0126848$ & \textbf{Favors left} \\
     & InsertionAUC & $\uparrow$ & $1000$ & $0.0146155$ & $[0.0100403,\, 0.0191907]$ & $0.198235$ & $5.40664 \times 10^{-10}$ & $2.81145 \times 10^{-8}$ & \textbf{Favors left} \\
    \midrule
    \multirow{9}{*}{ImageNet $\times$ ResNet-50} & FaithfulnessCorrelation ($\mathrm{FC}$) & $\uparrow$ & $1000$ & $0.0294815$ & $[0.023097,\, 0.0358659]$ & $0.286551$ & $6.62724 \times 10^{-19}$ & $4.5728 \times 10^{-17}$ & \textbf{Favors left} \\
     & MonotonicityCorrelation ($\mathrm{MC}$) & $\uparrow$ & $1000$ & $0.0103137$ & $[-0.00225067,\, 0.022878]$ & $0.0509387$ & $0.107534$ & $1$ & Not detected \\
     & FaithfulnessEstimate ($\mathrm{FE}$) & $\uparrow$ & $1000$ & $0.0329904$ & $[0.0188636,\, 0.0471171]$ & $0.144917$ & $5.17106 \times 10^{-6}$ & $0.000217185$ & \textbf{Favors left} \\
     & PointingGame ($\mathrm{PG}$) & $\uparrow$ & $1000$ & $0.008$ & $[-0.0112301,\, 0.0272301]$ & $0.0258156$ & $0.414489$ & $1$ & Not detected \\
     & TopKIntersection ($\mathrm{TopK}$) & $\uparrow$ & $1000$ & $0.001116$ & $[-0.00265272,\, 0.00488472]$ & $0.0183758$ & $0.561309$ & $1$ & Not detected \\
     & RelevanceRankAccuracy ($\mathrm{RRA}$) & $\uparrow$ & $1000$ & $-0.00160141$ & $[-0.00291061,\, -0.000292209]$ & $-0.0759051$ & $0.0165624$ & $0.563122$ & Not detected \\
     & RandomLogit & $\downarrow$ & $1000$ & $-0.0520465$ & $[-0.0570176,\, -0.0470753]$ & $-0.649692$ & $1.62801 \times 10^{-78}$ & $1.43265 \times 10^{-76}$ & \textbf{Favors left} \\
     & DeletionAUC & $\downarrow$ & $1000$ & $-0.00727655$ & $[-0.00909918,\, -0.00545392]$ & $-0.247743$ & $1.2012 \times 10^{-14}$ & $7.92793 \times 10^{-13}$ & \textbf{Favors left} \\
     & InsertionAUC & $\uparrow$ & $1000$ & $0.00762129$ & $[0.00513257,\, 0.01011]$ & $0.190032$ & $2.60726 \times 10^{-9}$ & $1.27756 \times 10^{-7}$ & \textbf{Favors left} \\
    \midrule
    \multirow{9}{*}{Oxford Pets $\times$ PVT-v2-B2} & FaithfulnessCorrelation ($\mathrm{FC}$) & $\uparrow$ & $1480$ & $0.063483$ & $[0.0594881,\, 0.0674779]$ & $0.810255$ & $2.17039 \times 10^{-164}$ & $2.04017 \times 10^{-162}$ & \textbf{Favors left} \\
     & MonotonicityCorrelation ($\mathrm{MC}$) & $\uparrow$ & $1479$ & $-0.0180919$ & $[-0.0279892,\, -0.0081947]$ & $-0.0932376$ & $0.000347106$ & $0.01319$ & \underline{Favors right} \\
     & FaithfulnessEstimate ($\mathrm{FE}$) & $\uparrow$ & $1480$ & $0.0306934$ & $[0.0214148,\, 0.039972]$ & $0.168669$ & $1.17714 \times 10^{-10}$ & $6.35656 \times 10^{-9}$ & \textbf{Favors left} \\
     & PointingGame ($\mathrm{PG}$) & $\uparrow$ & $1475$ & $0.00745763$ & $[-0.0150764,\, 0.0299917]$ & $0.0169033$ & $0.516322$ & $1$ & Not detected \\
     & TopKIntersection ($\mathrm{TopK}$) & $\uparrow$ & $1475$ & $0.0155051$ & $[0.0110307,\, 0.0199795]$ & $0.17699$ & $1.54185 \times 10^{-11}$ & $8.78854 \times 10^{-10}$ & \textbf{Favors left} \\
     & RelevanceRankAccuracy ($\mathrm{RRA}$) & $\uparrow$ & $1475$ & $0.00647223$ & $[0.00475725,\, 0.00818721]$ & $0.192754$ & $2.2311 \times 10^{-13}$ & $1.38328 \times 10^{-11}$ & \textbf{Favors left} \\
     & RandomLogit & $\downarrow$ & $1480$ & $-0.0384293$ & $[-0.0421311,\, -0.0347275]$ & $-0.529325$ & $1.86218 \times 10^{-81}$ & $1.67596 \times 10^{-79}$ & \textbf{Favors left} \\
     & DeletionAUC & $\downarrow$ & $1480$ & $-0.0471754$ & $[-0.0522936,\, -0.0420573]$ & $-0.469979$ & $3.5703 \times 10^{-66}$ & $3.07046 \times 10^{-64}$ & \textbf{Favors left} \\
     & InsertionAUC & $\uparrow$ & $1480$ & $0.0146225$ & $[0.00995529,\, 0.0192898]$ & $0.159748$ & $1.02235 \times 10^{-9}$ & $5.11177 \times 10^{-8}$ & \textbf{Favors left} \\
    \midrule
    \multirow{9}{*}{Oxford Pets $\times$ ResNet-50} & FaithfulnessCorrelation ($\mathrm{FC}$) & $\uparrow$ & $1480$ & $0.0355702$ & $[0.0301753,\, 0.0409652]$ & $0.336179$ & $2.51679 \times 10^{-36}$ & $1.91276 \times 10^{-34}$ & \textbf{Favors left} \\
     & MonotonicityCorrelation ($\mathrm{MC}$) & $\uparrow$ & $1480$ & $-0.0119093$ & $[-0.0218965,\, -0.00192206]$ & $-0.0608014$ & $0.019464$ & $0.642312$ & Not detected \\
     & FaithfulnessEstimate ($\mathrm{FE}$) & $\uparrow$ & $1480$ & $0.089807$ & $[0.0781848,\, 0.101429]$ & $0.393996$ & $2.38497 \times 10^{-48}$ & $1.93183 \times 10^{-46}$ & \textbf{Favors left} \\
     & PointingGame ($\mathrm{PG}$) & $\uparrow$ & $1477$ & $0.0311442$ & $[0.00921071,\, 0.0530777]$ & $0.0724742$ & $0.00541598$ & $0.194975$ & Not detected \\
     & TopKIntersection ($\mathrm{TopK}$) & $\uparrow$ & $1477$ & $0.0307271$ & $[0.0267866,\, 0.0346677]$ & $0.397997$ & $3.8867 \times 10^{-49}$ & $3.18709 \times 10^{-47}$ & \textbf{Favors left} \\
     & RelevanceRankAccuracy ($\mathrm{RRA}$) & $\uparrow$ & $1477$ & $0.013116$ & $[0.011707,\, 0.0145249]$ & $0.475124$ & $2.53853 \times 10^{-67}$ & $2.20852 \times 10^{-65}$ & \textbf{Favors left} \\
     & RandomLogit & $\downarrow$ & $1480$ & $-0.0654336$ & $[-0.0691897,\, -0.0616775]$ & $-0.888262$ & $3.90407 \times 10^{-189}$ & $3.70887 \times 10^{-187}$ & \textbf{Favors left} \\
     & DeletionAUC & $\downarrow$ & $1480$ & $-0.0198031$ & $[-0.0221528,\, -0.0174534]$ & $-0.429726$ & $1.82486 \times 10^{-56}$ & $1.53288 \times 10^{-54}$ & \textbf{Favors left} \\
     & InsertionAUC & $\uparrow$ & $1480$ & $0.0207557$ & $[0.0178045,\, 0.0237069]$ & $0.358601$ & $8.12127 \times 10^{-41}$ & $6.4158 \times 10^{-39}$ & \textbf{Favors left} \\
    \bottomrule
  \end{tabular}%
  }
\end{table*}

\begin{table*}[t]
  \centering
  \caption{Complete paired Student $t$-test results for contrast $\mathbf{C-F}$ (Linear Chord $-$ Flow-IG; Path-deviation ablation ($\excurs{\xi_t} \to \mathbf{0}$, Hypothesis~\circnumA{A})) on generator \textbf{Flow-IG (PixelDiT)}. Differences and Cohen's $d_z$ are raw $\text{left} - \text{right}$ (positive favors left for $\uparrow$ metrics; negative favors left for $\downarrow$ metrics). Multiplicity adjustment uses Holm's step-down procedure across the $m=96$ applicable tests of family $\mathbf{C-F}$. Tally across the $48$ applicable tests for this generator: $\mathrm{L}=10$ Favors left, $\mathrm{ND}=25$ Not detected, $\mathrm{R}=13$ Favors right ($\mathrm{NA}=6$ Flowers-102 localization exclusions).}
  \label{tab:paired-cf-pixeldit}
  \scriptsize
  \setlength{\tabcolsep}{4pt}
  \renewcommand{\arraystretch}{0.96}
  \resizebox{\textwidth}{!}{%
  \begin{tabular}{ll c r r c r r r l}
    \toprule
    \textbf{Dataset $\times$ Backbone} & \textbf{Metric} & \textbf{Pol.} & $n$ & \textbf{Mean diff.\ ($\overline{\Delta}$)} & \textbf{95\% CI (unadjusted)} & \textbf{Cohen $d_z$} & $p_{\mathrm{raw}}$ & $p_{\mathrm{Holm}}$ & \textbf{Verdict} \\
    \midrule
    \multirow{9}{*}{Flowers-102 $\times$ PVT-v2-B2} & FaithfulnessCorrelation ($\mathrm{FC}$) & $\uparrow$ & $1020$ & $-0.00388755$ & $[-0.0078865,\, 0.000111402]$ & $-0.0597302$ & $0.05672$ & $1$ & Not detected \\
     & MonotonicityCorrelation ($\mathrm{MC}$) & $\uparrow$ & $1020$ & $-0.0255239$ & $[-0.0367449,\, -0.014303]$ & $-0.13976$ & $8.9586 \times 10^{-6}$ & $0.000367303$ & \underline{Favors right} \\
     & FaithfulnessEstimate ($\mathrm{FE}$) & $\uparrow$ & $1020$ & $0.00640486$ & $[-0.00374475,\, 0.0165545]$ & $0.0387726$ & $0.215891$ & $1$ & Not detected \\
     & PointingGame ($\mathrm{PG}$) & $\uparrow$ & \textnormal{---} & \textnormal{---} & $[\textnormal{---},\, \textnormal{---}]$ & \textnormal{---} & \textnormal{---} & \textnormal{---} & \textnormal{Not applicable} \\
     & TopKIntersection ($\mathrm{TopK}$) & $\uparrow$ & \textnormal{---} & \textnormal{---} & $[\textnormal{---},\, \textnormal{---}]$ & \textnormal{---} & \textnormal{---} & \textnormal{---} & \textnormal{Not applicable} \\
     & RelevanceRankAccuracy ($\mathrm{RRA}$) & $\uparrow$ & \textnormal{---} & \textnormal{---} & $[\textnormal{---},\, \textnormal{---}]$ & \textnormal{---} & \textnormal{---} & \textnormal{---} & \textnormal{Not applicable} \\
     & RandomLogit & $\downarrow$ & $1020$ & $-0.0262625$ & $[-0.0301687,\, -0.0223564]$ & $-0.413093$ & $8.2965 \times 10^{-37}$ & $6.3883 \times 10^{-35}$ & \textbf{Favors left} \\
     & DeletionAUC & $\downarrow$ & $1020$ & $0.00596666$ & $[0.00372543,\, 0.00820789]$ & $0.163572$ & $2.12198 \times 10^{-7}$ & $9.54889 \times 10^{-6}$ & \underline{Favors right} \\
     & InsertionAUC & $\uparrow$ & $1020$ & $-0.00793779$ & $[-0.0106709,\, -0.00520467]$ & $-0.178446$ & $1.57612 \times 10^{-8}$ & $7.56538 \times 10^{-7}$ & \underline{Favors right} \\
    \midrule
    \multirow{9}{*}{Flowers-102 $\times$ ResNet-50} & FaithfulnessCorrelation ($\mathrm{FC}$) & $\uparrow$ & $1020$ & $-0.00714062$ & $[-0.0133034,\, -0.000977883]$ & $-0.0711912$ & $0.0231935$ & $0.742191$ & Not detected \\
     & MonotonicityCorrelation ($\mathrm{MC}$) & $\uparrow$ & $1020$ & $-0.0453998$ & $[-0.055845,\, -0.0349545]$ & $-0.267053$ & $5.27729 \times 10^{-17}$ & $3.53579 \times 10^{-15}$ & \underline{Favors right} \\
     & FaithfulnessEstimate ($\mathrm{FE}$) & $\uparrow$ & $1020$ & $-0.00103744$ & $[-0.0114016,\, 0.00932677]$ & $-0.00615021$ & $0.844319$ & $1$ & Not detected \\
     & PointingGame ($\mathrm{PG}$) & $\uparrow$ & \textnormal{---} & \textnormal{---} & $[\textnormal{---},\, \textnormal{---}]$ & \textnormal{---} & \textnormal{---} & \textnormal{---} & \textnormal{Not applicable} \\
     & TopKIntersection ($\mathrm{TopK}$) & $\uparrow$ & \textnormal{---} & \textnormal{---} & $[\textnormal{---},\, \textnormal{---}]$ & \textnormal{---} & \textnormal{---} & \textnormal{---} & \textnormal{Not applicable} \\
     & RelevanceRankAccuracy ($\mathrm{RRA}$) & $\uparrow$ & \textnormal{---} & \textnormal{---} & $[\textnormal{---},\, \textnormal{---}]$ & \textnormal{---} & \textnormal{---} & \textnormal{---} & \textnormal{Not applicable} \\
     & RandomLogit & $\downarrow$ & $1020$ & $-0.0404107$ & $[-0.0441679,\, -0.0366535]$ & $-0.660839$ & $2.57051 \times 10^{-82}$ & $2.33916 \times 10^{-80}$ & \textbf{Favors left} \\
     & DeletionAUC & $\downarrow$ & $1020$ & $0.00351678$ & $[0.00246759,\, 0.00456598]$ & $0.205946$ & $7.63946 \times 10^{-11}$ & $4.2017 \times 10^{-9}$ & \underline{Favors right} \\
     & InsertionAUC & $\uparrow$ & $1020$ & $-0.000119616$ & $[-0.00137937,\, 0.00114014]$ & $-0.00583402$ & $0.852228$ & $1$ & Not detected \\
    \midrule
    \multirow{9}{*}{ImageNet $\times$ PVT-v2-B2} & FaithfulnessCorrelation ($\mathrm{FC}$) & $\uparrow$ & $1000$ & $-0.00534526$ & $[-0.0103469,\, -0.000343601]$ & $-0.0663177$ & $0.0362312$ & $1$ & Not detected \\
     & MonotonicityCorrelation ($\mathrm{MC}$) & $\uparrow$ & $1000$ & $-0.0382407$ & $[-0.0516171,\, -0.0248644]$ & $-0.177404$ & $2.61905 \times 10^{-8}$ & $1.23095 \times 10^{-6}$ & \underline{Favors right} \\
     & FaithfulnessEstimate ($\mathrm{FE}$) & $\uparrow$ & $1000$ & $-0.00875448$ & $[-0.0221729,\, 0.00466395]$ & $-0.0404858$ & $0.200746$ & $1$ & Not detected \\
     & PointingGame ($\mathrm{PG}$) & $\uparrow$ & $1000$ & $-0.019$ & $[-0.045241,\, 0.00724098]$ & $-0.0449312$ & $0.155672$ & $1$ & Not detected \\
     & TopKIntersection ($\mathrm{TopK}$) & $\uparrow$ & $1000$ & $-0.018747$ & $[-0.0236218,\, -0.0138722]$ & $-0.238644$ & $1.00402 \times 10^{-13}$ & $6.42573 \times 10^{-12}$ & \underline{Favors right} \\
     & RelevanceRankAccuracy ($\mathrm{RRA}$) & $\uparrow$ & $1000$ & $-0.00459414$ & $[-0.00598785,\, -0.00320043]$ & $-0.204553$ & $1.5468 \times 10^{-10}$ & $8.19803 \times 10^{-9}$ & \underline{Favors right} \\
     & RandomLogit & $\downarrow$ & $1000$ & $-0.0206661$ & $[-0.0263082,\, -0.015024]$ & $-0.227296$ & $1.28831 \times 10^{-12}$ & $7.72984 \times 10^{-11}$ & \textbf{Favors left} \\
     & DeletionAUC & $\downarrow$ & $1000$ & $-0.00314873$ & $[-0.00629614,\, -1.3143e-06]$ & $-0.0620806$ & $0.0499046$ & $1$ & Not detected \\
     & InsertionAUC & $\uparrow$ & $1000$ & $0.00217924$ & $[-0.0018429,\, 0.00620139]$ & $0.033622$ & $0.287939$ & $1$ & Not detected \\
    \midrule
    \multirow{9}{*}{ImageNet $\times$ ResNet-50} & FaithfulnessCorrelation ($\mathrm{FC}$) & $\uparrow$ & $1000$ & $-0.00620266$ & $[-0.0124834,\, 7.81113e-05]$ & $-0.0612829$ & $0.0529125$ & $1$ & Not detected \\
     & MonotonicityCorrelation ($\mathrm{MC}$) & $\uparrow$ & $1000$ & $-0.0324003$ & $[-0.0443313,\, -0.0204693]$ & $-0.168518$ & $1.22064 \times 10^{-7}$ & $5.61493 \times 10^{-6}$ & \underline{Favors right} \\
     & FaithfulnessEstimate ($\mathrm{FE}$) & $\uparrow$ & $1000$ & $0.00658758$ & $[-0.00684829,\, 0.0200235]$ & $0.0304253$ & $0.336216$ & $1$ & Not detected \\
     & PointingGame ($\mathrm{PG}$) & $\uparrow$ & $1000$ & $0.017$ & $[-0.00347051,\, 0.0374705]$ & $0.0515341$ & $0.10349$ & $1$ & Not detected \\
     & TopKIntersection ($\mathrm{TopK}$) & $\uparrow$ & $1000$ & $0.000578$ & $[-0.00237236,\, 0.00352836]$ & $0.012157$ & $0.700735$ & $1$ & Not detected \\
     & RelevanceRankAccuracy ($\mathrm{RRA}$) & $\uparrow$ & $1000$ & $-0.000711505$ & $[-0.00172881,\, 0.000305799]$ & $-0.0434012$ & $0.170226$ & $1$ & Not detected \\
     & RandomLogit & $\downarrow$ & $1000$ & $-0.0553218$ & $[-0.0603116,\, -0.0503319]$ & $-0.687993$ & $3.24515 \times 10^{-86}$ & $2.98554 \times 10^{-84}$ & \textbf{Favors left} \\
     & DeletionAUC & $\downarrow$ & $1000$ & $0.00190522$ & $[0.000246487,\, 0.00356395]$ & $0.071276$ & $0.0244157$ & $0.756887$ & Not detected \\
     & InsertionAUC & $\uparrow$ & $1000$ & $-0.00144672$ & $[-0.00403934,\, 0.00114589]$ & $-0.0346276$ & $0.273772$ & $1$ & Not detected \\
    \midrule
    \multirow{9}{*}{Oxford Pets $\times$ PVT-v2-B2} & FaithfulnessCorrelation ($\mathrm{FC}$) & $\uparrow$ & $1480$ & $-0.0105247$ & $[-0.0138753,\, -0.00717416]$ & $-0.160165$ & $9.26239 \times 10^{-10}$ & $4.72382 \times 10^{-8}$ & \underline{Favors right} \\
     & MonotonicityCorrelation ($\mathrm{MC}$) & $\uparrow$ & $1479$ & $-0.0332066$ & $[-0.0423608,\, -0.0240524]$ & $-0.185022$ & $1.73201 \times 10^{-12}$ & $1.02189 \times 10^{-10}$ & \underline{Favors right} \\
     & FaithfulnessEstimate ($\mathrm{FE}$) & $\uparrow$ & $1480$ & $0.0029998$ & $[-0.00480626,\, 0.0108059]$ & $0.0195945$ & $0.451081$ & $1$ & Not detected \\
     & PointingGame ($\mathrm{PG}$) & $\uparrow$ & $1475$ & $0.00745763$ & $[-0.0137827,\, 0.0286979]$ & $0.0179328$ & $0.491105$ & $1$ & Not detected \\
     & TopKIntersection ($\mathrm{TopK}$) & $\uparrow$ & $1475$ & $0.00826034$ & $[0.00479935,\, 0.0117213]$ & $0.121901$ & $3.10684 \times 10^{-6}$ & $0.000133594$ & \textbf{Favors left} \\
     & RelevanceRankAccuracy ($\mathrm{RRA}$) & $\uparrow$ & $1475$ & $0.00469369$ & $[0.00362764,\, 0.00575975]$ & $0.224877$ & $1.47917 \times 10^{-17}$ & $1.00584 \times 10^{-15}$ & \textbf{Favors left} \\
     & RandomLogit & $\downarrow$ & $1480$ & $-0.0283665$ & $[-0.0316507,\, -0.0250823]$ & $-0.440405$ & $5.51571 \times 10^{-59}$ & $4.68835 \times 10^{-57}$ & \textbf{Favors left} \\
     & DeletionAUC & $\downarrow$ & $1480$ & $-0.00302733$ & $[-0.00694358,\, 0.000888918]$ & $-0.0394151$ & $0.12965$ & $1$ & Not detected \\
     & InsertionAUC & $\uparrow$ & $1480$ & $-0.00915372$ & $[-0.0128,\, -0.00550741]$ & $-0.128002$ & $9.41457 \times 10^{-7}$ & $4.14241 \times 10^{-5}$ & \underline{Favors right} \\
    \midrule
    \multirow{9}{*}{Oxford Pets $\times$ ResNet-50} & FaithfulnessCorrelation ($\mathrm{FC}$) & $\uparrow$ & $1480$ & $-0.000840882$ & $[-0.0059092,\, 0.00422744]$ & $-0.00845948$ & $0.744892$ & $1$ & Not detected \\
     & MonotonicityCorrelation ($\mathrm{MC}$) & $\uparrow$ & $1480$ & $-0.0377562$ & $[-0.0477721,\, -0.0277403]$ & $-0.192208$ & $2.36842 \times 10^{-13}$ & $1.44474 \times 10^{-11}$ & \underline{Favors right} \\
     & FaithfulnessEstimate ($\mathrm{FE}$) & $\uparrow$ & $1480$ & $0.00664963$ & $[-0.00374057,\, 0.0170398]$ & $0.0326322$ & $0.209537$ & $1$ & Not detected \\
     & PointingGame ($\mathrm{PG}$) & $\uparrow$ & $1477$ & $0.00541638$ & $[-0.0166531,\, 0.0274858]$ & $0.0125266$ & $0.630291$ & $1$ & Not detected \\
     & TopKIntersection ($\mathrm{TopK}$) & $\uparrow$ & $1477$ & $0.0157955$ & $[0.0129085,\, 0.0186826]$ & $0.279249$ & $6.33091 \times 10^{-26}$ & $4.74818 \times 10^{-24}$ & \textbf{Favors left} \\
     & RelevanceRankAccuracy ($\mathrm{RRA}$) & $\uparrow$ & $1477$ & $0.00476358$ & $[0.00388594,\, 0.00564121]$ & $0.277034$ & $1.4836 \times 10^{-25}$ & $1.09786 \times 10^{-23}$ & \textbf{Favors left} \\
     & RandomLogit & $\downarrow$ & $1480$ & $-0.0607386$ & $[-0.0640739,\, -0.0574033]$ & $-0.928543$ & $5.01201 \times 10^{-202}$ & $4.81153 \times 10^{-200}$ & \textbf{Favors left} \\
     & DeletionAUC & $\downarrow$ & $1480$ & $-0.00139714$ & $[-0.00317712,\, 0.000382847]$ & $-0.0400217$ & $0.123856$ & $1$ & Not detected \\
     & InsertionAUC & $\uparrow$ & $1480$ & $0.00272423$ & $[6.54827e-06,\, 0.00544191]$ & $0.0511114$ & $0.0494512$ & $1$ & Not detected \\
    \bottomrule
  \end{tabular}%
  }
\end{table*}

\clearpage

\begin{table*}[t]
  \centering
  \caption{Complete paired Student $t$-test results for contrast $\mathbf{P-C}$ (Permuted Baseline $-$ Linear Chord; Coupling ablation ($\coupl{\hat{\pi}_{h_\theta}} \to \coupl{\hat{\pi}_{\mathrm{perm}}}$, Hypothesis~\circnumB{B})) on generator \textbf{Flow-IG (Diffusion)}. Differences and Cohen's $d_z$ are raw $\text{left} - \text{right}$ (positive favors left for $\uparrow$ metrics; negative favors left for $\downarrow$ metrics). Multiplicity adjustment uses Holm's step-down procedure across the $m=96$ applicable tests of family $\mathbf{P-C}$. Tally across the $48$ applicable tests for this generator: $\mathrm{L}=17$ Favors left, $\mathrm{ND}=28$ Not detected, $\mathrm{R}=3$ Favors right ($\mathrm{NA}=6$ Flowers-102 localization exclusions).}
  \label{tab:paired-pc-diffusion}
  \scriptsize
  \setlength{\tabcolsep}{4pt}
  \renewcommand{\arraystretch}{0.96}
  \resizebox{\textwidth}{!}{%
  \begin{tabular}{ll c r r c r r r l}
    \toprule
    \textbf{Dataset $\times$ Backbone} & \textbf{Metric} & \textbf{Pol.} & $n$ & \textbf{Mean diff.\ ($\overline{\Delta}$)} & \textbf{95\% CI (unadjusted)} & \textbf{Cohen $d_z$} & $p_{\mathrm{raw}}$ & $p_{\mathrm{Holm}}$ & \textbf{Verdict} \\
    \midrule
    \multirow{9}{*}{Flowers-102 $\times$ PVT-v2-B2} & FaithfulnessCorrelation ($\mathrm{FC}$) & $\uparrow$ & $1020$ & $-0.015804$ & $[-0.0240293,\, -0.00757875]$ & $-0.118054$ & $0.000172392$ & $0.0106883$ & \underline{Favors right} \\
     & MonotonicityCorrelation ($\mathrm{MC}$) & $\uparrow$ & $1020$ & $0.0070765$ & $[-0.00884196,\, 0.022995]$ & $0.0273138$ & $0.383233$ & $1$ & Not detected \\
     & FaithfulnessEstimate ($\mathrm{FE}$) & $\uparrow$ & $1020$ & $-0.049213$ & $[-0.0677349,\, -0.030691]$ & $-0.163251$ & $2.23939 \times 10^{-7}$ & $1.67954 \times 10^{-5}$ & \underline{Favors right} \\
     & PointingGame ($\mathrm{PG}$) & $\uparrow$ & \textnormal{---} & \textnormal{---} & $[\textnormal{---},\, \textnormal{---}]$ & \textnormal{---} & \textnormal{---} & \textnormal{---} & \textnormal{Not applicable} \\
     & TopKIntersection ($\mathrm{TopK}$) & $\uparrow$ & \textnormal{---} & \textnormal{---} & $[\textnormal{---},\, \textnormal{---}]$ & \textnormal{---} & \textnormal{---} & \textnormal{---} & \textnormal{Not applicable} \\
     & RelevanceRankAccuracy ($\mathrm{RRA}$) & $\uparrow$ & \textnormal{---} & \textnormal{---} & $[\textnormal{---},\, \textnormal{---}]$ & \textnormal{---} & \textnormal{---} & \textnormal{---} & \textnormal{Not applicable} \\
     & RandomLogit & $\downarrow$ & $1020$ & $-0.0353712$ & $[-0.0460136,\, -0.0247288]$ & $-0.204209$ & $1.09098 \times 10^{-10}$ & $8.946 \times 10^{-9}$ & \textbf{Favors left} \\
     & DeletionAUC & $\downarrow$ & $1020$ & $-0.00718397$ & $[-0.0106339,\, -0.00373408]$ & $-0.127945$ & $4.72883 \times 10^{-5}$ & $0.00307374$ & \textbf{Favors left} \\
     & InsertionAUC & $\uparrow$ & $1020$ & $0.0110491$ & $[0.00674895,\, 0.0153492]$ & $0.157874$ & $5.4481 \times 10^{-7}$ & $3.81367 \times 10^{-5}$ & \textbf{Favors left} \\
    \midrule
    \multirow{9}{*}{Flowers-102 $\times$ ResNet-50} & FaithfulnessCorrelation ($\mathrm{FC}$) & $\uparrow$ & $1020$ & $-0.000281008$ & $[-0.00878006,\, 0.00821805]$ & $-0.00203148$ & $0.948282$ & $1$ & Not detected \\
     & MonotonicityCorrelation ($\mathrm{MC}$) & $\uparrow$ & $1020$ & $0.03529$ & $[0.0219659,\, 0.0486141]$ & $0.162734$ & $2.44208 \times 10^{-7}$ & $1.80714 \times 10^{-5}$ & \textbf{Favors left} \\
     & FaithfulnessEstimate ($\mathrm{FE}$) & $\uparrow$ & $1020$ & $0.0229142$ & $[0.0082967,\, 0.0375317]$ & $0.0963154$ & $0.00215346$ & $0.11198$ & Not detected \\
     & PointingGame ($\mathrm{PG}$) & $\uparrow$ & \textnormal{---} & \textnormal{---} & $[\textnormal{---},\, \textnormal{---}]$ & \textnormal{---} & \textnormal{---} & \textnormal{---} & \textnormal{Not applicable} \\
     & TopKIntersection ($\mathrm{TopK}$) & $\uparrow$ & \textnormal{---} & \textnormal{---} & $[\textnormal{---},\, \textnormal{---}]$ & \textnormal{---} & \textnormal{---} & \textnormal{---} & \textnormal{Not applicable} \\
     & RelevanceRankAccuracy ($\mathrm{RRA}$) & $\uparrow$ & \textnormal{---} & \textnormal{---} & $[\textnormal{---},\, \textnormal{---}]$ & \textnormal{---} & \textnormal{---} & \textnormal{---} & \textnormal{Not applicable} \\
     & RandomLogit & $\downarrow$ & $1020$ & $-0.032114$ & $[-0.0378001,\, -0.0264279]$ & $-0.347012$ & $4.9674 \times 10^{-27}$ & $4.66936 \times 10^{-25}$ & \textbf{Favors left} \\
     & DeletionAUC & $\downarrow$ & $1020$ & $-0.0073879$ & $[-0.00883342,\, -0.00594238]$ & $-0.314023$ & $1.22313 \times 10^{-22}$ & $1.12528 \times 10^{-20}$ & \textbf{Favors left} \\
     & InsertionAUC & $\uparrow$ & $1020$ & $-0.00134415$ & $[-0.00292794,\, 0.000239652]$ & $-0.0521448$ & $0.0961463$ & $1$ & Not detected \\
    \midrule
    \multirow{9}{*}{ImageNet $\times$ PVT-v2-B2} & FaithfulnessCorrelation ($\mathrm{FC}$) & $\uparrow$ & $1000$ & $0.00240267$ & $[-0.00664217,\, 0.0114475]$ & $0.0164842$ & $0.60229$ & $1$ & Not detected \\
     & MonotonicityCorrelation ($\mathrm{MC}$) & $\uparrow$ & $1000$ & $0.0285476$ & $[0.011165,\, 0.0459302]$ & $0.101913$ & $0.00131066$ & $0.0720865$ & Not detected \\
     & FaithfulnessEstimate ($\mathrm{FE}$) & $\uparrow$ & $1000$ & $0.0292681$ & $[0.00980564,\, 0.0487306]$ & $0.0933192$ & $0.0032413$ & $0.162065$ & Not detected \\
     & PointingGame ($\mathrm{PG}$) & $\uparrow$ & $1000$ & $0.014$ & $[-0.018009,\, 0.046009]$ & $0.0271413$ & $0.390943$ & $1$ & Not detected \\
     & TopKIntersection ($\mathrm{TopK}$) & $\uparrow$ & $1000$ & $0.015315$ & $[0.0050019,\, 0.0256281]$ & $0.0921515$ & $0.00364687$ & $0.178696$ & Not detected \\
     & RelevanceRankAccuracy ($\mathrm{RRA}$) & $\uparrow$ & $1000$ & $0.00358599$ & $[0.00103764,\, 0.00613433]$ & $0.0873223$ & $0.00586139$ & $0.269624$ & Not detected \\
     & RandomLogit & $\downarrow$ & $1000$ & $-0.0301001$ & $[-0.0409662,\, -0.0192341]$ & $-0.171898$ & $6.85297 \times 10^{-8}$ & $5.27679 \times 10^{-6}$ & \textbf{Favors left} \\
     & DeletionAUC & $\downarrow$ & $1000$ & $-0.00907858$ & $[-0.0137138,\, -0.00444339]$ & $-0.121542$ & $0.000128963$ & $0.00825363$ & \textbf{Favors left} \\
     & InsertionAUC & $\uparrow$ & $1000$ & $0.00508523$ & $[-0.000518267,\, 0.0106887]$ & $0.0563153$ & $0.0752416$ & $1$ & Not detected \\
    \midrule
    \multirow{9}{*}{ImageNet $\times$ ResNet-50} & FaithfulnessCorrelation ($\mathrm{FC}$) & $\uparrow$ & $1000$ & $0.00752295$ & $[-0.00111685,\, 0.0161628]$ & $0.054033$ & $0.0878222$ & $1$ & Not detected \\
     & MonotonicityCorrelation ($\mathrm{MC}$) & $\uparrow$ & $1000$ & $0.0122416$ & $[-0.0023105,\, 0.0267937]$ & $0.052202$ & $0.0990993$ & $1$ & Not detected \\
     & FaithfulnessEstimate ($\mathrm{FE}$) & $\uparrow$ & $1000$ & $0.0061025$ & $[-0.0120278,\, 0.0242328]$ & $0.020887$ & $0.509082$ & $1$ & Not detected \\
     & PointingGame ($\mathrm{PG}$) & $\uparrow$ & $1000$ & $0.03$ & $[0.0053918,\, 0.0546082]$ & $0.0756512$ & $0.0169271$ & $0.710936$ & Not detected \\
     & TopKIntersection ($\mathrm{TopK}$) & $\uparrow$ & $1000$ & $0.018321$ & $[0.00765281,\, 0.0289892]$ & $0.10657$ & $0.000780372$ & $0.0444812$ & \textbf{Favors left} \\
     & RelevanceRankAccuracy ($\mathrm{RRA}$) & $\uparrow$ & $1000$ & $-0.000608507$ & $[-0.0032937,\, 0.00207669]$ & $-0.0140626$ & $0.656635$ & $1$ & Not detected \\
     & RandomLogit & $\downarrow$ & $1000$ & $-0.0244129$ & $[-0.0337301,\, -0.0150957]$ & $-0.162596$ & $3.27455 \times 10^{-7}$ & $2.39042 \times 10^{-5}$ & \textbf{Favors left} \\
     & DeletionAUC & $\downarrow$ & $1000$ & $-0.00514389$ & $[-0.00713865,\, -0.00314913]$ & $-0.16002$ & $4.9809 \times 10^{-7}$ & $3.53644 \times 10^{-5}$ & \textbf{Favors left} \\
     & InsertionAUC & $\uparrow$ & $1000$ & $0.0109946$ & $[0.00782248,\, 0.0141668]$ & $0.21508$ & $1.78108 \times 10^{-11}$ & $1.4783 \times 10^{-9}$ & \textbf{Favors left} \\
    \midrule
    \multirow{9}{*}{Oxford Pets $\times$ PVT-v2-B2} & FaithfulnessCorrelation ($\mathrm{FC}$) & $\uparrow$ & $1480$ & $0.00421341$ & $[-0.0030856,\, 0.0115124]$ & $0.0294335$ & $0.257679$ & $1$ & Not detected \\
     & MonotonicityCorrelation ($\mathrm{MC}$) & $\uparrow$ & $1479$ & $0.0183797$ & $[0.00632422,\, 0.0304353]$ & $0.077763$ & $0.00283043$ & $0.144352$ & Not detected \\
     & FaithfulnessEstimate ($\mathrm{FE}$) & $\uparrow$ & $1480$ & $0.00269393$ & $[-0.0115207,\, 0.0169085]$ & $0.00966327$ & $0.71013$ & $1$ & Not detected \\
     & PointingGame ($\mathrm{PG}$) & $\uparrow$ & $1475$ & $0.0169492$ & $[-0.0113521,\, 0.0452504]$ & $0.030588$ & $0.240283$ & $1$ & Not detected \\
     & TopKIntersection ($\mathrm{TopK}$) & $\uparrow$ & $1475$ & $0.00929966$ & $[0.000404719,\, 0.0181946]$ & $0.053399$ & $0.0404604$ & $1$ & Not detected \\
     & RelevanceRankAccuracy ($\mathrm{RRA}$) & $\uparrow$ & $1475$ & $-0.00536657$ & $[-0.00787247,\, -0.00286066]$ & $-0.109381$ & $2.81867 \times 10^{-5}$ & $0.00186032$ & \underline{Favors right} \\
     & RandomLogit & $\downarrow$ & $1480$ & $-0.0121352$ & $[-0.0206202,\, -0.00365013]$ & $-0.072923$ & $0.00509114$ & $0.244375$ & Not detected \\
     & DeletionAUC & $\downarrow$ & $1480$ & $-0.0239482$ & $[-0.029629,\, -0.0182675]$ & $-0.214952$ & $2.96555 \times 10^{-16}$ & $2.63934 \times 10^{-14}$ & \textbf{Favors left} \\
     & InsertionAUC & $\uparrow$ & $1480$ & $0.0227319$ & $[0.0174882,\, 0.0279757]$ & $0.22104$ & $4.43109 \times 10^{-17}$ & $3.98798 \times 10^{-15}$ & \textbf{Favors left} \\
    \midrule
    \multirow{9}{*}{Oxford Pets $\times$ ResNet-50} & FaithfulnessCorrelation ($\mathrm{FC}$) & $\uparrow$ & $1480$ & $-0.00137824$ & $[-0.0084263,\, 0.00566982]$ & $-0.00997074$ & $0.701343$ & $1$ & Not detected \\
     & MonotonicityCorrelation ($\mathrm{MC}$) & $\uparrow$ & $1480$ & $0.0318948$ & $[0.0210783,\, 0.0427112]$ & $0.150351$ & $8.8797 \times 10^{-9}$ & $7.01497 \times 10^{-7}$ & \textbf{Favors left} \\
     & FaithfulnessEstimate ($\mathrm{FE}$) & $\uparrow$ & $1480$ & $0.0232794$ & $[0.00973604,\, 0.0368227]$ & $0.0876432$ & $0.000766319$ & $0.0444465$ & \textbf{Favors left} \\
     & PointingGame ($\mathrm{PG}$) & $\uparrow$ & $1477$ & $0.0189573$ & $[-0.0063011,\, 0.0442158]$ & $0.0383076$ & $0.141172$ & $1$ & Not detected \\
     & TopKIntersection ($\mathrm{TopK}$) & $\uparrow$ & $1477$ & $0.0124299$ & $[0.00487274,\, 0.0199871]$ & $0.0839503$ & $0.00128125$ & $0.0717502$ & Not detected \\
     & RelevanceRankAccuracy ($\mathrm{RRA}$) & $\uparrow$ & $1477$ & $0.000415431$ & $[-0.0014468,\, 0.00227766]$ & $0.0113862$ & $0.661745$ & $1$ & Not detected \\
     & RandomLogit & $\downarrow$ & $1480$ & $-0.00309215$ & $[-0.00769422,\, 0.00150991]$ & $-0.0342595$ & $0.187711$ & $1$ & Not detected \\
     & DeletionAUC & $\downarrow$ & $1480$ & $-0.00867656$ & $[-0.0109376,\, -0.00641552]$ & $-0.195664$ & $8.9551 \times 10^{-14}$ & $7.70138 \times 10^{-12}$ & \textbf{Favors left} \\
     & InsertionAUC & $\uparrow$ & $1480$ & $0.00449101$ & $[0.00122584,\, 0.00775618]$ & $0.0701311$ & $0.00705512$ & $0.31748$ & Not detected \\
    \bottomrule
  \end{tabular}%
  }
\end{table*}

\begin{table*}[t]
  \centering
  \caption{Complete paired Student $t$-test results for contrast $\mathbf{P-C}$ (Permuted Baseline $-$ Linear Chord; Coupling ablation ($\coupl{\hat{\pi}_{h_\theta}} \to \coupl{\hat{\pi}_{\mathrm{perm}}}$, Hypothesis~\circnumB{B})) on generator \textbf{Flow-IG (PixelDiT)}. Differences and Cohen's $d_z$ are raw $\text{left} - \text{right}$ (positive favors left for $\uparrow$ metrics; negative favors left for $\downarrow$ metrics). Multiplicity adjustment uses Holm's step-down procedure across the $m=96$ applicable tests of family $\mathbf{P-C}$. Tally across the $48$ applicable tests for this generator: $\mathrm{L}=16$ Favors left, $\mathrm{ND}=28$ Not detected, $\mathrm{R}=4$ Favors right ($\mathrm{NA}=6$ Flowers-102 localization exclusions).}
  \label{tab:paired-pc-pixeldit}
  \scriptsize
  \setlength{\tabcolsep}{4pt}
  \renewcommand{\arraystretch}{0.96}
  \resizebox{\textwidth}{!}{%
  \begin{tabular}{ll c r r c r r r l}
    \toprule
    \textbf{Dataset $\times$ Backbone} & \textbf{Metric} & \textbf{Pol.} & $n$ & \textbf{Mean diff.\ ($\overline{\Delta}$)} & \textbf{95\% CI (unadjusted)} & \textbf{Cohen $d_z$} & $p_{\mathrm{raw}}$ & $p_{\mathrm{Holm}}$ & \textbf{Verdict} \\
    \midrule
    \multirow{9}{*}{Flowers-102 $\times$ PVT-v2-B2} & FaithfulnessCorrelation ($\mathrm{FC}$) & $\uparrow$ & $1020$ & $-0.0150613$ & $[-0.0234055,\, -0.00671721]$ & $-0.110904$ & $0.000415101$ & $0.0249061$ & \underline{Favors right} \\
     & MonotonicityCorrelation ($\mathrm{MC}$) & $\uparrow$ & $1020$ & $-0.0205784$ & $[-0.0351417,\, -0.00601498]$ & $-0.0868186$ & $0.00565964$ & $0.266003$ & Not detected \\
     & FaithfulnessEstimate ($\mathrm{FE}$) & $\uparrow$ & $1020$ & $-0.0592843$ & $[-0.0748193,\, -0.0437494]$ & $-0.234474$ & $1.50608 \times 10^{-13}$ & $1.28017 \times 10^{-11}$ & \underline{Favors right} \\
     & PointingGame ($\mathrm{PG}$) & $\uparrow$ & \textnormal{---} & \textnormal{---} & $[\textnormal{---},\, \textnormal{---}]$ & \textnormal{---} & \textnormal{---} & \textnormal{---} & \textnormal{Not applicable} \\
     & TopKIntersection ($\mathrm{TopK}$) & $\uparrow$ & \textnormal{---} & \textnormal{---} & $[\textnormal{---},\, \textnormal{---}]$ & \textnormal{---} & \textnormal{---} & \textnormal{---} & \textnormal{Not applicable} \\
     & RelevanceRankAccuracy ($\mathrm{RRA}$) & $\uparrow$ & \textnormal{---} & \textnormal{---} & $[\textnormal{---},\, \textnormal{---}]$ & \textnormal{---} & \textnormal{---} & \textnormal{---} & \textnormal{Not applicable} \\
     & RandomLogit & $\downarrow$ & $1020$ & $-0.0280505$ & $[-0.0333796,\, -0.0227213]$ & $-0.323404$ & $7.47931 \times 10^{-24}$ & $6.95576 \times 10^{-22}$ & \textbf{Favors left} \\
     & DeletionAUC & $\downarrow$ & $1020$ & $-0.00946945$ & $[-0.0128711,\, -0.00606783]$ & $-0.171042$ & $5.89542 \times 10^{-8}$ & $4.59843 \times 10^{-6}$ & \textbf{Favors left} \\
     & InsertionAUC & $\uparrow$ & $1020$ & $0.0132663$ & $[0.00923628,\, 0.0172964]$ & $0.202257$ & $1.62338 \times 10^{-10}$ & $1.31494 \times 10^{-8}$ & \textbf{Favors left} \\
    \midrule
    \multirow{9}{*}{Flowers-102 $\times$ ResNet-50} & FaithfulnessCorrelation ($\mathrm{FC}$) & $\uparrow$ & $1020$ & $-0.00100192$ & $[-0.00947632,\, 0.00747248]$ & $-0.00726421$ & $0.816584$ & $1$ & Not detected \\
     & MonotonicityCorrelation ($\mathrm{MC}$) & $\uparrow$ & $1020$ & $0.020849$ & $[0.0100414,\, 0.0316565]$ & $0.118528$ & $0.000162368$ & $0.0102292$ & \textbf{Favors left} \\
     & FaithfulnessEstimate ($\mathrm{FE}$) & $\uparrow$ & $1020$ & $0.0229452$ & $[0.0105056,\, 0.0353849]$ & $0.113331$ & $0.000309681$ & $0.0188906$ & \textbf{Favors left} \\
     & PointingGame ($\mathrm{PG}$) & $\uparrow$ & \textnormal{---} & \textnormal{---} & $[\textnormal{---},\, \textnormal{---}]$ & \textnormal{---} & \textnormal{---} & \textnormal{---} & \textnormal{Not applicable} \\
     & TopKIntersection ($\mathrm{TopK}$) & $\uparrow$ & \textnormal{---} & \textnormal{---} & $[\textnormal{---},\, \textnormal{---}]$ & \textnormal{---} & \textnormal{---} & \textnormal{---} & \textnormal{Not applicable} \\
     & RelevanceRankAccuracy ($\mathrm{RRA}$) & $\uparrow$ & \textnormal{---} & \textnormal{---} & $[\textnormal{---},\, \textnormal{---}]$ & \textnormal{---} & \textnormal{---} & \textnormal{---} & \textnormal{Not applicable} \\
     & RandomLogit & $\downarrow$ & $1020$ & $-0.0323246$ & $[-0.0367767,\, -0.0278725]$ & $-0.446097$ & $3.89594 \times 10^{-42}$ & $3.7401 \times 10^{-40}$ & \textbf{Favors left} \\
     & DeletionAUC & $\downarrow$ & $1020$ & $-0.00635538$ & $[-0.00777437,\, -0.00493639]$ & $-0.275186$ & $6.32008 \times 10^{-18}$ & $5.75127 \times 10^{-16}$ & \textbf{Favors left} \\
     & InsertionAUC & $\uparrow$ & $1020$ & $3.01635e-05$ & $[-0.00142714,\, 0.00148747]$ & $0.00127173$ & $0.96761$ & $1$ & Not detected \\
    \midrule
    \multirow{9}{*}{ImageNet $\times$ PVT-v2-B2} & FaithfulnessCorrelation ($\mathrm{FC}$) & $\uparrow$ & $1000$ & $-0.00129886$ & $[-0.0101737,\, 0.00757597]$ & $-0.00908191$ & $0.774022$ & $1$ & Not detected \\
     & MonotonicityCorrelation ($\mathrm{MC}$) & $\uparrow$ & $1000$ & $0.0193217$ & $[0.00387541,\, 0.034768]$ & $0.0776239$ & $0.0142705$ & $0.61363$ & Not detected \\
     & FaithfulnessEstimate ($\mathrm{FE}$) & $\uparrow$ & $1000$ & $0.0117593$ & $[-0.00522594,\, 0.0287445]$ & $0.0429619$ & $0.174588$ & $1$ & Not detected \\
     & PointingGame ($\mathrm{PG}$) & $\uparrow$ & $1000$ & $-0.002$ & $[-0.0336574,\, 0.0296574]$ & $-0.00392039$ & $0.901361$ & $1$ & Not detected \\
     & TopKIntersection ($\mathrm{TopK}$) & $\uparrow$ & $1000$ & $0.012405$ & $[0.00470561,\, 0.0201044]$ & $0.0999805$ & $0.00161599$ & $0.0856476$ & Not detected \\
     & RelevanceRankAccuracy ($\mathrm{RRA}$) & $\uparrow$ & $1000$ & $0.00130142$ & $[-0.000696432,\, 0.00329926]$ & $0.040423$ & $0.201445$ & $1$ & Not detected \\
     & RandomLogit & $\downarrow$ & $1000$ & $-0.00952113$ & $[-0.0173915,\, -0.00165072]$ & $-0.0750699$ & $0.0177887$ & $0.729337$ & Not detected \\
     & DeletionAUC & $\downarrow$ & $1000$ & $-0.0109942$ & $[-0.0154207,\, -0.00656765]$ & $-0.154126$ & $1.27207 \times 10^{-6}$ & $8.65006 \times 10^{-5}$ & \textbf{Favors left} \\
     & InsertionAUC & $\uparrow$ & $1000$ & $0.00138249$ & $[-0.00421808,\, 0.00698306]$ & $0.0153181$ & $0.628209$ & $1$ & Not detected \\
    \midrule
    \multirow{9}{*}{ImageNet $\times$ ResNet-50} & FaithfulnessCorrelation ($\mathrm{FC}$) & $\uparrow$ & $1000$ & $0.00749766$ & $[-0.00149378,\, 0.0164891]$ & $0.0517453$ & $0.102085$ & $1$ & Not detected \\
     & MonotonicityCorrelation ($\mathrm{MC}$) & $\uparrow$ & $1000$ & $0.00935981$ & $[-0.0039051,\, 0.0226247]$ & $0.0437862$ & $0.166471$ & $1$ & Not detected \\
     & FaithfulnessEstimate ($\mathrm{FE}$) & $\uparrow$ & $1000$ & $-0.00628281$ & $[-0.0226696,\, 0.010104]$ & $-0.0237922$ & $0.452002$ & $1$ & Not detected \\
     & PointingGame ($\mathrm{PG}$) & $\uparrow$ & $1000$ & $0$ & $[-0.0198286,\, 0.0198286]$ & $0$ & $1$ & $1$ & Not detected \\
     & TopKIntersection ($\mathrm{TopK}$) & $\uparrow$ & $1000$ & $0.004353$ & $[0.000240757,\, 0.00846524]$ & $0.0656878$ & $0.0380354$ & $1$ & Not detected \\
     & RelevanceRankAccuracy ($\mathrm{RRA}$) & $\uparrow$ & $1000$ & $-0.00175433$ & $[-0.00327403,\, -0.000234636]$ & $-0.0716357$ & $0.0237064$ & $0.924551$ & Not detected \\
     & RandomLogit & $\downarrow$ & $1000$ & $-0.0149534$ & $[-0.0206769,\, -0.00922992]$ & $-0.162126$ & $3.53637 \times 10^{-7}$ & $2.54619 \times 10^{-5}$ & \textbf{Favors left} \\
     & DeletionAUC & $\downarrow$ & $1000$ & $-0.00625723$ & $[-0.0081727,\, -0.00434177]$ & $-0.202713$ & $2.23457 \times 10^{-10}$ & $1.78766 \times 10^{-8}$ & \textbf{Favors left} \\
     & InsertionAUC & $\uparrow$ & $1000$ & $0.0133988$ & $[0.0100341,\, 0.0167634]$ & $0.247113$ & $1.39433 \times 10^{-14}$ & $1.21307 \times 10^{-12}$ & \textbf{Favors left} \\
    \midrule
    \multirow{9}{*}{Oxford Pets $\times$ PVT-v2-B2} & FaithfulnessCorrelation ($\mathrm{FC}$) & $\uparrow$ & $1480$ & $0.00428527$ & $[-0.00279061,\, 0.0113611]$ & $0.0308795$ & $0.235041$ & $1$ & Not detected \\
     & MonotonicityCorrelation ($\mathrm{MC}$) & $\uparrow$ & $1479$ & $0.0100046$ & $[-0.000792557,\, 0.0208017]$ & $0.0472618$ & $0.0693303$ & $1$ & Not detected \\
     & FaithfulnessEstimate ($\mathrm{FE}$) & $\uparrow$ & $1480$ & $-0.002289$ & $[-0.0130093,\, 0.0084313]$ & $-0.0108871$ & $0.675398$ & $1$ & Not detected \\
     & PointingGame ($\mathrm{PG}$) & $\uparrow$ & $1475$ & $-0.00949153$ & $[-0.0365554,\, 0.0175724]$ & $-0.0179124$ & $0.491599$ & $1$ & Not detected \\
     & TopKIntersection ($\mathrm{TopK}$) & $\uparrow$ & $1475$ & $0.000677288$ & $[-0.00503757,\, 0.00639215]$ & $0.00605308$ & $0.816203$ & $1$ & Not detected \\
     & RelevanceRankAccuracy ($\mathrm{RRA}$) & $\uparrow$ & $1475$ & $-0.00605186$ & $[-0.0077857,\, -0.00431803]$ & $-0.178275$ & $1.10507 \times 10^{-11}$ & $9.28256 \times 10^{-10}$ & \underline{Favors right} \\
     & RandomLogit & $\downarrow$ & $1480$ & $-0.0116098$ & $[-0.0161703,\, -0.00704928]$ & $-0.129803$ & $6.62925 \times 10^{-7}$ & $4.57419 \times 10^{-5}$ & \textbf{Favors left} \\
     & DeletionAUC & $\downarrow$ & $1480$ & $-0.0317498$ & $[-0.0373693,\, -0.0261304]$ & $-0.288085$ & $1.77309 \times 10^{-27}$ & $1.68444 \times 10^{-25}$ & \textbf{Favors left} \\
     & InsertionAUC & $\uparrow$ & $1480$ & $0.0205569$ & $[0.0155403,\, 0.0255735]$ & $0.208938$ & $1.85148 \times 10^{-15}$ & $1.6293 \times 10^{-13}$ & \textbf{Favors left} \\
    \midrule
    \multirow{9}{*}{Oxford Pets $\times$ ResNet-50} & FaithfulnessCorrelation ($\mathrm{FC}$) & $\uparrow$ & $1480$ & $0.000855279$ & $[-0.00637263,\, 0.00808319]$ & $0.00603348$ & $0.816483$ & $1$ & Not detected \\
     & MonotonicityCorrelation ($\mathrm{MC}$) & $\uparrow$ & $1480$ & $0.0173406$ & $[0.0067661,\, 0.0279151]$ & $0.0836137$ & $0.00132488$ & $0.0720865$ & Not detected \\
     & FaithfulnessEstimate ($\mathrm{FE}$) & $\uparrow$ & $1480$ & $0.0136915$ & $[0.00158592,\, 0.0257971]$ & $0.0576684$ & $0.0266688$ & $1$ & Not detected \\
     & PointingGame ($\mathrm{PG}$) & $\uparrow$ & $1477$ & $0.0162492$ & $[-0.00720267,\, 0.039701]$ & $0.0353646$ & $0.174316$ & $1$ & Not detected \\
     & TopKIntersection ($\mathrm{TopK}$) & $\uparrow$ & $1477$ & $-0.00442045$ & $[-0.00823238,\, -0.000608517]$ & $-0.0591882$ & $0.0230664$ & $0.922655$ & Not detected \\
     & RelevanceRankAccuracy ($\mathrm{RRA}$) & $\uparrow$ & $1477$ & $-0.00263409$ & $[-0.00382327,\, -0.0014449]$ & $-0.113056$ & $1.48821 \times 10^{-5}$ & $0.000997103$ & \underline{Favors right} \\
     & RandomLogit & $\downarrow$ & $1480$ & $-0.00685716$ & $[-0.0106833,\, -0.00303105]$ & $-0.0913817$ & $0.000452135$ & $0.026676$ & \textbf{Favors left} \\
     & DeletionAUC & $\downarrow$ & $1480$ & $-0.00525811$ & $[-0.00723181,\, -0.00328441]$ & $-0.135838$ & $1.9818 \times 10^{-7}$ & $1.50617 \times 10^{-5}$ & \textbf{Favors left} \\
     & InsertionAUC & $\uparrow$ & $1480$ & $0.00399657$ & $[0.000824058,\, 0.00716909]$ & $0.0642328$ & $0.0135823$ & $0.597623$ & Not detected \\
    \bottomrule
  \end{tabular}%
  }
\end{table*}

\clearpage

\begin{table*}[t]
  \centering
  \caption{Complete paired Student $t$-test results for contrast $\mathbf{P-F}$ (Permuted Baseline $-$ Flow-IG; Composite path-deviation + coupling ablation ($P$ vs.\ $F$)) on generator \textbf{Flow-IG (Diffusion)}. Differences and Cohen's $d_z$ are raw $\text{left} - \text{right}$ (positive favors left for $\uparrow$ metrics; negative favors left for $\downarrow$ metrics). Multiplicity adjustment uses Holm's step-down procedure across the $m=96$ applicable tests of family $\mathbf{P-F}$. Tally across the $48$ applicable tests for this generator: $\mathrm{L}=36$ Favors left, $\mathrm{ND}=10$ Not detected, $\mathrm{R}=2$ Favors right ($\mathrm{NA}=6$ Flowers-102 localization exclusions).}
  \label{tab:paired-pf-diffusion}
  \scriptsize
  \setlength{\tabcolsep}{4pt}
  \renewcommand{\arraystretch}{0.96}
  \resizebox{\textwidth}{!}{%
  \begin{tabular}{ll c r r c r r r l}
    \toprule
    \textbf{Dataset $\times$ Backbone} & \textbf{Metric} & \textbf{Pol.} & $n$ & \textbf{Mean diff.\ ($\overline{\Delta}$)} & \textbf{95\% CI (unadjusted)} & \textbf{Cohen $d_z$} & $p_{\mathrm{raw}}$ & $p_{\mathrm{Holm}}$ & \textbf{Verdict} \\
    \midrule
    \multirow{9}{*}{Flowers-102 $\times$ PVT-v2-B2} & FaithfulnessCorrelation ($\mathrm{FC}$) & $\uparrow$ & $1020$ & $0.0381201$ & $[0.0296391,\, 0.0466012]$ & $0.276165$ & $4.87784 \times 10^{-18}$ & $3.36571 \times 10^{-16}$ & \textbf{Favors left} \\
     & MonotonicityCorrelation ($\mathrm{MC}$) & $\uparrow$ & $1020$ & $-0.0110478$ & $[-0.0276442,\, 0.00554862]$ & $-0.0409002$ & $0.191763$ & $1$ & Not detected \\
     & FaithfulnessEstimate ($\mathrm{FE}$) & $\uparrow$ & $1020$ & $-0.0241721$ & $[-0.0426531,\, -0.00569112]$ & $-0.0803625$ & $0.0104128$ & $0.281145$ & Not detected \\
     & PointingGame ($\mathrm{PG}$) & $\uparrow$ & \textnormal{---} & \textnormal{---} & $[\textnormal{---},\, \textnormal{---}]$ & \textnormal{---} & \textnormal{---} & \textnormal{---} & \textnormal{Not applicable} \\
     & TopKIntersection ($\mathrm{TopK}$) & $\uparrow$ & \textnormal{---} & \textnormal{---} & $[\textnormal{---},\, \textnormal{---}]$ & \textnormal{---} & \textnormal{---} & \textnormal{---} & \textnormal{Not applicable} \\
     & RelevanceRankAccuracy ($\mathrm{RRA}$) & $\uparrow$ & \textnormal{---} & \textnormal{---} & $[\textnormal{---},\, \textnormal{---}]$ & \textnormal{---} & \textnormal{---} & \textnormal{---} & \textnormal{Not applicable} \\
     & RandomLogit & $\downarrow$ & $1020$ & $-0.0622751$ & $[-0.0721425,\, -0.0524078]$ & $-0.387774$ & $6.48329 \times 10^{-33}$ & $5.05697 \times 10^{-31}$ & \textbf{Favors left} \\
     & DeletionAUC & $\downarrow$ & $1020$ & $-0.0295565$ & $[-0.0334342,\, -0.0256789]$ & $-0.468325$ & $7.08905 \times 10^{-46}$ & $5.88391 \times 10^{-44}$ & \textbf{Favors left} \\
     & InsertionAUC & $\uparrow$ & $1020$ & $0.0256885$ & $[0.0207656,\, 0.0306114]$ & $0.320615$ & $1.7284 \times 10^{-23}$ & $1.24444 \times 10^{-21}$ & \textbf{Favors left} \\
    \midrule
    \multirow{9}{*}{Flowers-102 $\times$ ResNet-50} & FaithfulnessCorrelation ($\mathrm{FC}$) & $\uparrow$ & $1020$ & $0.0239514$ & $[0.0151636,\, 0.0327392]$ & $0.167462$ & $1.09614 \times 10^{-7}$ & $6.02875 \times 10^{-6}$ & \textbf{Favors left} \\
     & MonotonicityCorrelation ($\mathrm{MC}$) & $\uparrow$ & $1020$ & $0.0786968$ & $[0.0632709,\, 0.0941226]$ & $0.313452$ & $1.44653 \times 10^{-22}$ & $1.02704 \times 10^{-20}$ & \textbf{Favors left} \\
     & FaithfulnessEstimate ($\mathrm{FE}$) & $\uparrow$ & $1020$ & $0.12194$ & $[0.105669,\, 0.138211]$ & $0.460468$ & $1.53455 \times 10^{-44}$ & $1.25833 \times 10^{-42}$ & \textbf{Favors left} \\
     & PointingGame ($\mathrm{PG}$) & $\uparrow$ & \textnormal{---} & \textnormal{---} & $[\textnormal{---},\, \textnormal{---}]$ & \textnormal{---} & \textnormal{---} & \textnormal{---} & \textnormal{Not applicable} \\
     & TopKIntersection ($\mathrm{TopK}$) & $\uparrow$ & \textnormal{---} & \textnormal{---} & $[\textnormal{---},\, \textnormal{---}]$ & \textnormal{---} & \textnormal{---} & \textnormal{---} & \textnormal{Not applicable} \\
     & RelevanceRankAccuracy ($\mathrm{RRA}$) & $\uparrow$ & \textnormal{---} & \textnormal{---} & $[\textnormal{---},\, \textnormal{---}]$ & \textnormal{---} & \textnormal{---} & \textnormal{---} & \textnormal{Not applicable} \\
     & RandomLogit & $\downarrow$ & $1020$ & $-0.0908463$ & $[-0.0967362,\, -0.0849563]$ & $-0.947676$ & $4.51499 \times 10^{-144}$ & $4.24409 \times 10^{-142}$ & \textbf{Favors left} \\
     & DeletionAUC & $\downarrow$ & $1020$ & $-0.0168047$ & $[-0.0185725,\, -0.0150369]$ & $-0.584075$ & $4.89558 \times 10^{-67}$ & $4.25915 \times 10^{-65}$ & \textbf{Favors left} \\
     & InsertionAUC & $\uparrow$ & $1020$ & $0.00562164$ & $[0.00381175,\, 0.00743152]$ & $0.190843$ & $1.55005 \times 10^{-9}$ & $8.99028 \times 10^{-8}$ & \textbf{Favors left} \\
    \midrule
    \multirow{9}{*}{ImageNet $\times$ PVT-v2-B2} & FaithfulnessCorrelation ($\mathrm{FC}$) & $\uparrow$ & $1000$ & $0.0228194$ & $[0.0138245,\, 0.0318142]$ & $0.157429$ & $7.55033 \times 10^{-7}$ & $4.07718 \times 10^{-5}$ & \textbf{Favors left} \\
     & MonotonicityCorrelation ($\mathrm{MC}$) & $\uparrow$ & $1000$ & $0.0142245$ & $[-0.00286271,\, 0.0313118]$ & $0.0516583$ & $0.102662$ & $1$ & Not detected \\
     & FaithfulnessEstimate ($\mathrm{FE}$) & $\uparrow$ & $1000$ & $0.0357309$ & $[0.0161825,\, 0.0552793]$ & $0.113425$ & $0.000350968$ & $0.0136877$ & \textbf{Favors left} \\
     & PointingGame ($\mathrm{PG}$) & $\uparrow$ & $1000$ & $-0.02$ & $[-0.0518761,\, 0.0118761]$ & $-0.038935$ & $0.218526$ & $1$ & Not detected \\
     & TopKIntersection ($\mathrm{TopK}$) & $\uparrow$ & $1000$ & $-0.017693$ & $[-0.0282106,\, -0.00717536]$ & $-0.10439$ & $0.000997185$ & $0.0309127$ & \underline{Favors right} \\
     & RelevanceRankAccuracy ($\mathrm{RRA}$) & $\uparrow$ & $1000$ & $-0.00458864$ & $[-0.00718517,\, -0.0019921]$ & $-0.109664$ & $0.000546937$ & $0.0191428$ & \underline{Favors right} \\
     & RandomLogit & $\downarrow$ & $1000$ & $-0.0346136$ & $[-0.0453213,\, -0.0239059]$ & $-0.200596$ & $3.40037 \times 10^{-10}$ & $2.07423 \times 10^{-8}$ & \textbf{Favors left} \\
     & DeletionAUC & $\downarrow$ & $1000$ & $-0.0159248$ & $[-0.020852,\, -0.0109975]$ & $-0.200559$ & $3.42588 \times 10^{-10}$ & $2.07423 \times 10^{-8}$ & \textbf{Favors left} \\
     & InsertionAUC & $\uparrow$ & $1000$ & $0.0197007$ & $[0.0131211,\, 0.0262803]$ & $0.185804$ & $5.73329 \times 10^{-9}$ & $3.26798 \times 10^{-7}$ & \textbf{Favors left} \\
    \midrule
    \multirow{9}{*}{ImageNet $\times$ ResNet-50} & FaithfulnessCorrelation ($\mathrm{FC}$) & $\uparrow$ & $1000$ & $0.0370044$ & $[0.0283692,\, 0.0456397]$ & $0.265921$ & $1.41174 \times 10^{-16}$ & $9.59981 \times 10^{-15}$ & \textbf{Favors left} \\
     & MonotonicityCorrelation ($\mathrm{MC}$) & $\uparrow$ & $1000$ & $0.0225553$ & $[0.00747368,\, 0.0376369]$ & $0.0928058$ & $0.00341424$ & $0.102427$ & Not detected \\
     & FaithfulnessEstimate ($\mathrm{FE}$) & $\uparrow$ & $1000$ & $0.0390929$ & $[0.0203096,\, 0.0578762]$ & $0.129152$ & $4.778 \times 10^{-5}$ & $0.0021501$ & \textbf{Favors left} \\
     & PointingGame ($\mathrm{PG}$) & $\uparrow$ & $1000$ & $0.038$ & $[0.0122097,\, 0.0637903]$ & $0.0914327$ & $0.00391897$ & $0.109731$ & Not detected \\
     & TopKIntersection ($\mathrm{TopK}$) & $\uparrow$ & $1000$ & $0.019437$ & $[0.0083149,\, 0.0305591]$ & $0.108447$ & $0.000629655$ & $0.0214083$ & \textbf{Favors left} \\
     & RelevanceRankAccuracy ($\mathrm{RRA}$) & $\uparrow$ & $1000$ & $-0.00220991$ & $[-0.00500533,\, 0.000585502]$ & $-0.0490573$ & $0.12114$ & $1$ & Not detected \\
     & RandomLogit & $\downarrow$ & $1000$ & $-0.0764594$ & $[-0.0858936,\, -0.0670253]$ & $-0.502925$ & $6.14739 \times 10^{-51}$ & $5.22528 \times 10^{-49}$ & \textbf{Favors left} \\
     & DeletionAUC & $\downarrow$ & $1000$ & $-0.0124204$ & $[-0.0147493,\, -0.0100916]$ & $-0.330959$ & $2.14621 \times 10^{-24}$ & $1.60966 \times 10^{-22}$ & \textbf{Favors left} \\
     & InsertionAUC & $\uparrow$ & $1000$ & $0.0186159$ & $[0.0150924,\, 0.0221394]$ & $0.327859$ & $5.43637 \times 10^{-24}$ & $4.02291 \times 10^{-22}$ & \textbf{Favors left} \\
    \midrule
    \multirow{9}{*}{Oxford Pets $\times$ PVT-v2-B2} & FaithfulnessCorrelation ($\mathrm{FC}$) & $\uparrow$ & $1480$ & $0.0676964$ & $[0.060368,\, 0.0750248]$ & $0.471008$ & $1.98177 \times 10^{-66}$ & $1.70433 \times 10^{-64}$ & \textbf{Favors left} \\
     & MonotonicityCorrelation ($\mathrm{MC}$) & $\uparrow$ & $1479$ & $0.000287811$ & $[-0.0117839,\, 0.0123595]$ & $0.00121607$ & $0.962705$ & $1$ & Not detected \\
     & FaithfulnessEstimate ($\mathrm{FE}$) & $\uparrow$ & $1480$ & $0.0333873$ & $[0.0188519,\, 0.0479228]$ & $0.117118$ & $7.13805 \times 10^{-6}$ & $0.000349764$ & \textbf{Favors left} \\
     & PointingGame ($\mathrm{PG}$) & $\uparrow$ & $1475$ & $0.0244068$ & $[-0.00440716,\, 0.0532207]$ & $0.043263$ & $0.0968163$ & $1$ & Not detected \\
     & TopKIntersection ($\mathrm{TopK}$) & $\uparrow$ & $1475$ & $0.0248047$ & $[0.0158789,\, 0.0337305]$ & $0.141937$ & $5.85714 \times 10^{-8}$ & $3.28 \times 10^{-6}$ & \textbf{Favors left} \\
     & RelevanceRankAccuracy ($\mathrm{RRA}$) & $\uparrow$ & $1475$ & $0.00110567$ & $[-0.00143186,\, 0.00364319]$ & $0.0222547$ & $0.392851$ & $1$ & Not detected \\
     & RandomLogit & $\downarrow$ & $1480$ & $-0.0505644$ & $[-0.05873,\, -0.0423989]$ & $-0.315743$ & $1.97241 \times 10^{-32}$ & $1.51875 \times 10^{-30}$ & \textbf{Favors left} \\
     & DeletionAUC & $\downarrow$ & $1480$ & $-0.0711237$ & $[-0.0784674,\, -0.0637799]$ & $-0.493822$ & $3.49379 \times 10^{-72}$ & $3.14441 \times 10^{-70}$ & \textbf{Favors left} \\
     & InsertionAUC & $\uparrow$ & $1480$ & $0.0373545$ & $[0.0309828,\, 0.0437262]$ & $0.298925$ & $2.24698 \times 10^{-29}$ & $1.7077 \times 10^{-27}$ & \textbf{Favors left} \\
    \midrule
    \multirow{9}{*}{Oxford Pets $\times$ ResNet-50} & FaithfulnessCorrelation ($\mathrm{FC}$) & $\uparrow$ & $1480$ & $0.034192$ & $[0.0270739,\, 0.0413101]$ & $0.244924$ & $1.63442 \times 10^{-20}$ & $1.1441 \times 10^{-18}$ & \textbf{Favors left} \\
     & MonotonicityCorrelation ($\mathrm{MC}$) & $\uparrow$ & $1480$ & $0.0199855$ & $[0.00908614,\, 0.0308849]$ & $0.0934947$ & $0.000332743$ & $0.0133097$ & \textbf{Favors left} \\
     & FaithfulnessEstimate ($\mathrm{FE}$) & $\uparrow$ & $1480$ & $0.113086$ & $[0.0985838,\, 0.127589]$ & $0.397591$ & $3.83166 \times 10^{-49}$ & $3.21859 \times 10^{-47}$ & \textbf{Favors left} \\
     & PointingGame ($\mathrm{PG}$) & $\uparrow$ & $1477$ & $0.0501016$ & $[0.024262,\, 0.0759411]$ & $0.0989648$ & $0.000148552$ & $0.00623919$ & \textbf{Favors left} \\
     & TopKIntersection ($\mathrm{TopK}$) & $\uparrow$ & $1477$ & $0.0431571$ & $[0.0348649,\, 0.0514492]$ & $0.265643$ & $1.08283 \times 10^{-23}$ & $7.90463 \times 10^{-22}$ & \textbf{Favors left} \\
     & RelevanceRankAccuracy ($\mathrm{RRA}$) & $\uparrow$ & $1477$ & $0.0135314$ & $[0.0114081,\, 0.0156547]$ & $0.325274$ & $3.70635 \times 10^{-34}$ & $2.92802 \times 10^{-32}$ & \textbf{Favors left} \\
     & RandomLogit & $\downarrow$ & $1480$ & $-0.0685257$ & $[-0.0733771,\, -0.0636744]$ & $-0.720224$ & $1.87131 \times 10^{-136}$ & $1.74032 \times 10^{-134}$ & \textbf{Favors left} \\
     & DeletionAUC & $\downarrow$ & $1480$ & $-0.0284796$ & $[-0.0315191,\, -0.0254401]$ & $-0.477759$ & $4.09399 \times 10^{-68}$ & $3.60271 \times 10^{-66}$ & \textbf{Favors left} \\
     & InsertionAUC & $\uparrow$ & $1480$ & $0.0252467$ & $[0.0217997,\, 0.0286937]$ & $0.373455$ & $6.49794 \times 10^{-44}$ & $5.26333 \times 10^{-42}$ & \textbf{Favors left} \\
    \bottomrule
  \end{tabular}%
  }
\end{table*}

\begin{table*}[t]
  \centering
  \caption{Complete paired Student $t$-test results for contrast $\mathbf{P-F}$ (Permuted Baseline $-$ Flow-IG; Composite path-deviation + coupling ablation ($P$ vs.\ $F$)) on generator \textbf{Flow-IG (PixelDiT)}. Differences and Cohen's $d_z$ are raw $\text{left} - \text{right}$ (positive favors left for $\uparrow$ metrics; negative favors left for $\downarrow$ metrics). Multiplicity adjustment uses Holm's step-down procedure across the $m=96$ applicable tests of family $\mathbf{P-F}$. Tally across the $48$ applicable tests for this generator: $\mathrm{L}=19$ Favors left, $\mathrm{ND}=20$ Not detected, $\mathrm{R}=9$ Favors right ($\mathrm{NA}=6$ Flowers-102 localization exclusions).}
  \label{tab:paired-pf-pixeldit}
  \scriptsize
  \setlength{\tabcolsep}{4pt}
  \renewcommand{\arraystretch}{0.96}
  \resizebox{\textwidth}{!}{%
  \begin{tabular}{ll c r r c r r r l}
    \toprule
    \textbf{Dataset $\times$ Backbone} & \textbf{Metric} & \textbf{Pol.} & $n$ & \textbf{Mean diff.\ ($\overline{\Delta}$)} & \textbf{95\% CI (unadjusted)} & \textbf{Cohen $d_z$} & $p_{\mathrm{raw}}$ & $p_{\mathrm{Holm}}$ & \textbf{Verdict} \\
    \midrule
    \multirow{9}{*}{Flowers-102 $\times$ PVT-v2-B2} & FaithfulnessCorrelation ($\mathrm{FC}$) & $\uparrow$ & $1020$ & $-0.0189489$ & $[-0.0272888,\, -0.0106089]$ & $-0.1396$ & $9.17095 \times 10^{-6}$ & $0.000440206$ & \underline{Favors right} \\
     & MonotonicityCorrelation ($\mathrm{MC}$) & $\uparrow$ & $1020$ & $-0.0461023$ & $[-0.0594627,\, -0.0327419]$ & $-0.212016$ & $2.15322 \times 10^{-11}$ & $1.37806 \times 10^{-9}$ & \underline{Favors right} \\
     & FaithfulnessEstimate ($\mathrm{FE}$) & $\uparrow$ & $1020$ & $-0.0528795$ & $[-0.0669887,\, -0.0387702]$ & $-0.230275$ & $3.93612 \times 10^{-13}$ & $2.55848 \times 10^{-11}$ & \underline{Favors right} \\
     & PointingGame ($\mathrm{PG}$) & $\uparrow$ & \textnormal{---} & \textnormal{---} & $[\textnormal{---},\, \textnormal{---}]$ & \textnormal{---} & \textnormal{---} & \textnormal{---} & \textnormal{Not applicable} \\
     & TopKIntersection ($\mathrm{TopK}$) & $\uparrow$ & \textnormal{---} & \textnormal{---} & $[\textnormal{---},\, \textnormal{---}]$ & \textnormal{---} & \textnormal{---} & \textnormal{---} & \textnormal{Not applicable} \\
     & RelevanceRankAccuracy ($\mathrm{RRA}$) & $\uparrow$ & \textnormal{---} & \textnormal{---} & $[\textnormal{---},\, \textnormal{---}]$ & \textnormal{---} & \textnormal{---} & \textnormal{---} & \textnormal{Not applicable} \\
     & RandomLogit & $\downarrow$ & $1020$ & $-0.054313$ & $[-0.0596203,\, -0.0490057]$ & $-0.628774$ & $7.79372 \times 10^{-76}$ & $7.09228 \times 10^{-74}$ & \textbf{Favors left} \\
     & DeletionAUC & $\downarrow$ & $1020$ & $-0.0035028$ & $[-0.00657495,\, -0.000430642]$ & $-0.0700545$ & $0.0254783$ & $0.611478$ & Not detected \\
     & InsertionAUC & $\uparrow$ & $1020$ & $0.00532856$ & $[0.00175853,\, 0.00889858]$ & $0.0917069$ & $0.00347756$ & $0.102427$ & Not detected \\
    \midrule
    \multirow{9}{*}{Flowers-102 $\times$ ResNet-50} & FaithfulnessCorrelation ($\mathrm{FC}$) & $\uparrow$ & $1020$ & $-0.00814255$ & $[-0.0167379,\, 0.000452822]$ & $-0.0582049$ & $0.063327$ & $1$ & Not detected \\
     & MonotonicityCorrelation ($\mathrm{MC}$) & $\uparrow$ & $1020$ & $-0.0245508$ & $[-0.0346048,\, -0.0144968]$ & $-0.150034$ & $1.89919 \times 10^{-6}$ & $0.000100657$ & \underline{Favors right} \\
     & FaithfulnessEstimate ($\mathrm{FE}$) & $\uparrow$ & $1020$ & $0.0219078$ & $[0.0104294,\, 0.0333861]$ & $0.117269$ & $0.000190294$ & $0.00780205$ & \textbf{Favors left} \\
     & PointingGame ($\mathrm{PG}$) & $\uparrow$ & \textnormal{---} & \textnormal{---} & $[\textnormal{---},\, \textnormal{---}]$ & \textnormal{---} & \textnormal{---} & \textnormal{---} & \textnormal{Not applicable} \\
     & TopKIntersection ($\mathrm{TopK}$) & $\uparrow$ & \textnormal{---} & \textnormal{---} & $[\textnormal{---},\, \textnormal{---}]$ & \textnormal{---} & \textnormal{---} & \textnormal{---} & \textnormal{Not applicable} \\
     & RelevanceRankAccuracy ($\mathrm{RRA}$) & $\uparrow$ & \textnormal{---} & \textnormal{---} & $[\textnormal{---},\, \textnormal{---}]$ & \textnormal{---} & \textnormal{---} & \textnormal{---} & \textnormal{Not applicable} \\
     & RandomLogit & $\downarrow$ & $1020$ & $-0.0727353$ & $[-0.0769278,\, -0.0685428]$ & $-1.06595$ & $2.89107 \times 10^{-170}$ & $2.74651 \times 10^{-168}$ & \textbf{Favors left} \\
     & DeletionAUC & $\downarrow$ & $1020$ & $-0.0028386$ & $[-0.00406431,\, -0.00161289]$ & $-0.142292$ & $6.16764 \times 10^{-6}$ & $0.000308382$ & \textbf{Favors left} \\
     & InsertionAUC & $\uparrow$ & $1020$ & $-8.94529e-05$ & $[-0.00147981,\, 0.0013009]$ & $-0.00395306$ & $0.899559$ & $1$ & Not detected \\
    \midrule
    \multirow{9}{*}{ImageNet $\times$ PVT-v2-B2} & FaithfulnessCorrelation ($\mathrm{FC}$) & $\uparrow$ & $1000$ & $-0.00664412$ & $[-0.0153673,\, 0.0020791]$ & $-0.0472645$ & $0.135325$ & $1$ & Not detected \\
     & MonotonicityCorrelation ($\mathrm{MC}$) & $\uparrow$ & $1000$ & $-0.018919$ & $[-0.034692,\, -0.00314606]$ & $-0.074432$ & $0.0187782$ & $0.469454$ & Not detected \\
     & FaithfulnessEstimate ($\mathrm{FE}$) & $\uparrow$ & $1000$ & $0.00300478$ & $[-0.0138666,\, 0.0198762]$ & $0.0110519$ & $0.726794$ & $1$ & Not detected \\
     & PointingGame ($\mathrm{PG}$) & $\uparrow$ & $1000$ & $-0.021$ & $[-0.0520776,\, 0.0100776]$ & $-0.0419321$ & $0.185139$ & $1$ & Not detected \\
     & TopKIntersection ($\mathrm{TopK}$) & $\uparrow$ & $1000$ & $-0.006342$ & $[-0.0134833,\, 0.000799316]$ & $-0.055109$ & $0.0816938$ & $1$ & Not detected \\
     & RelevanceRankAccuracy ($\mathrm{RRA}$) & $\uparrow$ & $1000$ & $-0.00329272$ & $[-0.00519867,\, -0.00138678]$ & $-0.107206$ & $0.000725868$ & $0.0232278$ & \underline{Favors right} \\
     & RandomLogit & $\downarrow$ & $1000$ & $-0.0301872$ & $[-0.0377384,\, -0.022636]$ & $-0.248075$ & $1.11028 \times 10^{-14}$ & $7.43891 \times 10^{-13}$ & \textbf{Favors left} \\
     & DeletionAUC & $\downarrow$ & $1000$ & $-0.0141429$ & $[-0.0183937,\, -0.00989206]$ & $-0.206462$ & $1.05252 \times 10^{-10}$ & $6.6309 \times 10^{-9}$ & \textbf{Favors left} \\
     & InsertionAUC & $\uparrow$ & $1000$ & $0.00356173$ & $[-0.00180776,\, 0.00893122]$ & $0.0411626$ & $0.193327$ & $1$ & Not detected \\
    \midrule
    \multirow{9}{*}{ImageNet $\times$ ResNet-50} & FaithfulnessCorrelation ($\mathrm{FC}$) & $\uparrow$ & $1000$ & $0.00129499$ & $[-0.00750153,\, 0.0100915]$ & $0.00913548$ & $0.772726$ & $1$ & Not detected \\
     & MonotonicityCorrelation ($\mathrm{MC}$) & $\uparrow$ & $1000$ & $-0.0230405$ & $[-0.03597,\, -0.0101109]$ & $-0.110582$ & $0.000491414$ & $0.0181823$ & \underline{Favors right} \\
     & FaithfulnessEstimate ($\mathrm{FE}$) & $\uparrow$ & $1000$ & $0.000304769$ & $[-0.0153905,\, 0.016]$ & $0.00120497$ & $0.969612$ & $1$ & Not detected \\
     & PointingGame ($\mathrm{PG}$) & $\uparrow$ & $1000$ & $0.017$ & $[-0.00328133,\, 0.0372813]$ & $0.0520148$ & $0.100315$ & $1$ & Not detected \\
     & TopKIntersection ($\mathrm{TopK}$) & $\uparrow$ & $1000$ & $0.004931$ & $[0.00101396,\, 0.00884804]$ & $0.0781181$ & $0.0136656$ & $0.355306$ & Not detected \\
     & RelevanceRankAccuracy ($\mathrm{RRA}$) & $\uparrow$ & $1000$ & $-0.00246584$ & $[-0.00387951,\, -0.00105217]$ & $-0.108241$ & $0.000644742$ & $0.0214083$ & \underline{Favors right} \\
     & RandomLogit & $\downarrow$ & $1000$ & $-0.0702752$ & $[-0.0761543,\, -0.0643961]$ & $-0.74177$ & $2.80039 \times 10^{-97}$ & $2.57636 \times 10^{-95}$ & \textbf{Favors left} \\
     & DeletionAUC & $\downarrow$ & $1000$ & $-0.00435201$ & $[-0.00619799,\, -0.00250603]$ & $-0.146298$ & $4.21071 \times 10^{-6}$ & $0.000214746$ & \textbf{Favors left} \\
     & InsertionAUC & $\uparrow$ & $1000$ & $0.011952$ & $[0.00891722,\, 0.0149868]$ & $0.244391$ & $2.64717 \times 10^{-14}$ & $1.74713 \times 10^{-12}$ & \textbf{Favors left} \\
    \midrule
    \multirow{9}{*}{Oxford Pets $\times$ PVT-v2-B2} & FaithfulnessCorrelation ($\mathrm{FC}$) & $\uparrow$ & $1480$ & $-0.00623944$ & $[-0.0132552,\, 0.000776326]$ & $-0.0453464$ & $0.0812774$ & $1$ & Not detected \\
     & MonotonicityCorrelation ($\mathrm{MC}$) & $\uparrow$ & $1479$ & $-0.023202$ & $[-0.0337198,\, -0.0126842]$ & $-0.112518$ & $1.61171 \times 10^{-5}$ & $0.000757506$ & \underline{Favors right} \\
     & FaithfulnessEstimate ($\mathrm{FE}$) & $\uparrow$ & $1480$ & $0.000710799$ & $[-0.0096813,\, 0.0111029]$ & $0.00348752$ & $0.893288$ & $1$ & Not detected \\
     & PointingGame ($\mathrm{PG}$) & $\uparrow$ & $1475$ & $-0.0020339$ & $[-0.0290692,\, 0.0250014]$ & $-0.00384243$ & $0.882701$ & $1$ & Not detected \\
     & TopKIntersection ($\mathrm{TopK}$) & $\uparrow$ & $1475$ & $0.00893763$ & $[0.00398796,\, 0.0138873]$ & $0.0922263$ & $0.000409448$ & $0.015559$ & \textbf{Favors left} \\
     & RelevanceRankAccuracy ($\mathrm{RRA}$) & $\uparrow$ & $1475$ & $-0.00135817$ & $[-0.00275624,\, 3.99091e-05]$ & $-0.0496171$ & $0.0568991$ & $1$ & Not detected \\
     & RandomLogit & $\downarrow$ & $1480$ & $-0.0399763$ & $[-0.0442028,\, -0.0357497]$ & $-0.482268$ & $3.00806 \times 10^{-69}$ & $2.67717 \times 10^{-67}$ & \textbf{Favors left} \\
     & DeletionAUC & $\downarrow$ & $1480$ & $-0.0347772$ & $[-0.040084,\, -0.0294703]$ & $-0.334142$ & $6.27841 \times 10^{-36}$ & $5.02273 \times 10^{-34}$ & \textbf{Favors left} \\
     & InsertionAUC & $\uparrow$ & $1480$ & $0.0114032$ & $[0.0066722,\, 0.0161342]$ & $0.122899$ & $2.48332 \times 10^{-6}$ & $0.000129132$ & \textbf{Favors left} \\
    \midrule
    \multirow{9}{*}{Oxford Pets $\times$ ResNet-50} & FaithfulnessCorrelation ($\mathrm{FC}$) & $\uparrow$ & $1480$ & $1.43967e-05$ & $[-0.00691342,\, 0.00694222]$ & $0.000105959$ & $0.996748$ & $1$ & Not detected \\
     & MonotonicityCorrelation ($\mathrm{MC}$) & $\uparrow$ & $1480$ & $-0.0204156$ & $[-0.0303551,\, -0.0104762]$ & $-0.104731$ & $5.88418 \times 10^{-5}$ & $0.00258904$ & \underline{Favors right} \\
     & FaithfulnessEstimate ($\mathrm{FE}$) & $\uparrow$ & $1480$ & $0.0203412$ & $[0.00888122,\, 0.0318011]$ & $0.0905037$ & $0.000512654$ & $0.0184556$ & \textbf{Favors left} \\
     & PointingGame ($\mathrm{PG}$) & $\uparrow$ & $1477$ & $0.0216655$ & $[-0.00309314,\, 0.0464242]$ & $0.0446638$ & $0.0862786$ & $1$ & Not detected \\
     & TopKIntersection ($\mathrm{TopK}$) & $\uparrow$ & $1477$ & $0.0113751$ & $[0.00773976,\, 0.0150104]$ & $0.159707$ & $1.07287 \times 10^{-9}$ & $6.32992 \times 10^{-8}$ & \textbf{Favors left} \\
     & RelevanceRankAccuracy ($\mathrm{RRA}$) & $\uparrow$ & $1477$ & $0.00212949$ & $[0.00105983,\, 0.00319915]$ & $0.101612$ & $9.84333 \times 10^{-5}$ & $0.00423263$ & \textbf{Favors left} \\
     & RandomLogit & $\downarrow$ & $1480$ & $-0.0675958$ & $[-0.0712627,\, -0.0639288]$ & $-0.939912$ & $1.13018 \times 10^{-205}$ & $1.08497 \times 10^{-203}$ & \textbf{Favors left} \\
     & DeletionAUC & $\downarrow$ & $1480$ & $-0.00665524$ & $[-0.00867924,\, -0.00463125]$ & $-0.167659$ & $1.5114 \times 10^{-10}$ & $9.37066 \times 10^{-9}$ & \textbf{Favors left} \\
     & InsertionAUC & $\uparrow$ & $1480$ & $0.0067208$ & $[0.00349892,\, 0.00994269]$ & $0.106361$ & $4.51123 \times 10^{-5}$ & $0.00207517$ & \textbf{Favors left} \\
    \bottomrule
  \end{tabular}%
  }
\end{table*}

\clearpage

\begin{table*}[t]
  \centering
  \caption{Complete paired Student $t$-test results for contrast $\mathbf{G-F}$ (Gaussian Linear IG $-$ Flow-IG; End-to-end generator replacement ($G$ vs.\ $F$, Hypothesis~\circnumC{C})) on generator \textbf{Flow-IG (Diffusion)}. Differences and Cohen's $d_z$ are raw $\text{left} - \text{right}$ (positive favors left for $\uparrow$ metrics; negative favors left for $\downarrow$ metrics). Multiplicity adjustment uses Holm's step-down procedure across the $m=96$ applicable tests of family $\mathbf{G-F}$. Tally across the $48$ applicable tests for this generator: $\mathrm{L}=40$ Favors left, $\mathrm{ND}=8$ Not detected, $\mathrm{R}=0$ Favors right ($\mathrm{NA}=6$ Flowers-102 localization exclusions).}
  \label{tab:paired-gf-diffusion}
  \scriptsize
  \setlength{\tabcolsep}{4pt}
  \renewcommand{\arraystretch}{0.96}
  \resizebox{\textwidth}{!}{%
  \begin{tabular}{ll c r r c r r r l}
    \toprule
    \textbf{Dataset $\times$ Backbone} & \textbf{Metric} & \textbf{Pol.} & $n$ & \textbf{Mean diff.\ ($\overline{\Delta}$)} & \textbf{95\% CI (unadjusted)} & \textbf{Cohen $d_z$} & $p_{\mathrm{raw}}$ & $p_{\mathrm{Holm}}$ & \textbf{Verdict} \\
    \midrule
    \multirow{9}{*}{Flowers-102 $\times$ PVT-v2-B2} & FaithfulnessCorrelation ($\mathrm{FC}$) & $\uparrow$ & $1020$ & $0.0456062$ & $[0.0365769,\, 0.0546355]$ & $0.310337$ & $3.6014 \times 10^{-22}$ & $2.30489 \times 10^{-20}$ & \textbf{Favors left} \\
     & MonotonicityCorrelation ($\mathrm{MC}$) & $\uparrow$ & $1020$ & $-0.00297$ & $[-0.0191984,\, 0.0132584]$ & $-0.0112446$ & $0.719576$ & $1$ & Not detected \\
     & FaithfulnessEstimate ($\mathrm{FE}$) & $\uparrow$ & $1020$ & $0.0322189$ & $[0.0143635,\, 0.0500744]$ & $0.110867$ & $0.000416916$ & $0.0125075$ & \textbf{Favors left} \\
     & PointingGame ($\mathrm{PG}$) & $\uparrow$ & \textnormal{---} & \textnormal{---} & $[\textnormal{---},\, \textnormal{---}]$ & \textnormal{---} & \textnormal{---} & \textnormal{---} & \textnormal{Not applicable} \\
     & TopKIntersection ($\mathrm{TopK}$) & $\uparrow$ & \textnormal{---} & \textnormal{---} & $[\textnormal{---},\, \textnormal{---}]$ & \textnormal{---} & \textnormal{---} & \textnormal{---} & \textnormal{Not applicable} \\
     & RelevanceRankAccuracy ($\mathrm{RRA}$) & $\uparrow$ & \textnormal{---} & \textnormal{---} & $[\textnormal{---},\, \textnormal{---}]$ & \textnormal{---} & \textnormal{---} & \textnormal{---} & \textnormal{Not applicable} \\
     & RandomLogit & $\downarrow$ & $1020$ & $-0.0817123$ & $[-0.089779,\, -0.0736455]$ & $-0.622375$ & $1.47574 \times 10^{-74}$ & $1.28389 \times 10^{-72}$ & \textbf{Favors left} \\
     & DeletionAUC & $\downarrow$ & $1020$ & $-0.0346429$ & $[-0.0387214,\, -0.0305644]$ & $-0.521888$ & $2.42195 \times 10^{-55}$ & $1.986 \times 10^{-53}$ & \textbf{Favors left} \\
     & InsertionAUC & $\uparrow$ & $1020$ & $0.0325423$ & $[0.0275415,\, 0.037543]$ & $0.399832$ & $9.54185 \times 10^{-35}$ & $6.87013 \times 10^{-33}$ & \textbf{Favors left} \\
    \midrule
    \multirow{9}{*}{Flowers-102 $\times$ ResNet-50} & FaithfulnessCorrelation ($\mathrm{FC}$) & $\uparrow$ & $1020$ & $0.0252786$ & $[0.0165412,\, 0.0340161]$ & $0.177759$ & $1.78491 \times 10^{-8}$ & $7.67512 \times 10^{-7}$ & \textbf{Favors left} \\
     & MonotonicityCorrelation ($\mathrm{MC}$) & $\uparrow$ & $1020$ & $0.0409349$ & $[0.0266946,\, 0.0551752]$ & $0.17662$ & $2.19218 \times 10^{-8}$ & $9.20718 \times 10^{-7}$ & \textbf{Favors left} \\
     & FaithfulnessEstimate ($\mathrm{FE}$) & $\uparrow$ & $1020$ & $0.114566$ & $[0.0987412,\, 0.130391]$ & $0.444813$ & $6.35235 \times 10^{-42}$ & $4.89131 \times 10^{-40}$ & \textbf{Favors left} \\
     & PointingGame ($\mathrm{PG}$) & $\uparrow$ & \textnormal{---} & \textnormal{---} & $[\textnormal{---},\, \textnormal{---}]$ & \textnormal{---} & \textnormal{---} & \textnormal{---} & \textnormal{Not applicable} \\
     & TopKIntersection ($\mathrm{TopK}$) & $\uparrow$ & \textnormal{---} & \textnormal{---} & $[\textnormal{---},\, \textnormal{---}]$ & \textnormal{---} & \textnormal{---} & \textnormal{---} & \textnormal{Not applicable} \\
     & RelevanceRankAccuracy ($\mathrm{RRA}$) & $\uparrow$ & \textnormal{---} & \textnormal{---} & $[\textnormal{---},\, \textnormal{---}]$ & \textnormal{---} & \textnormal{---} & \textnormal{---} & \textnormal{Not applicable} \\
     & RandomLogit & $\downarrow$ & $1020$ & $-0.10587$ & $[-0.111403,\, -0.100336]$ & $-1.17557$ & $2.19617 \times 10^{-194}$ & $2.0644 \times 10^{-192}$ & \textbf{Favors left} \\
     & DeletionAUC & $\downarrow$ & $1020$ & $-0.0187981$ & $[-0.0206162,\, -0.01698]$ & $-0.635264$ & $3.89521 \times 10^{-77}$ & $3.42779 \times 10^{-75}$ & \textbf{Favors left} \\
     & InsertionAUC & $\uparrow$ & $1020$ & $0.0066804$ & $[0.0047253,\, 0.00863551]$ & $0.209941$ & $3.33155 \times 10^{-11}$ & $1.63246 \times 10^{-9}$ & \textbf{Favors left} \\
    \midrule
    \multirow{9}{*}{ImageNet $\times$ PVT-v2-B2} & FaithfulnessCorrelation ($\mathrm{FC}$) & $\uparrow$ & $1000$ & $0.0367156$ & $[0.0279097,\, 0.0455216]$ & $0.258731$ & $8.45166 \times 10^{-16}$ & $4.64842 \times 10^{-14}$ & \textbf{Favors left} \\
     & MonotonicityCorrelation ($\mathrm{MC}$) & $\uparrow$ & $1000$ & $0.0161933$ & $[-0.000435414,\, 0.032822]$ & $0.0604298$ & $0.0562959$ & $0.95703$ & Not detected \\
     & FaithfulnessEstimate ($\mathrm{FE}$) & $\uparrow$ & $1000$ & $0.0856032$ & $[0.0664882,\, 0.104718]$ & $0.277902$ & $6.53408 \times 10^{-18}$ & $3.92045 \times 10^{-16}$ & \textbf{Favors left} \\
     & PointingGame ($\mathrm{PG}$) & $\uparrow$ & $1000$ & $-0.015$ & $[-0.0459667,\, 0.0159667]$ & $-0.0300587$ & $0.342068$ & $1$ & Not detected \\
     & TopKIntersection ($\mathrm{TopK}$) & $\uparrow$ & $1000$ & $-0.000268$ & $[-0.0102836,\, 0.00974758]$ & $-0.00166048$ & $0.958134$ & $1$ & Not detected \\
     & RelevanceRankAccuracy ($\mathrm{RRA}$) & $\uparrow$ & $1000$ & $0.0027825$ & $[0.000164583,\, 0.00540041]$ & $0.0659559$ & $0.0372582$ & $0.782422$ & Not detected \\
     & RandomLogit & $\downarrow$ & $1000$ & $-0.0502409$ & $[-0.0609675,\, -0.0395143]$ & $-0.29065$ & $2.19549 \times 10^{-19}$ & $1.38316 \times 10^{-17}$ & \textbf{Favors left} \\
     & DeletionAUC & $\downarrow$ & $1000$ & $-0.0219526$ & $[-0.026688,\, -0.0172171]$ & $-0.287673$ & $4.90419 \times 10^{-19}$ & $2.99156 \times 10^{-17}$ & \textbf{Favors left} \\
     & InsertionAUC & $\uparrow$ & $1000$ & $0.0306343$ & $[0.02409,\, 0.0371786]$ & $0.290483$ & $2.29727 \times 10^{-19}$ & $1.42431 \times 10^{-17}$ & \textbf{Favors left} \\
    \midrule
    \multirow{9}{*}{ImageNet $\times$ ResNet-50} & FaithfulnessCorrelation ($\mathrm{FC}$) & $\uparrow$ & $1000$ & $0.0385227$ & $[0.0296477,\, 0.0473978]$ & $0.269352$ & $5.92318 \times 10^{-17}$ & $3.37621 \times 10^{-15}$ & \textbf{Favors left} \\
     & MonotonicityCorrelation ($\mathrm{MC}$) & $\uparrow$ & $1000$ & $0.01436$ & $[0.000537958,\, 0.0281819]$ & $0.0644699$ & $0.0417412$ & $0.834824$ & Not detected \\
     & FaithfulnessEstimate ($\mathrm{FE}$) & $\uparrow$ & $1000$ & $0.0313274$ & $[0.0140836,\, 0.0485712]$ & $0.112737$ & $0.000381013$ & $0.0118114$ & \textbf{Favors left} \\
     & PointingGame ($\mathrm{PG}$) & $\uparrow$ & $1000$ & $0.087$ & $[0.0625229,\, 0.111477]$ & $0.220563$ & $5.5651 \times 10^{-12}$ & $2.78255 \times 10^{-10}$ & \textbf{Favors left} \\
     & TopKIntersection ($\mathrm{TopK}$) & $\uparrow$ & $1000$ & $0.051602$ & $[0.0418958,\, 0.0613082]$ & $0.329906$ & $2.94521 \times 10^{-24}$ & $1.97329 \times 10^{-22}$ & \textbf{Favors left} \\
     & RelevanceRankAccuracy ($\mathrm{RRA}$) & $\uparrow$ & $1000$ & $0.00710865$ & $[0.00462828,\, 0.00958901]$ & $0.177847$ & $2.42112 \times 10^{-8}$ & $9.92658 \times 10^{-7}$ & \textbf{Favors left} \\
     & RandomLogit & $\downarrow$ & $1000$ & $-0.106849$ & $[-0.115277,\, -0.0984202]$ & $-0.786664$ & $1.07071 \times 10^{-106}$ & $9.63635 \times 10^{-105}$ & \textbf{Favors left} \\
     & DeletionAUC & $\downarrow$ & $1000$ & $-0.0108507$ & $[-0.0129548,\, -0.00874658]$ & $-0.320009$ & $5.54542 \times 10^{-23}$ & $3.60452 \times 10^{-21}$ & \textbf{Favors left} \\
     & InsertionAUC & $\uparrow$ & $1000$ & $0.020273$ & $[0.0169623,\, 0.0235837]$ & $0.379992$ & $3.65196 \times 10^{-31}$ & $2.59289 \times 10^{-29}$ & \textbf{Favors left} \\
    \midrule
    \multirow{9}{*}{Oxford Pets $\times$ PVT-v2-B2} & FaithfulnessCorrelation ($\mathrm{FC}$) & $\uparrow$ & $1480$ & $0.0815707$ & $[0.0741949,\, 0.0889465]$ & $0.563893$ & $7.95557 \times 10^{-91}$ & $7.08045 \times 10^{-89}$ & \textbf{Favors left} \\
     & MonotonicityCorrelation ($\mathrm{MC}$) & $\uparrow$ & $1479$ & $0.014747$ & $[0.00295018,\, 0.0265437]$ & $0.0637616$ & $0.0143164$ & $0.335022$ & Not detected \\
     & FaithfulnessEstimate ($\mathrm{FE}$) & $\uparrow$ & $1480$ & $0.0718283$ & $[0.0578739,\, 0.0857828]$ & $0.262456$ & $3.16355 \times 10^{-23}$ & $2.08794 \times 10^{-21}$ & \textbf{Favors left} \\
     & PointingGame ($\mathrm{PG}$) & $\uparrow$ & $1475$ & $0.0677966$ & $[0.0394752,\, 0.096118]$ & $0.122265$ & $2.90446 \times 10^{-6}$ & $0.000104561$ & \textbf{Favors left} \\
     & TopKIntersection ($\mathrm{TopK}$) & $\uparrow$ & $1475$ & $0.054678$ & $[0.0463765,\, 0.0629794]$ & $0.336409$ & $2.97312 \times 10^{-36}$ & $2.17038 \times 10^{-34}$ & \textbf{Favors left} \\
     & RelevanceRankAccuracy ($\mathrm{RRA}$) & $\uparrow$ & $1475$ & $0.0142894$ & $[0.0118808,\, 0.016698]$ & $0.303008$ & $5.22562 \times 10^{-30}$ & $3.65794 \times 10^{-28}$ & \textbf{Favors left} \\
     & RandomLogit & $\downarrow$ & $1480$ & $-0.0897617$ & $[-0.096542,\, -0.0829814]$ & $-0.675018$ & $8.31526 \times 10^{-123}$ & $7.56689 \times 10^{-121}$ & \textbf{Favors left} \\
     & DeletionAUC & $\downarrow$ & $1480$ & $-0.0576338$ & $[-0.0649504,\, -0.0503173]$ & $-0.401647$ & $4.80166 \times 10^{-50}$ & $3.79331 \times 10^{-48}$ & \textbf{Favors left} \\
     & InsertionAUC & $\uparrow$ & $1480$ & $0.0424581$ & $[0.0360216,\, 0.0488945]$ & $0.336347$ & $2.33412 \times 10^{-36}$ & $1.72725 \times 10^{-34}$ & \textbf{Favors left} \\
    \midrule
    \multirow{9}{*}{Oxford Pets $\times$ ResNet-50} & FaithfulnessCorrelation ($\mathrm{FC}$) & $\uparrow$ & $1480$ & $0.039122$ & $[0.0317751,\, 0.0464688]$ & $0.271515$ & $1.08539 \times 10^{-24}$ & $7.38065 \times 10^{-23}$ & \textbf{Favors left} \\
     & MonotonicityCorrelation ($\mathrm{MC}$) & $\uparrow$ & $1480$ & $-0.000213028$ & $[-0.0112542,\, 0.0108281]$ & $-0.000983775$ & $0.969815$ & $1$ & Not detected \\
     & FaithfulnessEstimate ($\mathrm{FE}$) & $\uparrow$ & $1480$ & $0.0969557$ & $[0.0833403,\, 0.110571]$ & $0.36309$ & $9.63177 \times 10^{-42}$ & $7.32014 \times 10^{-40}$ & \textbf{Favors left} \\
     & PointingGame ($\mathrm{PG}$) & $\uparrow$ & $1477$ & $0.070413$ & $[0.0445603,\, 0.0962657]$ & $0.139015$ & $1.05999 \times 10^{-7}$ & $4.13394 \times 10^{-6}$ & \textbf{Favors left} \\
     & TopKIntersection ($\mathrm{TopK}$) & $\uparrow$ & $1477$ & $0.0590041$ & $[0.0513575,\, 0.0666507]$ & $0.393847$ & $3.19626 \times 10^{-48}$ & $2.49309 \times 10^{-46}$ & \textbf{Favors left} \\
     & RelevanceRankAccuracy ($\mathrm{RRA}$) & $\uparrow$ & $1477$ & $0.0172676$ & $[0.0153593,\, 0.0191759]$ & $0.461855$ & $4.8293 \times 10^{-64}$ & $4.05661 \times 10^{-62}$ & \textbf{Favors left} \\
     & RandomLogit & $\downarrow$ & $1480$ & $-0.084275$ & $[-0.0887078,\, -0.0798422]$ & $-0.969377$ & $3.8939 \times 10^{-215}$ & $3.69921 \times 10^{-213}$ & \textbf{Favors left} \\
     & DeletionAUC & $\downarrow$ & $1480$ & $-0.0279562$ & $[-0.0308667,\, -0.0250457]$ & $-0.48976$ & $3.80098 \times 10^{-71}$ & $3.23084 \times 10^{-69}$ & \textbf{Favors left} \\
     & InsertionAUC & $\uparrow$ & $1480$ & $0.0296481$ & $[0.0262115,\, 0.0330848]$ & $0.43988$ & $7.35345 \times 10^{-59}$ & $6.10337 \times 10^{-57}$ & \textbf{Favors left} \\
    \bottomrule
  \end{tabular}%
  }
\end{table*}

\begin{table*}[t]
  \centering
  \caption{Complete paired Student $t$-test results for contrast $\mathbf{G-F}$ (Gaussian Linear IG $-$ Flow-IG; End-to-end generator replacement ($G$ vs.\ $F$, Hypothesis~\circnumC{C})) on generator \textbf{Flow-IG (PixelDiT)}. Differences and Cohen's $d_z$ are raw $\text{left} - \text{right}$ (positive favors left for $\uparrow$ metrics; negative favors left for $\downarrow$ metrics). Multiplicity adjustment uses Holm's step-down procedure across the $m=96$ applicable tests of family $\mathbf{G-F}$. Tally across the $48$ applicable tests for this generator: $\mathrm{L}=23$ Favors left, $\mathrm{ND}=20$ Not detected, $\mathrm{R}=5$ Favors right ($\mathrm{NA}=6$ Flowers-102 localization exclusions).}
  \label{tab:paired-gf-pixeldit}
  \scriptsize
  \setlength{\tabcolsep}{4pt}
  \renewcommand{\arraystretch}{0.96}
  \resizebox{\textwidth}{!}{%
  \begin{tabular}{ll c r r c r r r l}
    \toprule
    \textbf{Dataset $\times$ Backbone} & \textbf{Metric} & \textbf{Pol.} & $n$ & \textbf{Mean diff.\ ($\overline{\Delta}$)} & \textbf{95\% CI (unadjusted)} & \textbf{Cohen $d_z$} & $p_{\mathrm{raw}}$ & $p_{\mathrm{Holm}}$ & \textbf{Verdict} \\
    \midrule
    \multirow{9}{*}{Flowers-102 $\times$ PVT-v2-B2} & FaithfulnessCorrelation ($\mathrm{FC}$) & $\uparrow$ & $1020$ & $-0.0154783$ & $[-0.0238435,\, -0.00711306]$ & $-0.113687$ & $0.000296534$ & $0.00948909$ & \underline{Favors right} \\
     & MonotonicityCorrelation ($\mathrm{MC}$) & $\uparrow$ & $1020$ & $-0.0372375$ & $[-0.0512122,\, -0.0232629]$ & $-0.163721$ & $2.06957 \times 10^{-7}$ & $7.86436 \times 10^{-6}$ & \underline{Favors right} \\
     & FaithfulnessEstimate ($\mathrm{FE}$) & $\uparrow$ & $1020$ & $-0.0199508$ & $[-0.0344993,\, -0.00540233]$ & $-0.0842572$ & $0.00724132$ & $0.188274$ & Not detected \\
     & PointingGame ($\mathrm{PG}$) & $\uparrow$ & \textnormal{---} & \textnormal{---} & $[\textnormal{---},\, \textnormal{---}]$ & \textnormal{---} & \textnormal{---} & \textnormal{---} & \textnormal{Not applicable} \\
     & TopKIntersection ($\mathrm{TopK}$) & $\uparrow$ & \textnormal{---} & \textnormal{---} & $[\textnormal{---},\, \textnormal{---}]$ & \textnormal{---} & \textnormal{---} & \textnormal{---} & \textnormal{Not applicable} \\
     & RelevanceRankAccuracy ($\mathrm{RRA}$) & $\uparrow$ & \textnormal{---} & \textnormal{---} & $[\textnormal{---},\, \textnormal{---}]$ & \textnormal{---} & \textnormal{---} & \textnormal{---} & \textnormal{Not applicable} \\
     & RandomLogit & $\downarrow$ & $1020$ & $-0.0444093$ & $[-0.049933,\, -0.0388855]$ & $-0.493977$ & $2.47517 \times 10^{-50}$ & $1.98014 \times 10^{-48}$ & \textbf{Favors left} \\
     & DeletionAUC & $\downarrow$ & $1020$ & $-0.00403533$ & $[-0.00725089,\, -0.000819778]$ & $-0.0771059$ & $0.0139593$ & $0.335022$ & Not detected \\
     & InsertionAUC & $\uparrow$ & $1020$ & $0.0123768$ & $[0.00841778,\, 0.0163359]$ & $0.19208$ & $1.22055 \times 10^{-9}$ & $5.61453 \times 10^{-8}$ & \textbf{Favors left} \\
    \midrule
    \multirow{9}{*}{Flowers-102 $\times$ ResNet-50} & FaithfulnessCorrelation ($\mathrm{FC}$) & $\uparrow$ & $1020$ & $-0.00681$ & $[-0.0152907,\, 0.00167073]$ & $-0.0493376$ & $0.115401$ & $1$ & Not detected \\
     & MonotonicityCorrelation ($\mathrm{MC}$) & $\uparrow$ & $1020$ & $-0.0697405$ & $[-0.0816225,\, -0.0578586]$ & $-0.360629$ & $6.09383 \times 10^{-29}$ & $4.20474 \times 10^{-27}$ & \underline{Favors right} \\
     & FaithfulnessEstimate ($\mathrm{FE}$) & $\uparrow$ & $1020$ & $0.000902059$ & $[-0.0112238,\, 0.0130279]$ & $0.00457074$ & $0.883968$ & $1$ & Not detected \\
     & PointingGame ($\mathrm{PG}$) & $\uparrow$ & \textnormal{---} & \textnormal{---} & $[\textnormal{---},\, \textnormal{---}]$ & \textnormal{---} & \textnormal{---} & \textnormal{---} & \textnormal{Not applicable} \\
     & TopKIntersection ($\mathrm{TopK}$) & $\uparrow$ & \textnormal{---} & \textnormal{---} & $[\textnormal{---},\, \textnormal{---}]$ & \textnormal{---} & \textnormal{---} & \textnormal{---} & \textnormal{Not applicable} \\
     & RelevanceRankAccuracy ($\mathrm{RRA}$) & $\uparrow$ & \textnormal{---} & \textnormal{---} & $[\textnormal{---},\, \textnormal{---}]$ & \textnormal{---} & \textnormal{---} & \textnormal{---} & \textnormal{Not applicable} \\
     & RandomLogit & $\downarrow$ & $1020$ & $-0.084052$ & $[-0.088551,\, -0.0795531]$ & $-1.1479$ & $2.52658 \times 10^{-188}$ & $2.34972 \times 10^{-186}$ & \textbf{Favors left} \\
     & DeletionAUC & $\downarrow$ & $1020$ & $-0.00373088$ & $[-0.00506483,\, -0.00239693]$ & $-0.171844$ & $5.12235 \times 10^{-8}$ & $2.04894 \times 10^{-6}$ & \textbf{Favors left} \\
     & InsertionAUC & $\uparrow$ & $1020$ & $0.00257313$ & $[0.00107258,\, 0.00407369]$ & $0.10536$ & $0.000794224$ & $0.0230325$ & \textbf{Favors left} \\
    \midrule
    \multirow{9}{*}{ImageNet $\times$ PVT-v2-B2} & FaithfulnessCorrelation ($\mathrm{FC}$) & $\uparrow$ & $1000$ & $0.0097974$ & $[0.00124059,\, 0.0183542]$ & $0.0710516$ & $0.0248674$ & $0.547083$ & Not detected \\
     & MonotonicityCorrelation ($\mathrm{MC}$) & $\uparrow$ & $1000$ & $-0.0129341$ & $[-0.028765,\, 0.0028969]$ & $-0.0506993$ & $0.109195$ & $1$ & Not detected \\
     & FaithfulnessEstimate ($\mathrm{FE}$) & $\uparrow$ & $1000$ & $0.0405588$ & $[0.0240071,\, 0.0571105]$ & $0.152061$ & $1.75381 \times 10^{-6}$ & $6.48909 \times 10^{-5}$ & \textbf{Favors left} \\
     & PointingGame ($\mathrm{PG}$) & $\uparrow$ & $1000$ & $-0.019$ & $[-0.0499583,\, 0.0119583]$ & $-0.0380848$ & $0.228741$ & $1$ & Not detected \\
     & TopKIntersection ($\mathrm{TopK}$) & $\uparrow$ & $1000$ & $-0.003463$ & $[-0.0108825,\, 0.00395655]$ & $-0.0289634$ & $0.359939$ & $1$ & Not detected \\
     & RelevanceRankAccuracy ($\mathrm{RRA}$) & $\uparrow$ & $1000$ & $0.00190745$ & $[1.72332e-05,\, 0.00379766]$ & $0.0626204$ & $0.0479515$ & $0.863126$ & Not detected \\
     & RandomLogit & $\downarrow$ & $1000$ & $-0.0185287$ & $[-0.027163,\, -0.00989437]$ & $-0.133166$ & $2.77106 \times 10^{-5}$ & $0.000914451$ & \textbf{Favors left} \\
     & DeletionAUC & $\downarrow$ & $1000$ & $-0.0186829$ & $[-0.0229834,\, -0.0143824]$ & $-0.269589$ & $5.57728 \times 10^{-17}$ & $3.23482 \times 10^{-15}$ & \textbf{Favors left} \\
     & InsertionAUC & $\uparrow$ & $1000$ & $0.015911$ & $[0.0107032,\, 0.0211188]$ & $0.189592$ & $2.83211 \times 10^{-9}$ & $1.27445 \times 10^{-7}$ & \textbf{Favors left} \\
    \midrule
    \multirow{9}{*}{ImageNet $\times$ ResNet-50} & FaithfulnessCorrelation ($\mathrm{FC}$) & $\uparrow$ & $1000$ & $-0.000700467$ & $[-0.00955053,\, 0.00814959]$ & $-0.00491152$ & $0.876604$ & $1$ & Not detected \\
     & MonotonicityCorrelation ($\mathrm{MC}$) & $\uparrow$ & $1000$ & $-0.0315438$ & $[-0.0447821,\, -0.0183055]$ & $-0.147862$ & $3.32979 \times 10^{-6}$ & $0.000116543$ & \underline{Favors right} \\
     & FaithfulnessEstimate ($\mathrm{FE}$) & $\uparrow$ & $1000$ & $-0.00729444$ & $[-0.022949,\, 0.00836014]$ & $-0.0289151$ & $0.36074$ & $1$ & Not detected \\
     & PointingGame ($\mathrm{PG}$) & $\uparrow$ & $1000$ & $0.02$ & $[-0.00165004,\, 0.04165]$ & $0.0573252$ & $0.0701653$ & $1$ & Not detected \\
     & TopKIntersection ($\mathrm{TopK}$) & $\uparrow$ & $1000$ & $0.014868$ & $[0.0109857,\, 0.0187503]$ & $0.237651$ & $1.26067 \times 10^{-13}$ & $6.55547 \times 10^{-12}$ & \textbf{Favors left} \\
     & RelevanceRankAccuracy ($\mathrm{RRA}$) & $\uparrow$ & $1000$ & $0.00209575$ & $[0.000594851,\, 0.00359664]$ & $0.0866488$ & $0.00625196$ & $0.175055$ & Not detected \\
     & RandomLogit & $\downarrow$ & $1000$ & $-0.0819627$ & $[-0.0877681,\, -0.0761573]$ & $-0.876109$ & $8.39289 \times 10^{-126}$ & $7.72146 \times 10^{-124}$ & \textbf{Favors left} \\
     & DeletionAUC & $\downarrow$ & $1000$ & $-0.00262681$ & $[-0.00461494,\, -0.000638677]$ & $-0.0819894$ & $0.00966079$ & $0.24152$ & Not detected \\
     & InsertionAUC & $\uparrow$ & $1000$ & $0.0133057$ & $[0.0102891,\, 0.0163223]$ & $0.273711$ & $1.93895 \times 10^{-17}$ & $1.14398 \times 10^{-15}$ & \textbf{Favors left} \\
    \midrule
    \multirow{9}{*}{Oxford Pets $\times$ PVT-v2-B2} & FaithfulnessCorrelation ($\mathrm{FC}$) & $\uparrow$ & $1480$ & $0.0063545$ & $[-0.000868036,\, 0.013577]$ & $0.0448605$ & $0.0845888$ & $1$ & Not detected \\
     & MonotonicityCorrelation ($\mathrm{MC}$) & $\uparrow$ & $1479$ & $-0.00631106$ & $[-0.0171699,\, 0.00454778]$ & $-0.0296441$ & $0.254451$ & $1$ & Not detected \\
     & FaithfulnessEstimate ($\mathrm{FE}$) & $\uparrow$ & $1480$ & $0.02351$ & $[0.0128488,\, 0.0341712]$ & $0.11244$ & $1.62274 \times 10^{-5}$ & $0.000551731$ & \textbf{Favors left} \\
     & PointingGame ($\mathrm{PG}$) & $\uparrow$ & $1475$ & $0.0372881$ & $[0.0104514,\, 0.0641249]$ & $0.0709658$ & $0.0064965$ & $0.175405$ & Not detected \\
     & TopKIntersection ($\mathrm{TopK}$) & $\uparrow$ & $1475$ & $0.0358997$ & $[0.0308159,\, 0.0409834]$ & $0.360677$ & $4.13114 \times 10^{-41}$ & $3.09836 \times 10^{-39}$ & \textbf{Favors left} \\
     & RelevanceRankAccuracy ($\mathrm{RRA}$) & $\uparrow$ & $1475$ & $0.0127785$ & $[0.0112211,\, 0.0143358]$ & $0.419081$ & $8.073 \times 10^{-54}$ & $6.53913 \times 10^{-52}$ & \textbf{Favors left} \\
     & RandomLogit & $\downarrow$ & $1480$ & $-0.043603$ & $[-0.0481004,\, -0.0391056]$ & $-0.494342$ & $2.57169 \times 10^{-72}$ & $2.21165 \times 10^{-70}$ & \textbf{Favors left} \\
     & DeletionAUC & $\downarrow$ & $1480$ & $-0.0213584$ & $[-0.0266286,\, -0.0160881]$ & $-0.206638$ & $3.68552 \times 10^{-15}$ & $1.99018 \times 10^{-13}$ & \textbf{Favors left} \\
     & InsertionAUC & $\uparrow$ & $1480$ & $0.0184332$ & $[0.0136529,\, 0.0232136]$ & $0.196614$ & $6.83572 \times 10^{-14}$ & $3.62293 \times 10^{-12}$ & \textbf{Favors left} \\
    \midrule
    \multirow{9}{*}{Oxford Pets $\times$ ResNet-50} & FaithfulnessCorrelation ($\mathrm{FC}$) & $\uparrow$ & $1480$ & $0.00393013$ & $[-0.0031255,\, 0.0109858]$ & $0.0284017$ & $0.274732$ & $1$ & Not detected \\
     & MonotonicityCorrelation ($\mathrm{MC}$) & $\uparrow$ & $1480$ & $-0.0335134$ & $[-0.0438559,\, -0.0231709]$ & $-0.165221$ & $2.74834 \times 10^{-10}$ & $1.29172 \times 10^{-8}$ & \underline{Favors right} \\
     & FaithfulnessEstimate ($\mathrm{FE}$) & $\uparrow$ & $1480$ & $-0.0118512$ & $[-0.0232712,\, -0.00043128]$ & $-0.0529142$ & $0.0419636$ & $0.834824$ & Not detected \\
     & PointingGame ($\mathrm{PG}$) & $\uparrow$ & $1477$ & $0.0176032$ & $[-0.00673249,\, 0.041939]$ & $0.0369201$ & $0.156138$ & $1$ & Not detected \\
     & TopKIntersection ($\mathrm{TopK}$) & $\uparrow$ & $1477$ & $0.0149228$ & $[0.0114229,\, 0.0184227]$ & $0.217624$ & $1.38944 \times 10^{-16}$ & $7.78088 \times 10^{-15}$ & \textbf{Favors left} \\
     & RelevanceRankAccuracy ($\mathrm{RRA}$) & $\uparrow$ & $1477$ & $0.00366871$ & $[0.00259001,\, 0.00474741]$ & $0.17359$ & $3.57309 \times 10^{-11}$ & $1.71508 \times 10^{-9}$ & \textbf{Favors left} \\
     & RandomLogit & $\downarrow$ & $1480$ & $-0.0759771$ & $[-0.0797696,\, -0.0721846]$ & $-1.02148$ & $6.93134 \times 10^{-232}$ & $6.65409 \times 10^{-230}$ & \textbf{Favors left} \\
     & DeletionAUC & $\downarrow$ & $1480$ & $-0.00627578$ & $[-0.00835107,\, -0.0042005]$ & $-0.154192$ & $3.7226 \times 10^{-9}$ & $1.63794 \times 10^{-7}$ & \textbf{Favors left} \\
     & InsertionAUC & $\uparrow$ & $1480$ & $0.01113$ & $[0.00800211,\, 0.014258]$ & $0.181432$ & $4.44764 \times 10^{-12}$ & $2.2683 \times 10^{-10}$ & \textbf{Favors left} \\
    \bottomrule
  \end{tabular}%
  }
\end{table*}

\begin{table*}[t]
  \centering
  \caption{Complete paired Student $t$-test results for contrast $\mathbf{G_{16}-S}$ (Expected Gaussian IG ($16\!\times\!100$) $-$ Gaussian SPI ($16\!\times\!100$); reference averaging vs.\ stochastic path deviations ($G_{16}$ vs.\ $S$, Hypothesis~\circnumD{D})). Differences and Cohen's $d_z$ are raw $\text{left} - \text{right}$ (positive favors left for $\uparrow$ metrics; negative favors left for $\downarrow$ metrics). Multiplicity adjustment uses Holm's step-down procedure across the $m=48$ applicable tests of family $\mathbf{G_{16}-S}$. Tally across the $48$ applicable tests: $\mathrm{L}=40$ Favors left, $\mathrm{ND}=7$ Not detected, $\mathrm{R}=1$ Favors right ($\mathrm{NA}=6$ Flowers-102 localization exclusions).}
  \label{tab:paired-g16s-gaussian}
  \scriptsize
  \setlength{\tabcolsep}{4pt}
  \renewcommand{\arraystretch}{0.96}
  \resizebox{\textwidth}{!}{%
  \begin{tabular}{ll c r r c r r r l}
    \toprule
    \textbf{Dataset $\times$ Backbone} & \textbf{Metric} & \textbf{Pol.} & $n$ & \textbf{Mean diff.\ ($\overline{\Delta}$)} & \textbf{95\% CI (unadjusted)} & \textbf{Cohen $d_z$} & $p_{\mathrm{raw}}$ & $p_{\mathrm{Holm}}$ & \textbf{Verdict} \\
    \midrule
    \multirow{9}{*}{Flowers-102 $\times$ PVT-v2-B2} & FaithfulnessCorrelation ($\mathrm{FC}$) & $\uparrow$ & $1020$ & $0.160598$ & $[0.15289,\, 0.168305]$ & $1.28017$ & $4.53986 \times 10^{-217}$ & $1.90674 \times 10^{-215}$ & \textbf{Favors left} \\
     & MonotonicityCorrelation ($\mathrm{MC}$) & $\uparrow$ & $1020$ & $0.112822$ & $[0.0947414,\, 0.130903]$ & $0.383389$ & $2.93783 \times 10^{-32}$ & $5.58188 \times 10^{-31}$ & \textbf{Favors left} \\
     & FaithfulnessEstimate ($\mathrm{FE}$) & $\uparrow$ & $1020$ & $0.129764$ & $[0.113927,\, 0.145601]$ & $0.50344$ & $5.16782 \times 10^{-52}$ & $1.34363 \times 10^{-50}$ & \textbf{Favors left} \\
     & PointingGame ($\mathrm{PG}$) & $\uparrow$ & \textnormal{---} & \textnormal{---} & $[\textnormal{---},\, \textnormal{---}]$ & \textnormal{---} & \textnormal{---} & \textnormal{---} & \textnormal{Not applicable} \\
     & TopKIntersection ($\mathrm{TopK}$) & $\uparrow$ & \textnormal{---} & \textnormal{---} & $[\textnormal{---},\, \textnormal{---}]$ & \textnormal{---} & \textnormal{---} & \textnormal{---} & \textnormal{Not applicable} \\
     & RelevanceRankAccuracy ($\mathrm{RRA}$) & $\uparrow$ & \textnormal{---} & \textnormal{---} & $[\textnormal{---},\, \textnormal{---}]$ & \textnormal{---} & \textnormal{---} & \textnormal{---} & \textnormal{Not applicable} \\
     & RandomLogit & $\downarrow$ & $1020$ & $-0.127986$ & $[-0.13331,\, -0.122661]$ & $-1.47681$ & $1.89324 \times 10^{-258}$ & $8.14093 \times 10^{-257}$ & \textbf{Favors left} \\
     & DeletionAUC & $\downarrow$ & $1020$ & $-0.0049638$ & $[-0.0084038,\, -0.00152379]$ & $-0.0886582$ & $0.00472397$ & $0.0377918$ & \textbf{Favors left} \\
     & InsertionAUC & $\uparrow$ & $1020$ & $0.0471077$ & $[0.0403033,\, 0.0539121]$ & $0.425369$ & $9.32005 \times 10^{-39}$ & $2.05041 \times 10^{-37}$ & \textbf{Favors left} \\
    \midrule
    \multirow{9}{*}{Flowers-102 $\times$ ResNet-50} & FaithfulnessCorrelation ($\mathrm{FC}$) & $\uparrow$ & $1020$ & $0.0837433$ & $[0.0754336,\, 0.0920531]$ & $0.619194$ & $6.3359 \times 10^{-74}$ & $1.90077 \times 10^{-72}$ & \textbf{Favors left} \\
     & MonotonicityCorrelation ($\mathrm{MC}$) & $\uparrow$ & $1020$ & $0.188635$ & $[0.174195,\, 0.203076]$ & $0.802625$ & $3.05879 \times 10^{-112}$ & $1.13175 \times 10^{-110}$ & \textbf{Favors left} \\
     & FaithfulnessEstimate ($\mathrm{FE}$) & $\uparrow$ & $1020$ & $0.148376$ & $[0.134641,\, 0.162111]$ & $0.663725$ & $6.6118 \times 10^{-83}$ & $2.11578 \times 10^{-81}$ & \textbf{Favors left} \\
     & PointingGame ($\mathrm{PG}$) & $\uparrow$ & \textnormal{---} & \textnormal{---} & $[\textnormal{---},\, \textnormal{---}]$ & \textnormal{---} & \textnormal{---} & \textnormal{---} & \textnormal{Not applicable} \\
     & TopKIntersection ($\mathrm{TopK}$) & $\uparrow$ & \textnormal{---} & \textnormal{---} & $[\textnormal{---},\, \textnormal{---}]$ & \textnormal{---} & \textnormal{---} & \textnormal{---} & \textnormal{Not applicable} \\
     & RelevanceRankAccuracy ($\mathrm{RRA}$) & $\uparrow$ & \textnormal{---} & \textnormal{---} & $[\textnormal{---},\, \textnormal{---}]$ & \textnormal{---} & \textnormal{---} & \textnormal{---} & \textnormal{Not applicable} \\
     & RandomLogit & $\downarrow$ & $1020$ & $-0.157527$ & $[-0.162878,\, -0.152176]$ & $-1.80883$ & $0$ & $0$ & \textbf{Favors left} \\
     & DeletionAUC & $\downarrow$ & $1020$ & $-0.000665282$ & $[-0.00238625,\, 0.00105569]$ & $-0.0237518$ & $0.448284$ & $1$ & Not detected \\
     & InsertionAUC & $\uparrow$ & $1020$ & $0.0223793$ & $[0.0194543,\, 0.0253044]$ & $0.470092$ & $3.53457 \times 10^{-46}$ & $8.83642 \times 10^{-45}$ & \textbf{Favors left} \\
    \midrule
    \multirow{9}{*}{ImageNet $\times$ PVT-v2-B2} & FaithfulnessCorrelation ($\mathrm{FC}$) & $\uparrow$ & $1000$ & $0.10848$ & $[0.10032,\, 0.11664]$ & $0.824953$ & $7.8883 \times 10^{-115}$ & $2.99756 \times 10^{-113}$ & \textbf{Favors left} \\
     & MonotonicityCorrelation ($\mathrm{MC}$) & $\uparrow$ & $1000$ & $0.182699$ & $[0.165344,\, 0.200054]$ & $0.653253$ & $3.18314 \times 10^{-79}$ & $9.86772 \times 10^{-78}$ & \textbf{Favors left} \\
     & FaithfulnessEstimate ($\mathrm{FE}$) & $\uparrow$ & $1000$ & $0.148491$ & $[0.133514,\, 0.163467]$ & $0.615252$ & $9.6071 \times 10^{-72}$ & $2.78606 \times 10^{-70}$ & \textbf{Favors left} \\
     & PointingGame ($\mathrm{PG}$) & $\uparrow$ & $1000$ & $-0.021$ & $[-0.0457222,\, 0.00372221]$ & $-0.0527116$ & $0.09585$ & $0.47925$ & Not detected \\
     & TopKIntersection ($\mathrm{TopK}$) & $\uparrow$ & $1000$ & $-0.035202$ & $[-0.0421745,\, -0.0282295]$ & $-0.313294$ & $3.90325 \times 10^{-22}$ & $6.63552 \times 10^{-21}$ & \underline{Favors right} \\
     & RelevanceRankAccuracy ($\mathrm{RRA}$) & $\uparrow$ & $1000$ & $0.00272275$ & $[0.000673369,\, 0.00477213]$ & $0.0824441$ & $0.00926705$ & $0.0563206$ & Not detected \\
     & RandomLogit & $\downarrow$ & $1000$ & $-0.161889$ & $[-0.169929,\, -0.153849]$ & $-1.24947$ & $2.29426 \times 10^{-206}$ & $9.40645 \times 10^{-205}$ & \textbf{Favors left} \\
     & DeletionAUC & $\downarrow$ & $1000$ & $-0.0206013$ & $[-0.0247875,\, -0.0164151]$ & $-0.305386$ & $3.72224 \times 10^{-21}$ & $5.58336 \times 10^{-20}$ & \textbf{Favors left} \\
     & InsertionAUC & $\uparrow$ & $1000$ & $0.0433395$ & $[0.0365716,\, 0.0501075]$ & $0.397378$ & $9.92691 \times 10^{-34}$ & $1.98538 \times 10^{-32}$ & \textbf{Favors left} \\
    \midrule
    \multirow{9}{*}{ImageNet $\times$ ResNet-50} & FaithfulnessCorrelation ($\mathrm{FC}$) & $\uparrow$ & $1000$ & $0.105424$ & $[0.0971248,\, 0.113724]$ & $0.788247$ & $4.9557 \times 10^{-107}$ & $1.78405 \times 10^{-105}$ & \textbf{Favors left} \\
     & MonotonicityCorrelation ($\mathrm{MC}$) & $\uparrow$ & $1000$ & $0.0788678$ & $[0.0637607,\, 0.0939749]$ & $0.323962$ & $1.73176 \times 10^{-23}$ & $3.11717 \times 10^{-22}$ & \textbf{Favors left} \\
     & FaithfulnessEstimate ($\mathrm{FE}$) & $\uparrow$ & $1000$ & $0.130235$ & $[0.112986,\, 0.147485]$ & $0.468516$ & $4.83092 \times 10^{-45}$ & $1.15942 \times 10^{-43}$ & \textbf{Favors left} \\
     & PointingGame ($\mathrm{PG}$) & $\uparrow$ & $1000$ & $-0.007$ & $[-0.0242226,\, 0.0102226]$ & $-0.0252217$ & $0.425304$ & $1$ & Not detected \\
     & TopKIntersection ($\mathrm{TopK}$) & $\uparrow$ & $1000$ & $-0.001798$ & $[-0.00605592,\, 0.00245992]$ & $-0.0262039$ & $0.407506$ & $1$ & Not detected \\
     & RelevanceRankAccuracy ($\mathrm{RRA}$) & $\uparrow$ & $1000$ & $0.00246049$ & $[0.000642236,\, 0.00427874]$ & $0.0839734$ & $0.0080458$ & $0.0563206$ & Not detected \\
     & RandomLogit & $\downarrow$ & $1000$ & $-0.181774$ & $[-0.187743,\, -0.175806]$ & $-1.88996$ & $0$ & $0$ & \textbf{Favors left} \\
     & DeletionAUC & $\downarrow$ & $1000$ & $-0.00718153$ & $[-0.00907657,\, -0.00528648]$ & $-0.235164$ & $2.22085 \times 10^{-13}$ & $2.66502 \times 10^{-12}$ & \textbf{Favors left} \\
     & InsertionAUC & $\uparrow$ & $1000$ & $0.0457931$ & $[0.0405842,\, 0.051002]$ & $0.545542$ & $1.42512 \times 10^{-58}$ & $3.84784 \times 10^{-57}$ & \textbf{Favors left} \\
    \midrule
    \multirow{9}{*}{Oxford Pets $\times$ PVT-v2-B2} & FaithfulnessCorrelation ($\mathrm{FC}$) & $\uparrow$ & $1480$ & $0.208656$ & $[0.202693,\, 0.214619]$ & $1.78421$ & $0$ & $0$ & \textbf{Favors left} \\
     & MonotonicityCorrelation ($\mathrm{MC}$) & $\uparrow$ & $1479$ & $0.0988418$ & $[0.0845299,\, 0.113154]$ & $0.35226$ & $1.6903 \times 10^{-39}$ & $3.88768 \times 10^{-38}$ & \textbf{Favors left} \\
     & FaithfulnessEstimate ($\mathrm{FE}$) & $\uparrow$ & $1480$ & $0.128222$ & $[0.116892,\, 0.139553]$ & $0.577008$ & $1.84496 \times 10^{-94}$ & $6.08836 \times 10^{-93}$ & \textbf{Favors left} \\
     & PointingGame ($\mathrm{PG}$) & $\uparrow$ & $1475$ & $0.0535593$ & $[0.029146,\, 0.0779726]$ & $0.112052$ & $1.79262 \times 10^{-5}$ & $0.000161336$ & \textbf{Favors left} \\
     & TopKIntersection ($\mathrm{TopK}$) & $\uparrow$ & $1475$ & $0.0235342$ & $[0.0178745,\, 0.029194]$ & $0.212378$ & $7.30508 \times 10^{-16}$ & $9.49661 \times 10^{-15}$ & \textbf{Favors left} \\
     & RelevanceRankAccuracy ($\mathrm{RRA}$) & $\uparrow$ & $1475$ & $0.0198664$ & $[0.0169023,\, 0.0228304]$ & $0.342324$ & $2.05773 \times 10^{-37}$ & $4.32123 \times 10^{-36}$ & \textbf{Favors left} \\
     & RandomLogit & $\downarrow$ & $1480$ & $-0.104535$ & $[-0.109122,\, -0.0999474]$ & $-1.16192$ & $7.92173 \times 10^{-277}$ & $3.48556 \times 10^{-275}$ & \textbf{Favors left} \\
     & DeletionAUC & $\downarrow$ & $1480$ & $-0.0133733$ & $[-0.0181867,\, -0.00855989]$ & $-0.141664$ & $5.89531 \times 10^{-8}$ & $5.89531 \times 10^{-7}$ & \textbf{Favors left} \\
     & InsertionAUC & $\uparrow$ & $1480$ & $0.0604824$ & $[0.0553798,\, 0.065585]$ & $0.604378$ & $3.54428 \times 10^{-102}$ & $1.2405 \times 10^{-100}$ & \textbf{Favors left} \\
    \midrule
    \multirow{9}{*}{Oxford Pets $\times$ ResNet-50} & FaithfulnessCorrelation ($\mathrm{FC}$) & $\uparrow$ & $1480$ & $0.106408$ & $[0.0990879,\, 0.113728]$ & $0.741192$ & $7.01064 \times 10^{-143}$ & $2.80426 \times 10^{-141}$ & \textbf{Favors left} \\
     & MonotonicityCorrelation ($\mathrm{MC}$) & $\uparrow$ & $1480$ & $0.107674$ & $[0.0961015,\, 0.119247]$ & $0.474397$ & $2.83872 \times 10^{-67}$ & $7.94842 \times 10^{-66}$ & \textbf{Favors left} \\
     & FaithfulnessEstimate ($\mathrm{FE}$) & $\uparrow$ & $1480$ & $0.170711$ & $[0.158288,\, 0.183133]$ & $0.700691$ & $1.60766 \times 10^{-130}$ & $6.26987 \times 10^{-129}$ & \textbf{Favors left} \\
     & PointingGame ($\mathrm{PG}$) & $\uparrow$ & $1477$ & $0.00880162$ & $[-0.0135432,\, 0.0311464]$ & $0.0201048$ & $0.439845$ & $1$ & Not detected \\
     & TopKIntersection ($\mathrm{TopK}$) & $\uparrow$ & $1477$ & $0.0176256$ & $[0.0134871,\, 0.0217641]$ & $0.217377$ & $1.50035 \times 10^{-16}$ & $2.10049 \times 10^{-15}$ & \textbf{Favors left} \\
     & RelevanceRankAccuracy ($\mathrm{RRA}$) & $\uparrow$ & $1477$ & $0.0274692$ & $[0.0251485,\, 0.0297898]$ & $0.604163$ & $6.5123 \times 10^{-102}$ & $2.21418 \times 10^{-100}$ & \textbf{Favors left} \\
     & RandomLogit & $\downarrow$ & $1480$ & $-0.137093$ & $[-0.141696,\, -0.13249]$ & $-1.51856$ & $0$ & $0$ & \textbf{Favors left} \\
     & DeletionAUC & $\downarrow$ & $1480$ & $-0.0060827$ & $[-0.00782851,\, -0.0043369]$ & $-0.177653$ & $1.19964 \times 10^{-11}$ & $1.31961 \times 10^{-10}$ & \textbf{Favors left} \\
     & InsertionAUC & $\uparrow$ & $1480$ & $0.0237528$ & $[0.0189382,\, 0.0285673]$ & $0.251554$ & $1.60808 \times 10^{-21}$ & $2.57293 \times 10^{-20}$ & \textbf{Favors left} \\
    \bottomrule
  \end{tabular}%
  }
\end{table*}

\clearpage

\section{Qualitative Attribution Visualizations}
\label{app:qualitative_vis}

\begin{figure*}[p]
  \centering
  \includegraphics[width=0.95\textwidth]{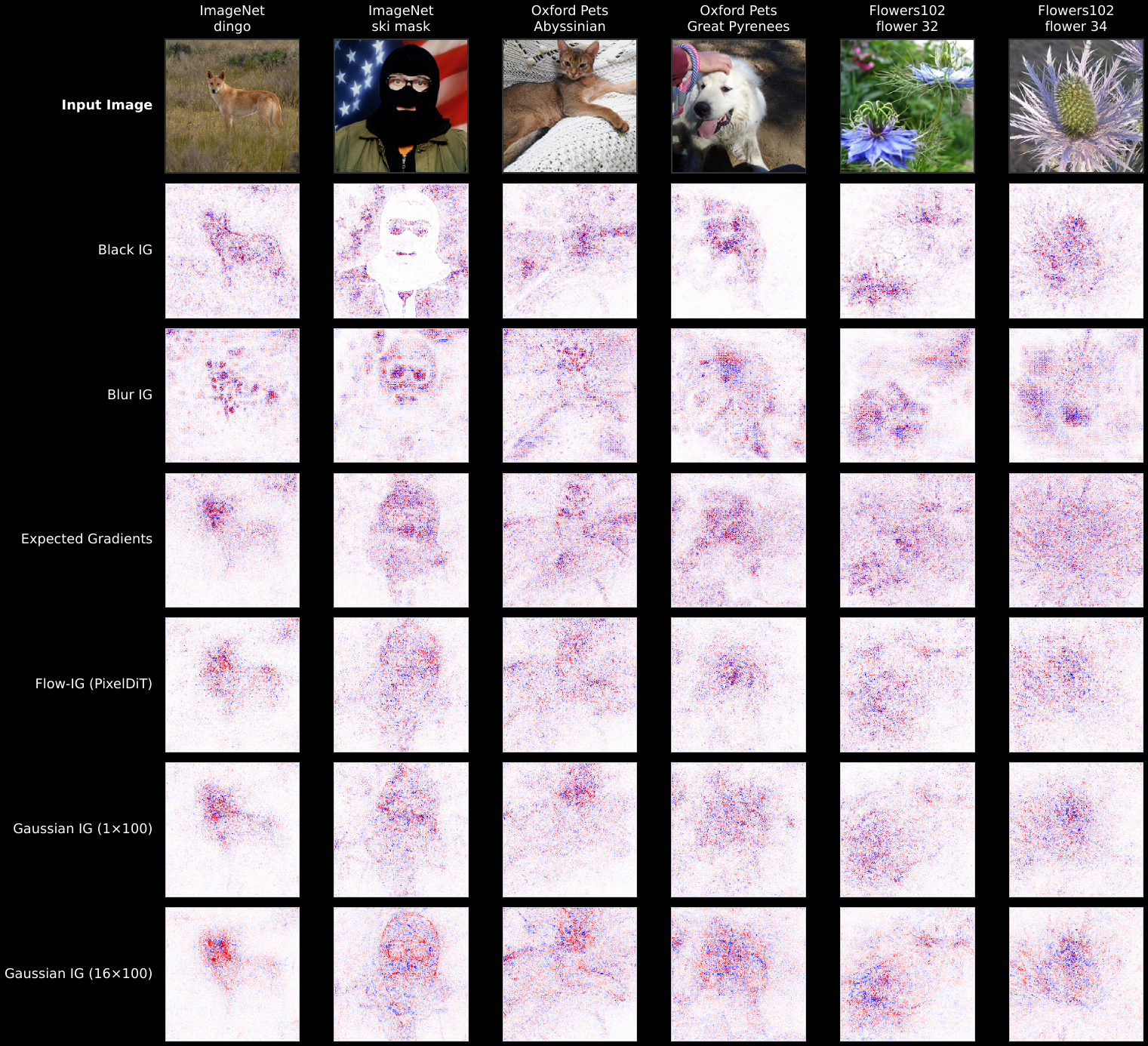}
  \caption{\textbf{ResNet-50 qualitative attribution grid (signed channel sum $S = \sum_c A_c$).} Six shared random samples ($\mathrm{seed}=42$) across ImageNet, Oxford-IIIT Pet, and Oxford Flowers-102, rendered with diverging red ($+$) / blue ($-$) on white ($0$) and $99.5$th-percentile clipping of $\lvert S\rvert$.}
  \label{fig:qual_resnet_signed}
\end{figure*}

\begin{figure*}[p]
  \centering
  \includegraphics[width=0.95\textwidth]{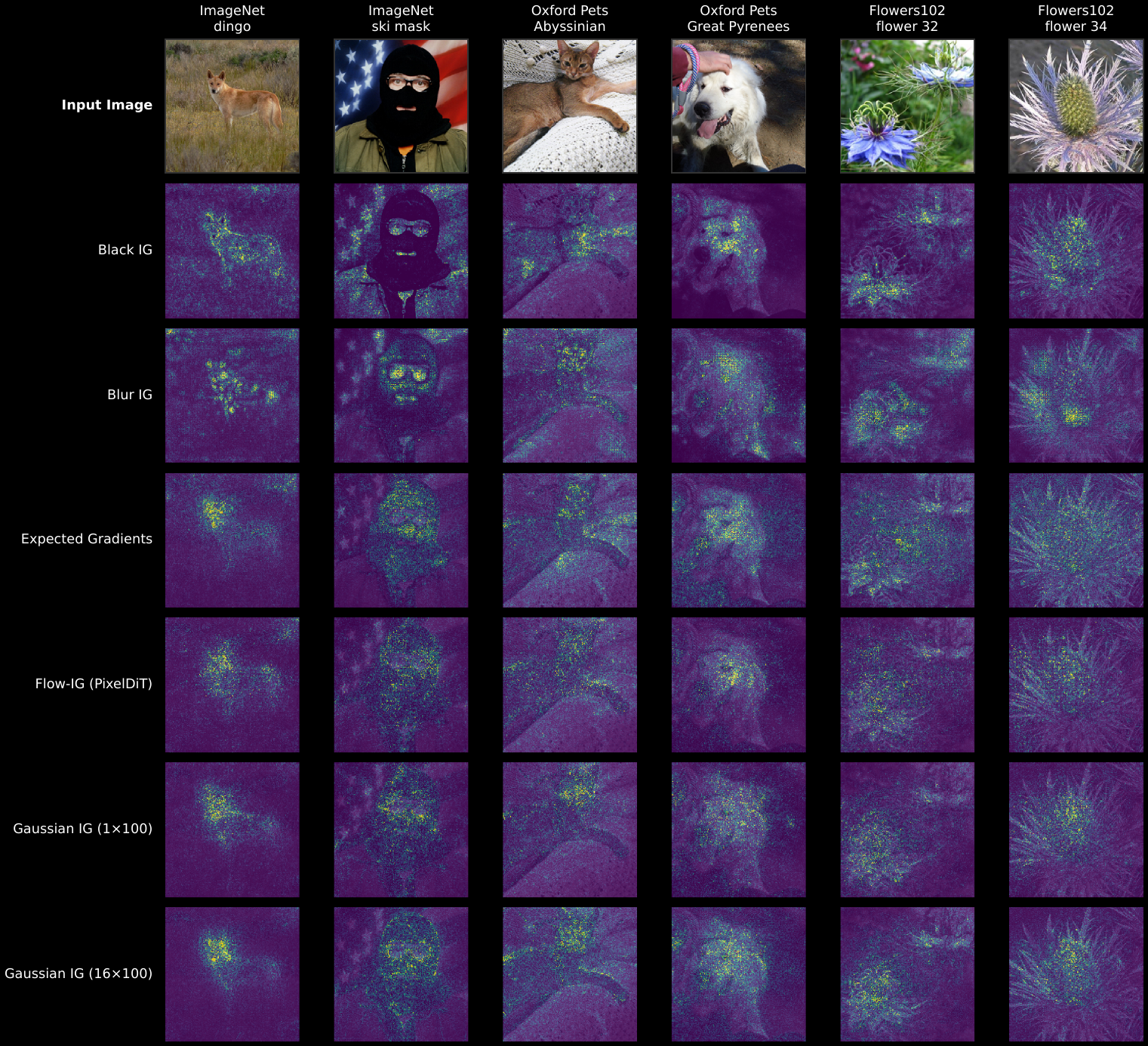}
  \caption{\textbf{ResNet-50 qualitative attribution grid (magnitude $\lvert\sum_c A_c\rvert$).} Identical samples and explainers as Figure~\ref{fig:qual_resnet_signed}, rendered in the \texttt{viridis} colormap (purple $= 0$, yellow $=$ $99.5$th-percentile clipped maximum) with a $0.15$ grayscale overlay of the input image to match the unsigned input of spatial localization metrics ($\mathrm{PG}$, $\mathrm{TopK}$, $\mathrm{RRA}$).}
  \label{fig:qual_resnet_mag}
\end{figure*}

\begin{figure*}[p]
  \centering
  \includegraphics[width=0.95\textwidth]{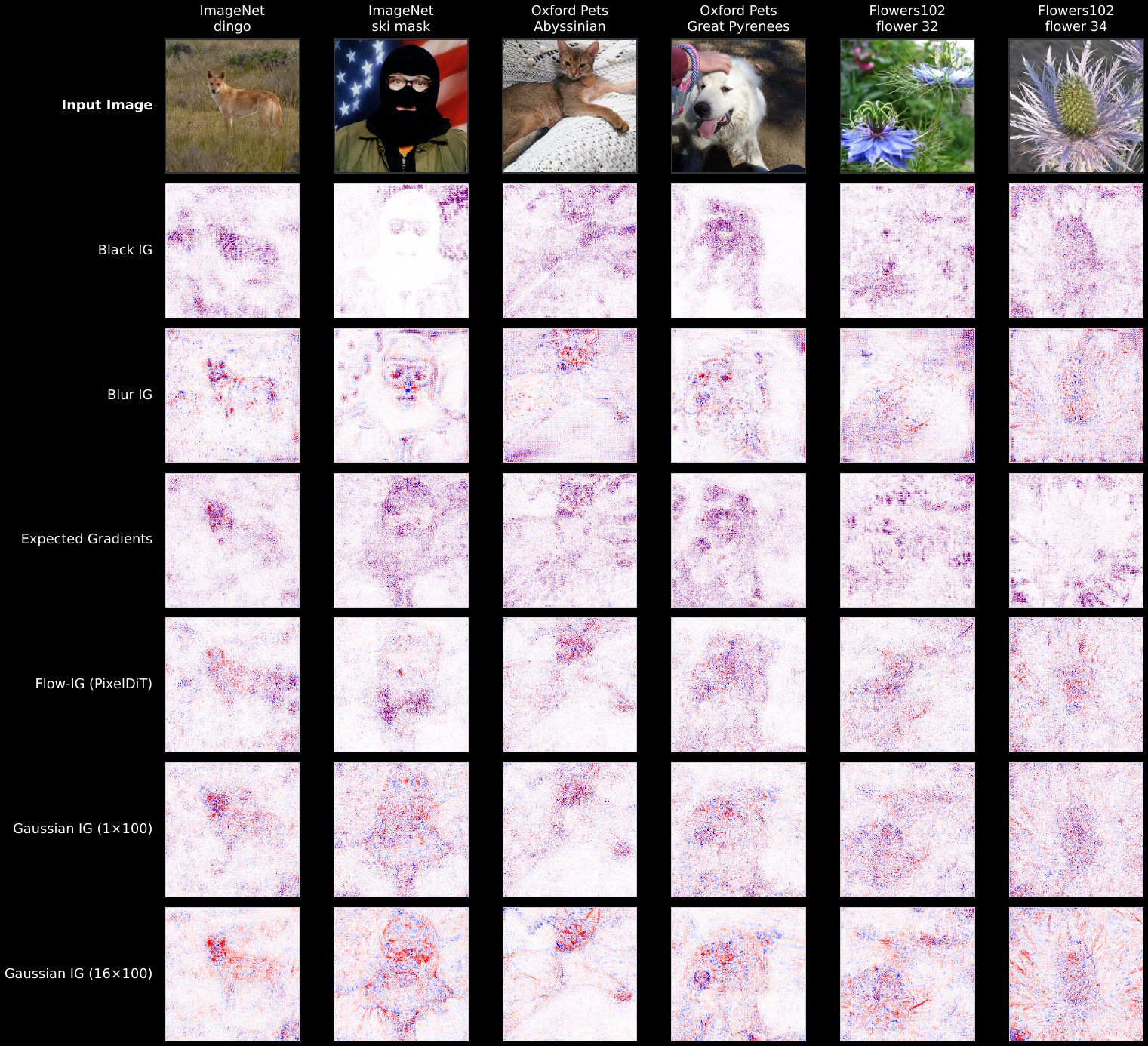}
  \caption{\textbf{PVT-v2-B2 qualitative attribution grid (signed channel sum $S = \sum_c A_c$).} Identical seed-42 samples and rendering protocol as Figure~\ref{fig:qual_resnet_signed} on the hierarchical Vision Transformer backbone.}
  \label{fig:qual_pvt_signed}
\end{figure*}

\begin{figure*}[p]
  \centering
  \includegraphics[width=0.95\textwidth]{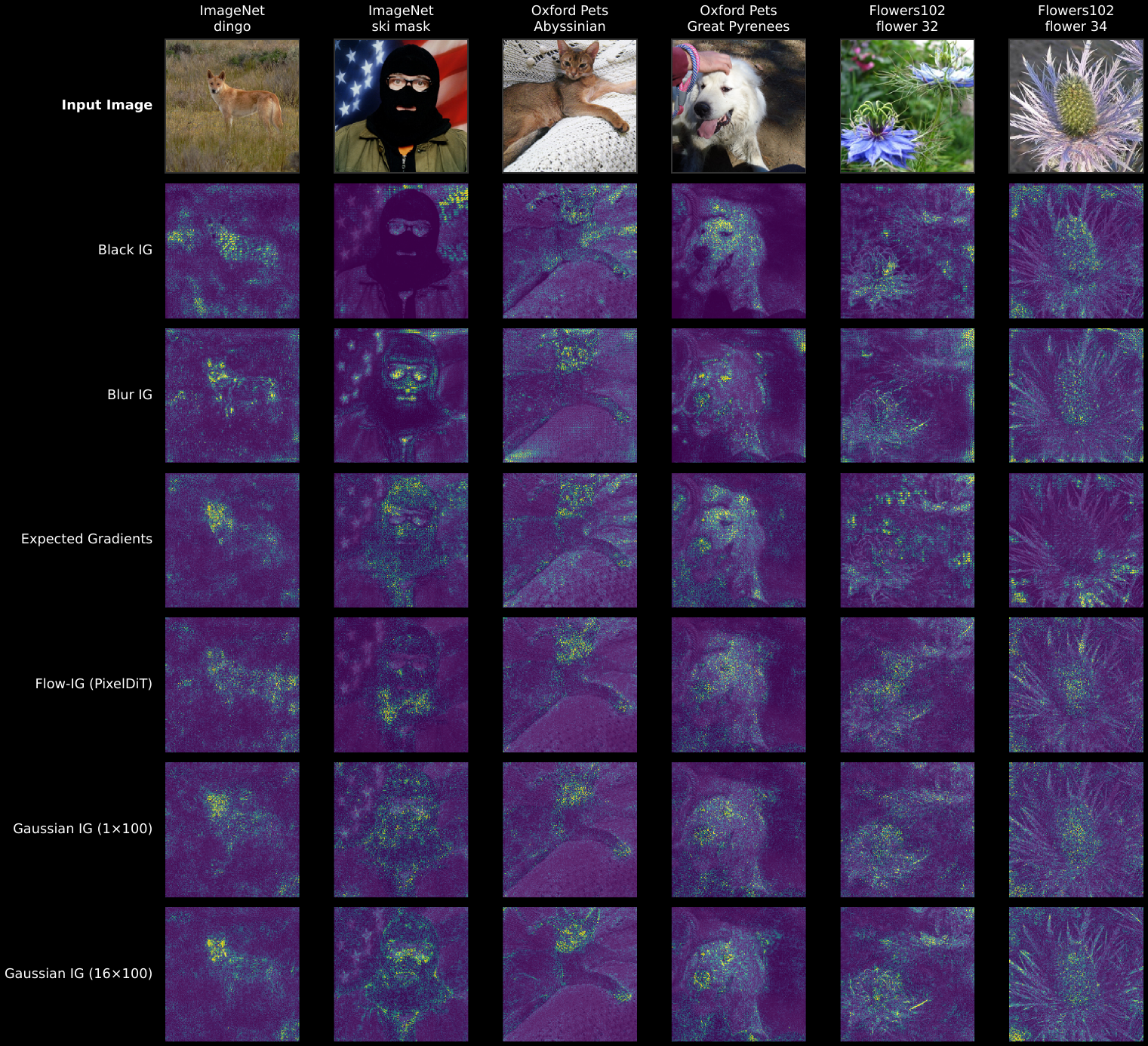}
  \caption{\textbf{PVT-v2-B2 qualitative attribution grid (magnitude $\lvert\sum_c A_c\rvert$).} Identical samples and explainers as Figure~\ref{fig:qual_pvt_signed}, rendered in the \texttt{viridis} colormap with $99.5$th-percentile clipping and a $0.15$ grayscale overlay of the input image.}
  \label{fig:qual_pvt_mag}
\end{figure*}

Figures~\ref{fig:qual_resnet_signed}--\ref{fig:qual_pvt_mag} provide qualitative attribution grids across six shared evaluation samples drawn uniformly at random ($\mathrm{seed}=42$, two per benchmark across ImageNet, Oxford-IIIT Pet, and Oxford Flowers-102) on both \textbf{ResNet-50} and \textbf{PVT-v2-B2}. Each grid compares \textbf{Black IG}, \textbf{Blur IG}, \textbf{Expected Gradients}, \textbf{Flow-IG (PixelDiT)}, \textbf{Gaussian IG} ($1\!\times\!100$), and \textbf{Expected Gaussian IG} ($16\!\times\!100$) on identical inputs.

\paragraph{Rendering protocol.}
Matching the channel-reduction convention used by our evaluation metrics (Appendix~\ref{app:metrics}), saved native-RGB attribution tensors $A \in \mathbb{R}^{3 \times 224 \times 224}$ are reduced via the signed channel sum $S_{h,w} = \sum_{c \in \{R,G,B\}} A_{c,h,w}$ and normalized per map after clipping $\lvert S_{h,w}\rvert$ at its $99.5$th percentile so isolated gradient spikes do not compress visual contrast. For each backbone we present two complementary views:
\begin{itemize}
  \item \textbf{Signed channel sum ($S_{h,w}$; Figures~\ref{fig:qual_resnet_signed} and~\ref{fig:qual_pvt_signed}):} diverging red ($+$) / blue ($-$) on white ($0$), matching the signed ranking evaluated by Deletion AUC, Insertion AUC, and Faithfulness Correlation.
  \item \textbf{Magnitude ($\lvert S_{h,w}\rvert$; Figures~\ref{fig:qual_resnet_mag} and~\ref{fig:qual_pvt_mag}):} rendered with the \texttt{viridis} colormap (dark purple $= 0$, bright yellow $=$ clipped maximum) blended with a $0.15$-strength grayscale overlay of the input image, matching the unsigned attribution magnitude evaluated by spatial localization metrics ($\mathrm{PG}$, $\mathrm{TopK}$, $\mathrm{RRA}$).
\end{itemize}

\clearpage

\end{document}